\PassOptionsToPackage{table}{xcolor}
\documentclass[10pt,twocolumn,letterpaper]{article}

\usepackage{cvpr}              %

\usepackage{amsmath}
\usepackage{graphicx}
\usepackage{bm}
\usepackage{booktabs}
\usepackage{multirow}
\usepackage{bigstrut}
\usepackage{amsmath,amssymb,amsfonts}
\usepackage{amsmath}
\DeclareMathOperator*{\argmax}{arg\,max}

\usepackage{bm}

\usepackage{algorithm}
\usepackage{algpseudocode}
\usepackage{enumitem}

\usepackage{booktabs}
\usepackage{graphicx}
\usepackage{makecell}
\newcommand{\meanstd}[2]{%
  \ensuremath{#1{\mskip2mu{\scriptstyle\pm}\mskip1mu{\scriptscriptstyle #2}}}%
}
\newcommand{\meannostd}[2]{%
  \ensuremath{#1}%
}

\newcommand{\bestone}[1]{\ensuremath{\bm{#1}}}
\newcommand{\besttwo}[1]{\underline{#1}}
\newcommand{\bestthree}[1]{\underline{#1}}

\usepackage{algpseudocode}

\algrenewcommand\algorithmicrequire{\textbf{Input:}}
\algrenewcommand\algorithmicensure{\textbf{Output:}}

\renewcommand{\paragraph}[1]{\par\noindent\textbf{#1}\;}

\usepackage{xr-hyper}
\usepackage{refcount}
\usepackage[accsupp]{axessibility}

\DeclareMathOperator{\tr}{tr}

\makeatletter
  \providecommand*{\theHALG@line}{\thealgorithm.\arabic{ALG@line}}
  \makeatother

\definecolor{cvprblue}{rgb}{0.21,0.49,0.74}
\usepackage[pagebackref,breaklinks,colorlinks,allcolors=cvprblue]{hyperref}

\def\paperID{36795} %
\def\confName{CVPR}
\def\confYear{2026}

\title{Uncertainty-driven 3D Gaussian Splatting Active Mapping via \\ Anisotropic Visibility Field}

\author{Shangjie Xue\thanks{Equal contribution. \quad $^\dagger$xsj@gatech.edu}\hspace{4pt}$^\dagger$
\quad
Jesse Dill\footnotemark[1]
\quad
Dhruv Ahuja\footnotemark[1]
\quad
Frank Dellaert
\quad
Panagiotis Tsiotras
\quad
Danfei Xu\\
    Georgia Institute of Technology
}

\begin{document}
\maketitle
\begin{abstract}

We present Gaussian Splatting Anisotropic Visibility Field (GAVIS), a novel framework for uncertainty quantification and active mapping in 3DGS. Our key insight is that regions unseen from the training views yield unreliable predictions from the 3DGS. 
To address this, we introduce a principled and efficient method for quantifying the visibility field in 3DGS, defined as the anisotropic visibility of each particle with respect to the training views, and represented using spherical harmonics. The resulting visibility field is integrated into a Bayesian Network-based uncertainty-aware 3DGS rasterizer, enabling real-time (200 FPS) uncertainty quantification for synthesized views. Active mapping is further performed within a maximum information gain framework building on this formulation. 
Extensive experiments across diverse environments demonstrate that GAVIS consistently and significantly outperforms prior approaches in both accuracy and efficiency. Moreover, beyond standalone use, our method can be applied post-hoc to improve the performance of existing approaches.

\end{abstract}
   
\vspace{-15.0px}

\section{Introduction}

When an autonomous robot is deployed in a new environment, it should explore and understand its surroundings. This process, known as active mapping~\cite{yamauchi1997frontier,bircher2016receding,lluvia2021active}, is essential for applications such as household robots, space robots, and search-and-rescue robots. One common goal of active mapping is to achieve comprehensive visual coverage of the scene, formulated as a planning objective that reduces uncertainty in the reconstructed map~\cite{bircher2018receding,xue2024neural}. A suitable representation is critical for active mapping, as it must not only enable high-quality and fast reconstruction but also enable accurate uncertainty quantification~\cite{placed2023survey}.

\begin{figure}[t]
    \centering
    \includegraphics[width=.8\linewidth,trim={68bp 50bp 265bp 44bp},clip]{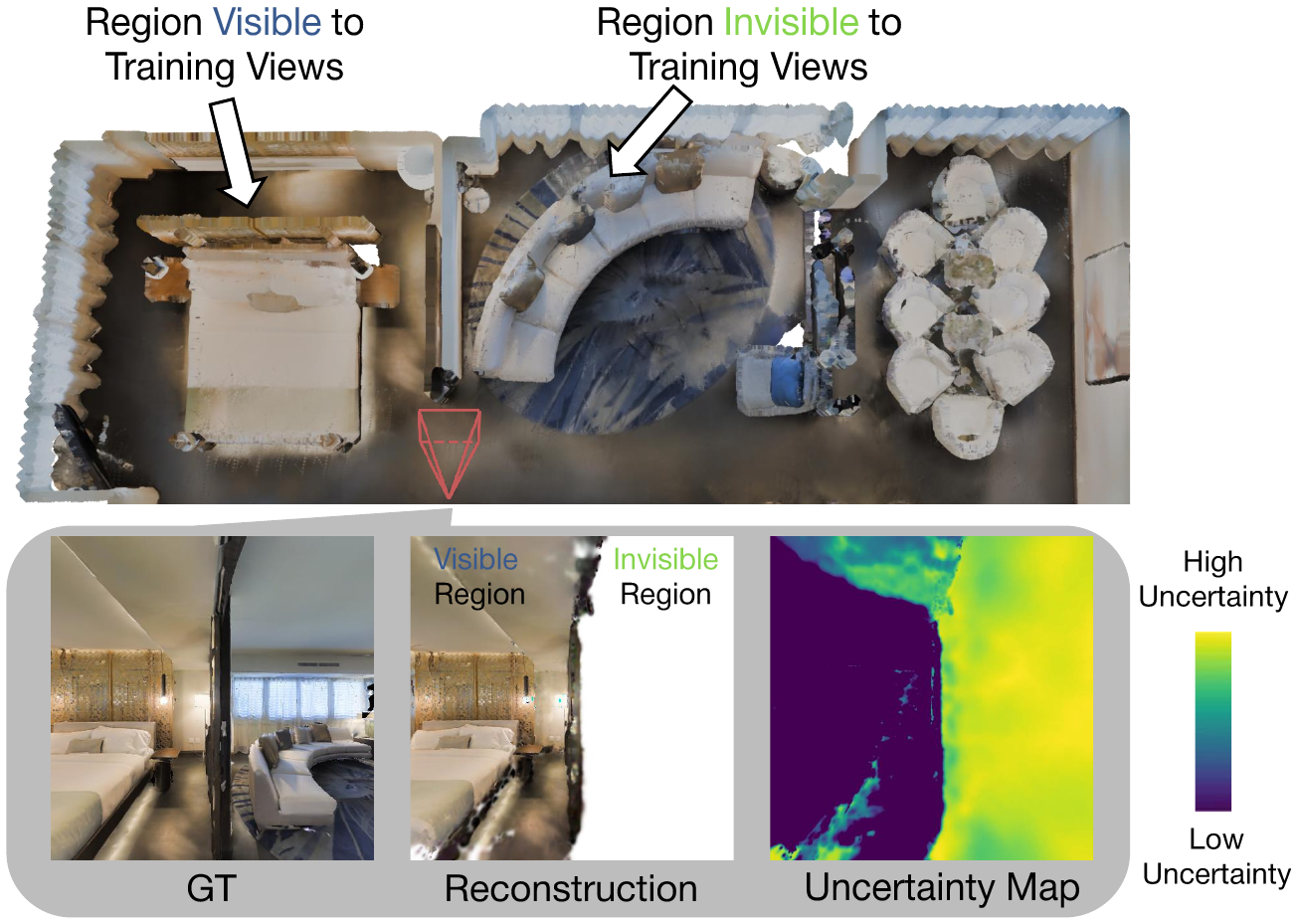}
    \vspace{-5pt}
    \caption{\textbf{GAVIS overview.} Gaussian Splatting Anisotropic Visibility Field (GAVIS) quantifies uncertainty in 3DGS by modeling \emph{visibility}, i.e., whether a region is observed by the training views. Observed regions have low uncertainty (left room), whereas unobserved regions have high uncertainty (right room).}
    \label{fig:teaser}
    \vspace{-18pt}
\end{figure}

\begin{figure*}[t]
    \centering
    \vspace{-13pt}
    \newlength{\pipelineheight}\settoheight{\pipelineheight}{\includegraphics[width=.75\linewidth,trim={8bp 262bp 0bp 17bp},clip]{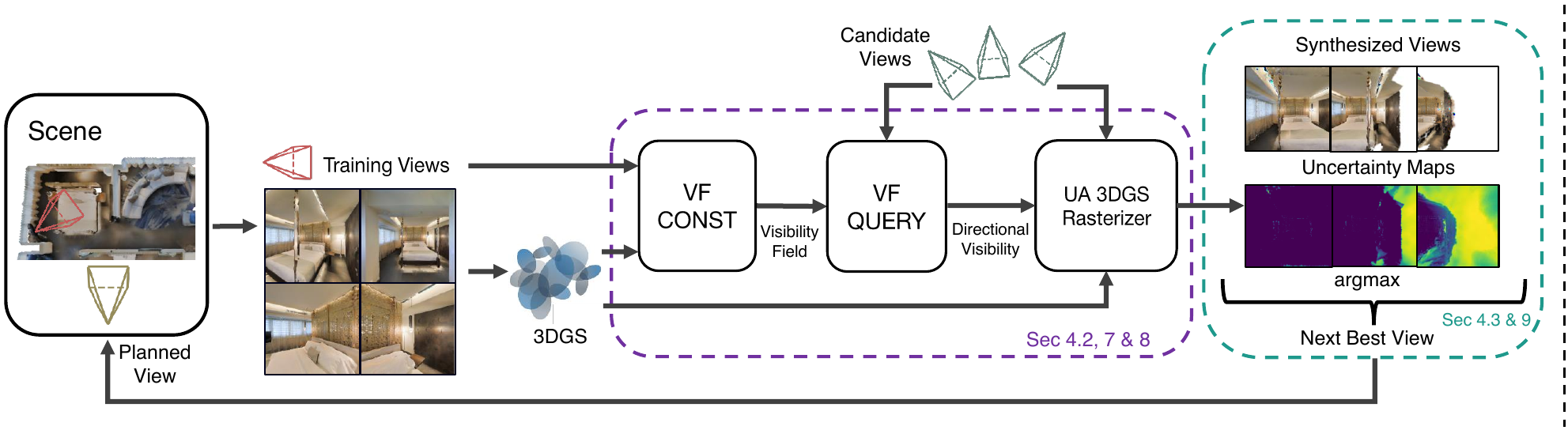}}%
    \includegraphics[width=.75\linewidth,trim={8bp 262bp 0bp 17bp},clip]{figs/pipeline_part.pdf}%
    \includegraphics[height=\pipelineheight,trim={18bp 16bp 19bp 9bp},clip]{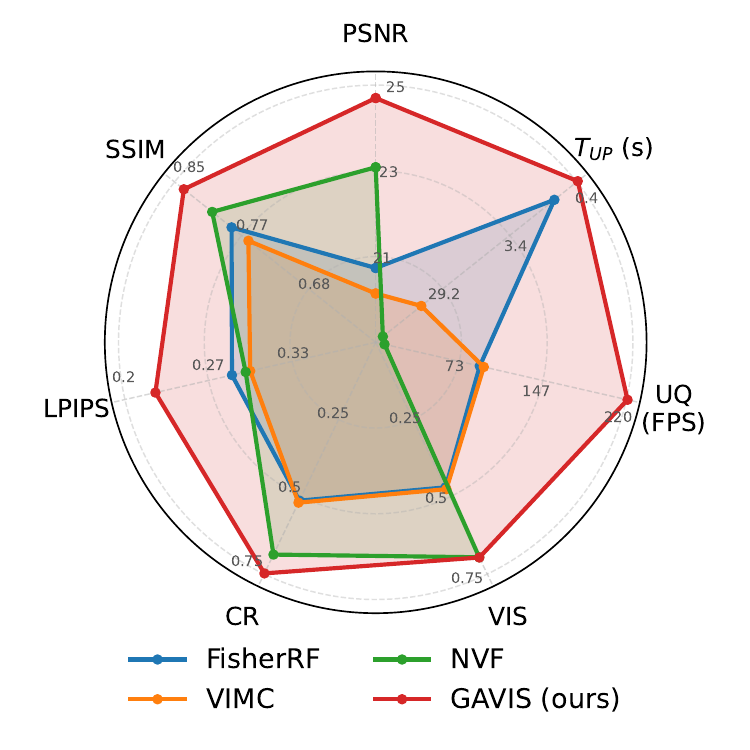}
    \vspace{-10pt}
    \caption{\textbf{GAVIS framework.} (Left) Given a trained 3DGS, GAVIS constructs a visibility field (VF CONST) to represent regions invisible to the training views. It then quantifies uncertainty over sampled candidate views using an uncertainty-aware 3DGS rasterizer (UA 3DGS Rasterizer) that queries the visibility field (VF QUERY). Finally, the maximum-uncertainty view is selected as the next observation. (Right) GAVIS achieves top performance across all evaluation metrics (see Sec.~\ref{sec:exp} for details).
}
    \label{fig:pipeline}
    \vspace{-14pt}
\end{figure*}

Recent advancements in radiance fields~\cite{mildenhall2021nerf} have demonstrated remarkable capabilities in high-quality scene reconstruction. Notably, the emergence of 3D Gaussian Splatting (3DGS)~\cite{kerbl20233d} has shown superior reconstruction quality and rendering speed.
However, due to the large number of trainable parameters in 3DGS, accurately and efficiently quantifying its uncertainty remains challenging. 
Recent approaches to uncertainty quantification in 3DGS 
adopted established epistemic uncertainty quantification methods from machine learning, including variational inference~\cite{lyu2024manifold} and Laplace approximation~\cite{jiang2023fisherrf}, to tackle this challenge. 
However, a gap exists between the objective of machine learning uncertainty quantification and active mapping. In active mapping, the robot aims to achieve high-quality reconstruction by providing visual coverage of the entire scene, where predictions for regions never visible in the training views are always considered unreliable. However, as an approximation method, existing learning-based uncertainty quantification techniques are unable to guarantee this, and often underestimate the uncertainty in these unseen, out-of-distribution regions~\cite{gawlikowski2023survey,wang2021rethinking,ovadia2019can}.

We propose 3D Gaussian Splatting Anisotropic Visibility Field (GAVIS), a principled method that reliably assigns high uncertainty to unobserved regions for active mapping, while incurring minimal computational overhead. This method can be used independently or integrated seamlessly, as a post-hoc module, into existing learning-based uncertainty quantification framework that estimates uncertainty in 3DGS parameters or volumetric radiance field outputs, thereby improving its performance.

GAVIS extends the 3D Gaussian Splatting framework to quantify the extent to which each region is covered by the training views, inspired by NVF~\cite{xue2024neural}, which shows that visibility is closely related to regional uncertainty.
However, NVF relies on training a neural network to approximate a view-independent (isotropic) visibility field for NeRF, making it computationally expensive and impractical for real-world applications, while also introducing approximation errors and neglecting the inherently view-dependent nature of visibility. 
In contrast, GAVIS explicitly models the view-dependent visibility of each Gaussian particle relative to all training views (see Sec.~\ref{sec:method.formulation} for details), with analytical, efficient, gradient-free computation. 
The uncertainty-aware 3DGS rasterization is then modeled as a Bayesian network, which integrates both field-based uncertainty and visibility into the ray-based camera observations. Within this framework, the color distribution along a ray is modeled as a Gaussian Mixture Model (GMM). The entropy of the GMM serves as an objective to guide the agent in selecting the optimal poses for active mapping.

Extensive evaluations are conducted across diverse environments, including standard NeRF Blender assets, indoor scenes, and space robot scenarios. The results demonstrate that our method outperforms all existing approaches in both accuracy and efficiency. Moreover, we also demonstrate that our method can be used independently or as a post-hoc method to enhance the performance of existing methods. To summarize, our main contributions are:

\begin{itemize}
    \item We propose a principled uncertainty quantification method for 3DGS that incorporates anisotropic visibility.
\item The proposed method can be used independently or integrated seamlessly as a post-hoc enhancement to improve the performance of existing frameworks.
\item Through extensive evaluations across diverse environments, we demonstrate that our method is both effective for active mapping tasks and computationally more efficient than existing approaches.
\end{itemize}

\vspace{-1.0px}
\section{Related Work}

\paragraph{Uncertainty Quantification in Radiance Fields.}
With the introduction of NeRF~\cite{mildenhall2021nerf},
subsequent work has improved efficiency~\cite{muller2022instant,kerbl20233d,fridovich2022plenoxels} and enabled online SLAM~\cite{keetha2024splatam,matsuki2024gaussian} with explicit representations such as 3DGS~\cite{kerbl20233d}. Uncertainty quantification in radiance fields~\cite{klasson2025sources} has seen growing interest for artifact removal~\cite{goli2023bayes,warburg2023nerfbusters} and active mapping~\cite{lee2022uncertainty,xue2024neural,jiang2023fisherrf}. Parameter uncertainty is estimated via ensembles~\cite{sunderhauf2023density}, variational inference~\cite{shen2021stochastic,lyu2024manifold,li2024variational}, conditional flows~\cite{shen2022conditional}, Fisher information via Laplace approximation~\cite{goli2023bayes,jiang2023fisherrf,li2024bavsnerf}, or Monte Carlo dropout~\cite{gal2016dropout}. Alternatively, radiance field outputs can be modeled as probability distributions~\cite{pan2022activenerf,jin2023neu,neurar,lee2022uncertainty,xue2024neural,lee2024bayesian,yan2023active,wilson2025modeling}, including as Gaussian-distributed RGB~\cite{neurar,pan2022activenerf}, occupancy from NeRF density~\cite{yan2023active,lee2022uncertainty,feng2024naruto,shen2024estimating} or 3DGS opacity~\cite{chen2025activegamer,li2025activesplat,jin2024gs}, or as a volume rendering weighted GMM to incorporate visibility into ray-wise uncertainty~\cite{xue2024neural}.

\paragraph{Active Mapping.}
Active mapping is typically posed as an optimization over information gain~\cite{carlone2014active,valencia2017active,connolly1985determination,whaite1997autonomous,sim2005global,stachniss2005information,stachniss2004exploration,carrillo2012comparison,mu2016information,atanasov2015decentralized,papachristos2017uncertainty,kompis2021informed,meng2017two,respall2021fast,batinovic2022shadowcasting}, and augmented with frontier-based~\cite{yamauchi1997frontier,yamauchi1998frontier,keidar2012robot,keidar2014efficient,holz2010evaluating,qiao2018sample,dornhege2013frontier,senarathne2016towards,lu2020optimal,gomez2019topological,selin2019efficient,umari2017autonomous,dai2020fast,respall2021fast} or sampling-based~\cite{qiao2018sample,selin2019efficient,bircher2016receding,shen2012stochastic,sutantyo2013collective,bircher2018receding,umari2017autonomous,papachristos2017uncertainty,dang2019graph,zhang2018perception,kopanas2023improving} heuristics to reduce computational cost~\cite{placed2023survey}. Recent radiance field active mapping methods pair voxel grids~\cite{jin2025activegs,xu2025hgs,jin2024gs,zeng2025multi,chen2025activegamer,feng2024naruto} or Voronoi graphs~\cite{li2025activesplat,kuang2024active}, while a complementary line of work learns planning policies end-to-end~\cite{kollar2008trajectory,niroui2019deep,chen2020autonomous,wen2020path,zhu2017target}.

\paragraph{Radiance Fields Visibility.}
Visibility has received growing attention in radiance fields for uncertainty quantification and active mapping, providing a measure of where the scene is poorly observed.
Matsuki et al.~\cite{matsuki2024gaussian} define single-view visibility via an occupancy threshold; in 3DGS this reduces to particle opacity~\cite{chen2025activegamer,li2025activesplat}; however, single-view metrics ignore prior observations. Multi-view methods~\cite{xue2024neural,nakayama2024provnerf} aggregate visibility across views, yet face limitations, specifically, NVF~\cite{xue2024neural} requires neural-network training to approximate positional visibility, and ProvNeRF~\cite{nakayama2024provnerf} models per-point provenance via additional post-hoc optimization, both incur minutes to hours of runtime, limiting real-time applicability.
Beyond active mapping, visibility also serves as a cue for inpainting by identifying high-uncertainty regions~\cite{shih2024extranerf}, and has been incorporated into radiance-field pipelines for relighting~\cite{srinivasan2021nerv}, material decomposition~\cite{zhang2021nerfactor}, occlusion-aware view synthesis~\cite{chen2022hallucinated,zhang2024gaussian,shi2021self,huang20243d}, and scene reconstruction~\cite{zhang2025visibility,somraj2023vip}.

\vspace{-1.0px}
\section{Background}
\label{sec:bg}

This section reviews the background on radiance fields, uncertainty-aware volume rendering, and active mapping, with further details available in appendix Sec.~\ref{sec:appx_uvr},~\ref{sec:appx_am}, and \cite{mildenhall2021nerf,kerbl20233d,xue2024neural}.

\paragraph{Radiance field.}
A radiance field \cite{mildenhall2021nerf} $F_\Theta$ maps a 3D location $\bm{x}\in\mathbb{R}^3$ and a view direction $\bm{d}\in\mathbb{S}^2$ to color $\bm{c}(\bm{x},\bm{d})\in\mathbb{R}^3$ and volume density $\sigma(\bm{x})\in\mathbb{R}_{\ge 0}$.
Given a camera ray $\bm{r}(t)=\bm{o}+t\bm{d}$ with near/far bounds $t_n,t_f$, the pixel color is computed via volume rendering:
\begin{equation}
\bm{C}(\bm{r}) = \int_{t_n}^{t_f} T(t)\,\sigma\!\big(\bm{r}(t)\big)\,\bm{c}\!\big(\bm{r}(t),\bm{d}\big)\,\mathrm{d}t,
\label{eq:rf_vr}
\end{equation}
where $T(t) = \exp\!\left(-\int_{t_n}^{t} \sigma\!\big(\bm{r}(s)\big)\,\mathrm{d}s\right)$ represents the (accumulated) transmittance from the near bounds $t_n$ to point $t$. In practice, the integral is discretized and computed numerically.
In NeRF~\cite{mildenhall2021nerf}, $F_\Theta$ is a neural network with parameters $\Theta$; in 3DGS~\cite{kerbl20233d}, the scene is explicitly represented by a trainable mixture of ellipsoidal Gaussian primitives parameterized by $\Theta$.

\paragraph{Active Mapping.} Active mapping, or next-best-view (NBV) planning, aims to determine the robot’s next action that most effectively reduces the uncertainty of the reconstructed map. A practical formulation is to select the action with the highest predicted entropy in observation:
\begin{equation}
\tau^* = \argmax_{\tau} \; \mathcal{H}\!\left(\bm{Z}_{\tau}\right),
\label{eq:i_gain}
\end{equation}
\noindent where $\tau$ denotes the action (i.e., a candidate camera pose) and $\bm{Z}_\tau$ is the random variable which represents the observation obtained after executing action $\tau$, and $\mathcal{H}$ is the entropy.
This objective has been widely adopted in methods that represent scenes using radiance fields \cite{lee2022uncertainty,yan2023active,xue2024neural,lyu2024manifold}, and can be interpreted as an approximation of the fundamental objective of maximizing the expected information gain of the map obtained from new observations (see Sec.~\ref{sec:appx_am} for details). However, a key challenge remains: accurately and efficiently quantifying the uncertainty of synthesized views generated by a radiance field.

\paragraph{Uncertainty-aware Volume Rendering.} To quantify synthesized-view uncertainty, NVF~\cite{xue2024neural} extends deterministic volume rendering of NeRF to a stochastic formulation. By assuming the emitted color $\bm{c}$ is Gaussian, the probability density functions (PDF) of the pixel colors are modeled as Gaussian mixture models (GMM) induced by a Bayesian-network view of uncertainty-aware volume rendering, specifically $p(\bm{z}) = \sum_i w_i \mathcal{N}(\bm{\mu}_{\bm{c}_i}, \bm{Q}_{\bm{c}_i})$, where $p(\bm{z})$ is the PDF of observed pixel color, $\bm{\mu}_{\bm{c}_i}$ and $\bm{Q}_{\bm{c}_i}$ are the mean and variance of emission color at point $i$, $w_i$ is the mixture weight, which can be obtained from \eqref{eq:rf_vr} (details in Sec.~\ref{sec:appx_uvr}).
NVF~\cite{xue2024neural} highlighted the crucial role of visibility in uncertainty quantification and active mapping, as predictions for regions unseen by training views are unreliable. Accordingly, a visibility correction term is incorporated into the GMM of the pixel-color PDF:
\begin{equation} \footnotesize
p(\bm{z}_0) = \sum_i w_i^*\, v_i\, \mathcal{N}\!\big(\bm{\mu}_{\bm{c}_i}, \bm{Q}_{\bm{c}_i}\big)
+ \mathcal{N}\!\big(\bm{\mu}_{0}, \bm{Q}_{0}\big)\,\sum_i w_i^*(1-v_i).
\label{eq:gmm}
\end{equation}
\vspace{-5pt}

\noindent where $w_i^*$ denotes the visibility-corrected mixture weight (see Sec.~\ref{sec:appx_uvr} for details), $\mathcal{N}(\bm{\mu}_{0}, \bm{Q}_{0})$ is a prior distribution with large variance representing emitted color of invisible regions, $v_i$ represents the probability that point $\bm{x}_i$ is visible in at least one training view, quantified as
$v_i = V(\bm{x}_i)$, where $V:\mathbb{R}^3 \to [0,1]$ is the (isotropic) \textbf{visibility field} as a function of position.
In NVF~\cite{xue2024neural}, a neural network is used to both learn the radiance field and approximate the visibility field $V_{\theta}(\bm{x}_i)$.
Moreover, its accuracy is limited since it relies on a neural approximation and models visibility solely as a function of position, ignoring directional anisotropy. Details are discussed in the next section.

\vspace{-1.0px}

\section{Method}

The main challenge in applying radiance-field-based uncertainty quantification to active mapping lies in accurately estimating uncertainty that enables effective exploration while maintaining a practical runtime. Visibility-field-based approach NVF \cite{xue2024neural}, which assigns high uncertainty to unseen regions, could potentially facilitate effective exploration. However, NVF typically require retraining a neural network for the visibility field at every planning step, which takes \textbf{several minutes} and severely limits their applicability to real-world robotic systems.

To address this challenge, we propose 3D Gaussian Splatting Anisotropic Visibility Field (GAVIS), which explicitly models the visibility field using the 3DGS representation. This formulation enables analytical construction of the visibility field without training, achieving construction \textbf{within 1 second}, which is \textbf{500× faster} than NVF \cite{xue2024neural}, while providing more accurate uncertainty quantification that leads to more effective exploration.

In this section, we first formulate the visibility field within the 3DGS representation (Sec.~\ref{sec:method.formulation}). In particular, we introduce a direction-aware visibility field that can account for visibility change given a specific viewing direction. We then introduce an efficient method to construct and query the visibility field (Sec.~\ref{sec:method.sh}) by representing it with spherical harmonics. Finally, we summarize the overall active mapping pipeline that integrates the visibility field with uncertainty-aware 3DGS rasterization (Sec.~\ref{sec:method.am}).

\subsection{Formulation} \label{sec:method.formulation}

In this subsection, we introduce a novel formulation of \emph{visibility field} in 3DGS. Notably, a straightforward extension of the Neural Visibility Field~\cite{xue2024neural} (as a function of position only) from NeRF to 3DGS (i.e. assigning a scalar visibility value to each Gaussian) is insufficient for reliable uncertainty quantification. Since a 3DGS particle can occlude itself (i.e. self-occlusion), observing a particle from one direction provides little information about its appearance from the opposite side. 
As a simple example, viewing only one side of a wall reveals nothing about the opposite side (Fig.~\ref{fig:ablation_illustration}). A reliable uncertainty quantification method should assign high uncertainty there to drive exploration.

To account for this, a visibility field should depend on the viewing direction. 
We therefore define the \textbf{Anisotropic Visibility Field} $V^{(i)}(\bm{d})$ for 3DGS, 
where for each Gaussian particle $i$, the visibility is formulated as a function of the rendering direction $\bm{d}$, 
with $\bm{d}$ being the unit vector pointing toward the synthesized view.

To obtain $V^{(i)}(\bm{d})$, which is the visibility with respect to the entire training set (views observed so far), we first compute the visibility with respect to a single training view. Let $V^{(i)}_{\bm{p}}(\bm{d})$ denote the visibility from a specific camera pose $\bm{p} \in \mathcal{P}$, where $\mathcal{P} = \{\bm{p}_1, \bm{p}_2, \ldots\}$ is the set of camera poses in the training set. We have

\begin{equation} \label{eq:vis_p}
V^{(i)}_{\mathbf{p}}(\mathbf{d}) =
\underbrace{\Phi_{i,\mathbf{p}}}_{\text{FOV}}
\underbrace{T_{\mathbf{p}}(t_i^{\mathbf{p}})}_{\text{Transmittance}}
\hspace{0.28em}
\underbrace{\nu(\mathbf{d}; \mathbf{d}_p)}_{\text{Directional Visibility}}
\end{equation}
\noindent where
\vspace{-10pt}
\begin{equation*}
 \nu (\bm{d}; \bm{d}_p) := \zeta \exp (\kappa \bm{d}\cdot \bm{d}_p), ~\zeta = \exp(-\kappa) 
\end{equation*}

\noindent This equation consists of three components: 
     
\noindent
\textbf{FOV} $\Phi_{i,\bm{p}} \in \{0, 1\}$ is a binary indicator denoting whether the particle $i$ lies within the field of view of camera $\bm{p}$, where $\Phi_{i,\bm{p}} = 1$ if it is within the field of view and $0$ otherwise. 

\noindent
\textbf{Transmittance} $T_{\bm{p}}(t_i^{\bm{p}})$ denotes 
the probability that the ray travels along $\bm{d}_p$ from $\bm{x_p}$ to particle $i$ without occlusion, where $\bm{d_p}=\frac{\bm{x_i}-\bm{x_p}}{||\bm{x_i}-\bm{x_p}||}$, $\bm{x_p}$ is the position of camera view $\bm{p}$. This term can be directly obtained from the radiance field output. The first two terms $\Phi_{i,\bm{p}} T_{\bm{p}}(t_i^{\bm{p}})$ define isotropic visibility, identical to the formulation of NVF~\cite{xue2024neural}.

\noindent
\textbf{Directional Visibility Function} $\nu(\bm{d}; \bm{d_p})$ captures how visibility $V^{(i)}_{\bm{p}}(\bm{d})$ changes as the rendering direction $\bm{d}$ of a novel view deviates from the training view direction $\bm{d_p}$. Viewing an object from one direction does not imply knowledge of its appearance from the opposite side; the larger the angular difference between $\bm{d}$ and $\bm{d_p}$, the lower the visibility and the higher the uncertainty. We use the spherical function $\zeta \exp(\kappa \bm{d} \cdot \bm{d_p})$, proportional to the widely used von Mises-Fisher distribution, an analogue of the Gaussian distribution for spherical data. Here, $\kappa$ is analogous to the reciprocal of variance in the Gaussian distribution, controlling the concentration around the mean direction, and $\zeta = \exp(-\kappa)$ is a constant factor ensuring $\nu(\bm{d_p}; \bm{d_p}) = 1$ when the rendering direction matches the training direction. 

\begin{figure}
    \centering
    \includegraphics[width=.8\linewidth,trim={14bp 51bp 421bp 2bp},clip]{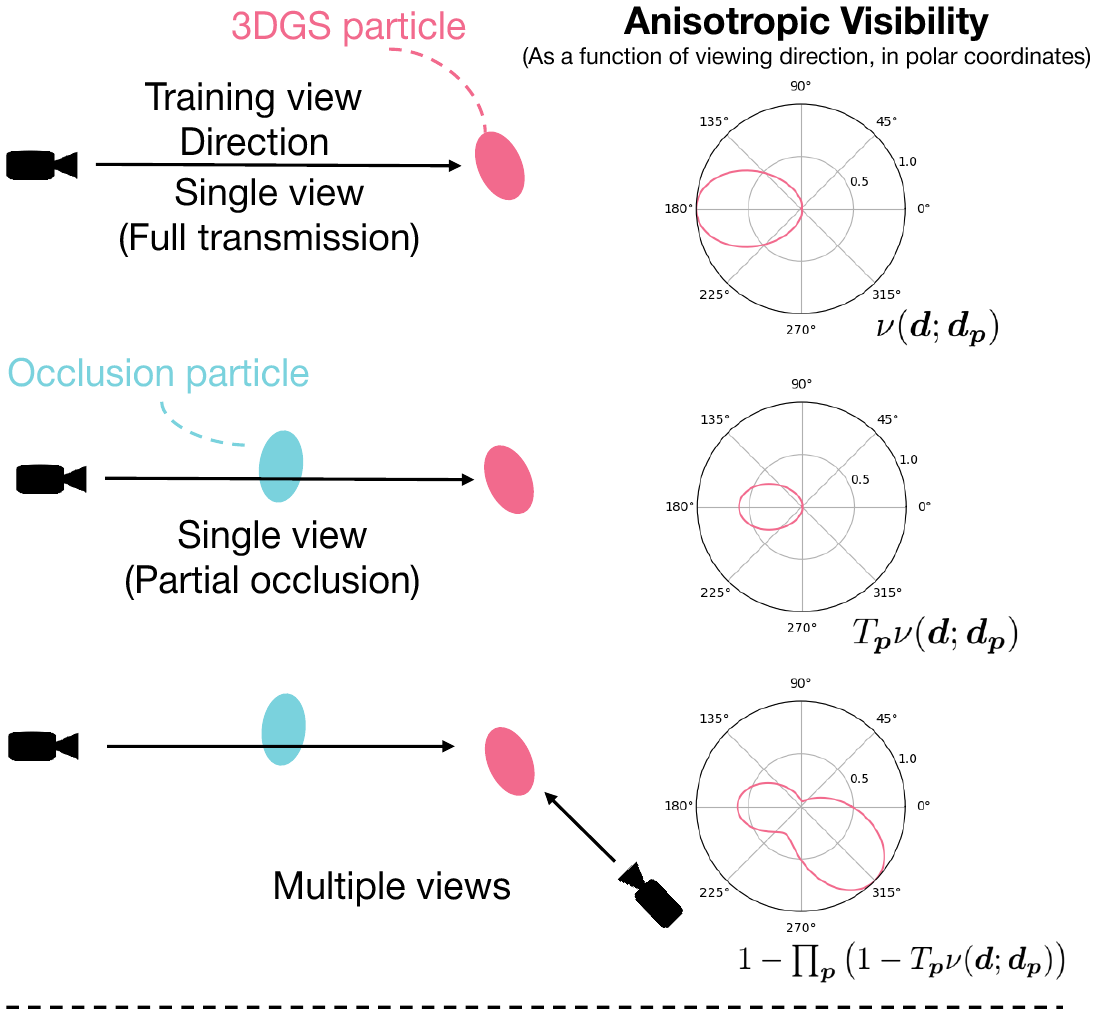}\\[-1pt]
    \includegraphics[width=.8\linewidth,trim={48bp 162bp 243bp 202bp},clip]{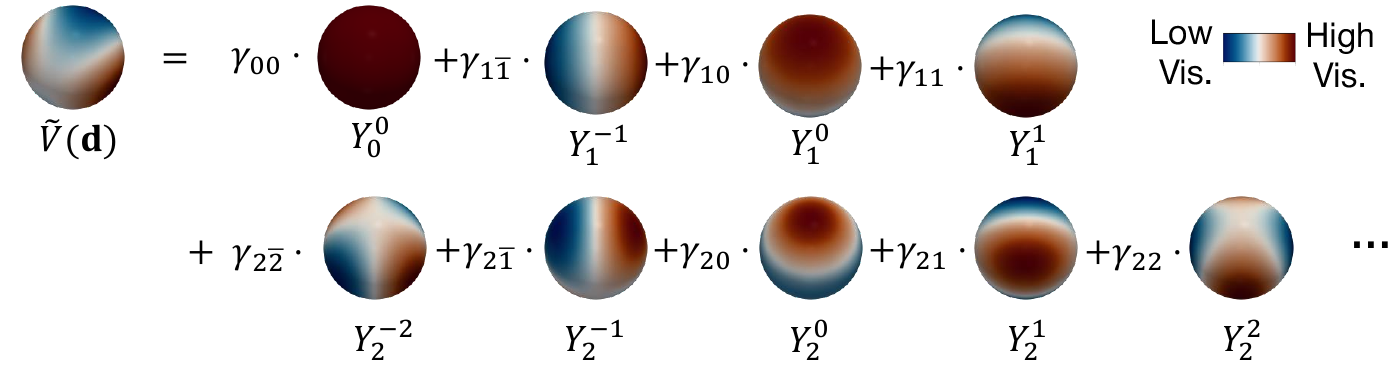}
    \vspace{-8pt}
    \caption{\textbf{Schematic of directional visibility.} (Top) A 2D illustration of how visibility varies with viewing direction. Anisotropic visibility is plotted in polar coordinates for (1) a single training view, (2) a single training view with occlusion, and (3) multiple training views. (Bottom) A 3D illustration of spherical harmonics expansion of $\tilde{V}(\bm{d})$. See Secs.~\ref{sec:method.formulation} and \ref{sec:method.sh} for details.} 
    \label{fig:avf}
    \vspace{-10pt}
\end{figure}

Therefore, the anisotropic visibility field for the entire training set can be computed from the per-training-view anisotropic visibility $V^{(i)}_{\bm{p}}(\bm{d})$ (i.e., the probability of being visible from a single view). Specifically, the probability that particle $i$ is visible to at least one camera in $\mathcal{P}$ is given by
\begin{equation}    \label{eq:vis_all}
V^{(i)}(\bm{d}) = 1 - \prod_{\bm{p} \in \mathcal{P}} (1 - V^{(i)}_{\bm{p}}(\bm{d})).
\end{equation}
\noindent An illustrative example of the anisotropic visibility field under different occlusion conditions is provided in Fig.~\ref{fig:avf}. In the following subsection, we present an accurate and efficient algorithm for visibility field construction and querying, built upon the above formulation.

\subsection{Efficient Construction and Querying of Anisotropic Visibility Field} \label{sec:method.sh}

So far, we have introduced the formulation of the anisotropic visibility field. However, directly applying the analytical expression obtained by combining \eqref{eq:vis_p} and \eqref{eq:vis_all} for uncertainty quantification and active mapping is impractical in real-world scenarios. Querying visibility from a single direction $\bm{d}$ requires accessing all training view directions $\bm{d_p}$, where $p \in \mathcal{P}$, causing both runtime and memory cost to scale with the length of the history trajectory. 
For real-world robotic applications, it is crucial to ensure fast construction and querying of the visibility field.
Following the same philosophy as radiance fields, we seek a representation of the anisotropic visibility field $V^{(i)}(\bm{d})$ that can be stored in constant memory and queried in constant time, independent of the trajectory length. Unlike NVF~\cite{xue2024neural}, which relies on a neural network to approximate the (isotropic) visibility field, we propose an analytical representation of the anisotropic visibility field based on spherical harmonics that is both more accurate and computationally efficient. In this subsection, we present the core components of our method that enable accurate and efficient construction and querying of $V^{(i)}(\bm{d})$ for uncertainty-driven active mapping.

\paragraph{Overview.} Inspired by volumetric radiance field~\cite{muller2022instant, fridovich2022plenoxels}, and 3DGS, where the directional appearance component (color) is represented using spherical harmonics~\cite{kerbl20233d}, we express the \textbf{anisotropic visibility field} in the orthonormal basis of \textbf{spherical harmonics (SH)} $Y_\ell^m(\bm{d})$, as in Fig.~\ref{fig:avf}
\begin{equation} \label{eq:vis_sh}
    \tilde{V}^{(i)}(\bm{d}) = \sum_{\ell=0}^L \sum_{m=-\ell}^\ell \gamma_{\ell m}^{\mathcal{P}}\, Y_\ell^m(\bm{d}),
\end{equation}
where $\tilde{V}^{(i)}(\bm{d})$ is an auxiliary function related to visibility field $V^{(i)}(\bm{d})$, and 
$\gamma_{\ell m}^{\mathcal{P}}$ denotes the spherical harmonics coefficients, $L$ denotes the maximum degree of the spherical harmonics. Next, we describe how auxiliary function $\tilde{V}^{(i)}(\bm{d})$ is used to compute the visibility field $V^{(i)}(\bm{d})$, and show how the coefficients $\gamma_{\ell m}^{\mathcal{P}}$ are computed analytically during visibility-field construction.

\paragraph{SH Representation of $\nu(\bm{d}; \bm{d_p})$.}
We notice that the directional visibility function $\nu(\bm{d}; \bm{d_p})$ can be decomposed into spherical harmonics as %
\begin{equation} \label{eq:SH_vMF}
\nu(\bm{d}; \bm{d_p})
= \zeta \sum_{\ell=0}^\infty \sum_{m=-\ell}^\ell a_{\ell m}\, Y_\ell^m(\bm{d}),
\end{equation}
where the coefficients $a_{\ell m}$ can be computed analytically as
\begin{equation} \label{eq:vmf_coeff}
a_{\ell m}
= 4\pi\, i_\ell(\kappa)\, Y_\ell^{m*}(\bm{d_p}),
\end{equation}
and $Y_\ell^m(\bm{d})$ represents the spherical harmonic basis function, and $Y_\ell^{m*}(\bm{d_p})$ is its complex conjugate,
 $i_\ell(\kappa)$ denotes the modified spherical Bessel function of the first kind~\cite{olver2010nist}.
We include the full derivation in appendix Sec.~\ref{sec:appx_avf_vmf_coeff}. %

\paragraph{SH Representation of $V^{(i)}(\bm{d})$.} Since spherical harmonics form a linear basis, the sum of two functions expressed in spherical harmonics can be obtained by simply adding their corresponding coefficients. 
However, the multiplication of two spherical harmonics~\cite{wigner2012group} is computationally expensive~\cite{luo2024enabling}. 
Directly substituting \eqref{eq:SH_vMF} and \eqref{eq:vis_p} into \eqref{eq:vis_all} to compute the SH coefficients of $V^{(i)}(\bm{d})$ is impractical, 
as it requires a computational complexity of $O(|\mathcal{P}|^4 L^3)$ and a memory cost of $O(|\mathcal{P}|^2 L^2)$ to store the parameters of $V^{(i)}(\bm{d})$ (see Sec.~\ref{sec:appx_avf_sh} for details). 
Consequently, the query-time complexity scales quadratically with the number of training views $|\mathcal{P}|$, which is not suitable for real-time applications. To address this, we propose an efficient method that bypasses the direct multiplication of spherical harmonics, 
achieving computational complexity that is linear in $|\mathcal{P}|$ during construction and constant during query time. Specifically, we can obtain a lower bound of $V^{(i)}(\bm{d})$ by applying the arithmetic-geometric mean (AM-GM) inequality to \eqref{eq:vis_all}:

\begin{equation} \label{eq:vis_bound}
V^{(i)}(\bm{d}) \;\ge\; 1 - \Bigl(1 - \tfrac{\tilde{V}^{(i)}(\bm{d})}{|\mathcal{P}|} \Bigr)^{|\mathcal{P}|}.
\end{equation}
\noindent where 
\begin{equation} \label{eq:vis_tilde}
\tilde{V}^{(i)}(\bm{d}) := \sum_{\bm{p}} V^{(i)}_{\bm{p}}(\bm{d})
\end{equation}
\noindent
By substituting \eqref{eq:vis_p} into \eqref{eq:vis_tilde} and comparing it with \eqref{eq:vis_sh}, 
we can obtain the SH coefficients of $\tilde{V}^{(i)}(\bm{d})$ as
\begin{equation} \label{eq:vis_gamma}
    \gamma_{\ell m}^{\mathcal{P}} 
    = 4\pi\, \zeta\, i_\ell(\kappa)
    \sum_{\bm{p} \in \mathcal{P}} 
    \Phi_{i,\bm{p}}\, T_{\bm{p}}(t_i^{\bm{p}})\, Y_\ell^{m*}(\bm{d_p}).
\end{equation}

\noindent More details are provided in Sec.~\ref{sec:appx_avf_vis_bound}.

\paragraph{Visibility Field Construction.} During the construction stage of the visibility field, given a trained 3DGS, we first use \eqref{eq:vis_gamma} to compute the SH coefficients of $\tilde{V}^{(i)}(\bm{d})$, 
where the single-view particle visibility $\Phi_{i,\bm{p}}\, T_{\bm{p}}(t_i^{\bm{p}})$ is efficiently computed using a modified 3DGS rasterizer 
(see appendix Alg.~\ref{alg:pos_vis} for details). 
For each particle, the set of coefficients $\{\gamma_{\ell m}^{\mathcal{P}}\}$ is stored as the parameters of the visibility field, 
requiring $(L+1)^2$ parameters per particle. 
In practice, we find that $L=2$ is sufficient to accurately capture visibility anisotropy. This SH degree is also commonly used in 3DGS to represent view-dependent color~\cite{kerbl20233d}.

\paragraph{Constant-time Visibility Field Query.} 
So far, we have introduced an efficient algorithm that constructs the visibility field using spherical harmonics and stores its parameters in the coefficients $\{\gamma_{\ell m}^{\mathcal{P}}\}$.
When querying the visibility field in a given direction $\bm{d}$, we first compute $\tilde{V}^{(i)}(\bm{d})$ from \eqref{eq:vis_sh} 
using the visibility field parameters $\{\gamma_{\ell m}^{\mathcal{P}}\}$. 
We then apply the lower bound in \eqref{eq:vis_bound} to estimate the anisotropic visibility (see Alg.~\ref{alg:visibility_query} for details). 
Importantly, this formulation enables \textbf{constant-time querying of the visibility field}.

\begin{algorithm}[t]
\caption{Query Anisotropic Visibility Field}
\label{alg:visibility_query}
\begin{algorithmic}[1]
\Require SH coefficients $\{\gamma_{\ell m}^{(g)}\}$ for particle $g$; query direction $\bm{d}$; SH degree $L$; number of training views $|\mathcal{P}|$
\State $\tilde{V} \gets \sum_{\ell=0}^{L}\sum_{m=-\ell}^{\ell} \gamma_{\ell m}^{(g)}\,Y_\ell^{m}(\bm{d})$ 
\State ${V}^{(g)}(\bm{d}) \gets 1- (1-\frac{\tilde{V}}{|\mathcal{P}|})^{|\mathcal{P}|}$ \Comment{AM--GM lower-bound estimator}
\State \textbf{return} ${V}^{(g)}(\bm{d})$
\end{algorithmic}
\end{algorithm}

\paragraph{Visibility Field Density Control.} A core component of 3DGS is its adaptive density control~\cite{kerbl20233d}: particles are densified in regions of interest and pruned in free space, yielding high-quality constructions without substantial memory or runtime overhead. However, this mechanism complicates uncertainty estimation. Empty regions may correspond either to true free space (pruned) or to underexplored areas with too few initialized particles (e.g., from sparse SfM/depth initialization). Because particles in unseen areas contribute no loss, they receive no gradient and thus are not densified, making both cases appear similarly empty. As a result, particle-centric uncertainty measures often overlook such regions and falsely assign uniformly low uncertainty to both, 
even though distinguishing them is crucial for active mapping, 
where the uncertainty quantification method should assign high uncertainty to underexplored areas.

To tackle this challenge, we propose using \emph{virtual particles} to distinguish free space from underexplored regions. 
Given a trained 3DGS, we uniformly sample particles across the scene with zero opacity and compute their visibility as 
$1 - \prod_{\bm{p} \in \mathcal{P}} (1 - \Phi_{i,\bm{p}}\, T_{\bm{p}}(t_i^{\bm{p}}))$ 
with respect to all training views. 
Low visibility indicates that a particle lies in an underexplored (unseen) region, 
while high visibility suggests it resides in free space. 
Therefore, we prune virtual particles in free space using a visibility threshold 
and concatenate the remaining particles with the trained 3DGS for uncertainty quantification, 
where the visibility of virtual particles is set to $V^{(i)}(\bm{d}) = 0$ for all directions.
We find that using a relatively small number of virtual particles (approximately 5--10\% of the total particles)
is sufficient to distinguish underexplored regions from free space. 
Further implementation details are provided in appendix Sec.~\ref{sec:appx_avf_density}.

\vspace{-2.0px}
\subsection{Active Mapping Pipeline} \label{sec:method.am}

So far, we have introduced our core contribution: an accurate and efficient method to compute the 3DGS visibility field $V^{(i)}(\bm{d})$. 
We then discuss how to apply it to uncertainty quantification and active mapping. 

Our approach extends the NeRF-based uncertainty-aware volume rendering pipeline in NVF \cite{xue2024neural} to 3DGS rasterization, 
while providing significant improvements in visibility estimation and computational efficiency.
Given the observed images and corresponding camera poses, a 3DGS is first trained. 
We then construct the visibility field by computing the visibility field parameters $\{\gamma_{\ell m}^{\mathcal{P}}\}$ using \eqref{eq:vis_gamma}, 
and integrating virtual particles that identify the underexplored regions. 
Once the visibility field is constructed, uncertainty quantification is performed by evaluating the entropy of synthesized views within a candidate view set sampled from a prior distribution (see Sec.~\ref{sec:exp_setup}). 
For each candidate view, uncertainty-aware 3DGS rasterization (see Alg.~\ref{alg:uq}) is applied following \eqref{eq:gmm}, 
where $v_i$ is obtained by querying the visibility field $V^{(i)}(\bm{d})$ along ray direction $\bm{d}$ using \eqref{eq:vis_bound}. 
Finally, the next best view is selected as the one that maximizes entropy, as defined in \eqref{eq:i_gain}. 
An overview of this pipeline is illustrated in Fig.~\ref{fig:pipeline}, and further details are provided in appendix Sec.~\ref{sec:appx_uvr}~\&~\ref{sec:appx_am}.

\vspace{-1.0px}

\section{Experiments} \label{sec:exp}

\renewcommand\meanstd[2]{%
  #1%
}

\begin{table*}[t]
  \centering
  \vspace{0pt}
  \resizebox{0.85\linewidth}{!}{%
    \begin{tabular}{cccccccccccccccc}
    \toprule
    \multirow{2}[4]{*}{Method} & \multicolumn{7}{c}{NeRF Synthetic}                    &       & \multicolumn{7}{c}{Space} \\
\cmidrule{2-8}\cmidrule{10-16}          & PSNR $\uparrow$ & SSIM $\uparrow$ & LPIPS $\downarrow$ & CR $\uparrow$ & VIS $\uparrow$ & UQ FPS $\uparrow$ & $T_{UP} \downarrow$ &       & PSNR $\uparrow$ & SSIM $\uparrow$ & LPIPS $\downarrow$ & CR $\uparrow$ & VIS $\uparrow$ & UQ FPS $\uparrow$ & $T_{UP} \downarrow$ \\
    \midrule
    FisherRF & \meanstd{22.34}{0.31} & \meanstd{0.870}{0.004} & \meanstd{0.119}{0.003} & \meanstd{0.626}{0.011} & \meanstd{0.376}{0.010} & \besttwo{\meannostd{146.3}{1.9}} & \besttwo{\meannostd{0.42}{0.00}} &       & \meanstd{24.17}{0.08} & \meanstd{0.834}{0.004} & \meanstd{0.158}{0.002} & \meanstd{0.547}{0.017} & \meanstd{0.474}{0.016} & \besttwo{\meannostd{141.5}{2.3}} & \besttwo{\meannostd{0.37}{0.01}} \\
    VIMC  & \besttwo{\meanstd{23.14}{0.25}} & \besttwo{\meanstd{0.880}{0.003}} & \besttwo{\meanstd{0.107}{0.003}} & \besttwo{\meanstd{0.651}{0.010}} & \meanstd{0.407}{0.011} & \meannostd{144.5}{0.4} & \meannostd{9.48}{0.15} &       & \besttwo{\meanstd{24.56}{0.55}} & \besttwo{\meanstd{0.841}{0.004}} & \besttwo{\meanstd{0.150}{0.002}} & \besttwo{\meanstd{0.612}{0.008}} & \meanstd{0.510}{0.009} & \meannostd{127.6}{0.6} & \meannostd{17.75}{0.68} \\
    NVF   & \meanstd{22.59}{0.26} & \meanstd{0.859}{0.005} & \meanstd{0.147}{0.004} & \meanstd{0.549}{0.014} & \besttwo{\meanstd{0.431}{0.005}} & \meannostd{11.9}{0.0} & \meannostd{149.13}{0.46} &       & \meanstd{23.76}{0.46} & \meanstd{0.796}{0.012} & \meanstd{0.202}{0.010} & \meanstd{0.499}{0.019} & \besttwo{\meanstd{0.564}{0.017}} & \meannostd{10.7}{0.0} & \meannostd{140.54}{0.92} \\
    GAVIS (ours) & \bestone{\meanstd{24.26}{0.25}} & \bestone{\meanstd{0.894}{0.002}} & \bestone{\meanstd{0.097}{0.002}} & \bestone{\meanstd{0.711}{0.009}} & \bestone{\meanstd{0.437}{0.006}} & \bestone{\meannostd{251.5}{1.0}} & \bestone{\meannostd{0.17}{0.00}} &       & \bestone{\meanstd{26.14}{0.10}} & \bestone{\meanstd{0.857}{0.003}} & \bestone{\meanstd{0.140}{0.002}} & \bestone{\meanstd{0.630}{0.019}} & \bestone{\meanstd{0.582}{0.017}} & \bestone{\meannostd{232.4}{1.3}} & \bestone{\meannostd{0.17}{0.00}} \\
    \midrule
    \multirow{2}[4]{*}{Method} & \multicolumn{7}{c}{Gibson}                            &       & \multicolumn{7}{c}{HM3D} \\
\cmidrule{2-8}\cmidrule{10-16}          & PSNR $\uparrow$ & SSIM $\uparrow$ & LPIPS $\downarrow$ & CR $\uparrow$ & VIS $\uparrow$ & UQ FPS $\uparrow$ & $T_{UP} \downarrow$ &       & PSNR $\uparrow$ & SSIM $\uparrow$ & LPIPS $\downarrow$ & CR $\uparrow$ & VIS $\uparrow$ & UQ FPS $\uparrow$ & $T_{UP} \downarrow$ \\
    \midrule
    FisherRF & \meanstd{18.11}{0.50} & \meanstd{0.720}{0.007} & \meanstd{0.419}{0.006} & \meanstd{0.431}{0.031} & \meanstd{0.469}{0.035} & \meannostd{39.8}{0.2} & \besttwo{\meannostd{0.90}{0.01}} &       & \meanstd{18.32}{0.44} & \meanstd{0.693}{0.009} & \meanstd{0.446}{0.009} & \meanstd{0.447}{0.025} & \meanstd{0.558}{0.030} & \meannostd{37.7}{0.2} & \besttwo{\meannostd{1.59}{0.01}} \\
    VIMC  & \meanstd{15.70}{0.20} & \meanstd{0.668}{0.003} & \meanstd{0.465}{0.003} & \meanstd{0.337}{0.013} & \meanstd{0.366}{0.014} & \besttwo{\meannostd{57.0}{1.1}} & \meannostd{90.47}{2.34} &       & \meanstd{17.15}{0.26} & \meanstd{0.645}{0.006} & \meanstd{0.477}{0.006} & \meanstd{0.476}{0.017} & \meanstd{0.618}{0.020} & \besttwo{\meannostd{50.1}{0.3}} & \meannostd{114.51}{1.95} \\
    NVF   & \besttwo{\meanstd{23.29}{0.12}} & \besttwo{\meanstd{0.798}{0.001}} & \besttwo{\meanstd{0.402}{0.002}} & \bestone{\meanstd{0.880}{0.002}} & \bestone{\meanstd{0.915}{0.002}} & \meannostd{4.2}{0.0} & \meannostd{219.87}{0.77} &       & \besttwo{\meanstd{22.69}{0.14}} & \besttwo{\meanstd{0.760}{0.002}} & \besttwo{\meanstd{0.434}{0.002}} & \besttwo{\meanstd{0.819}{0.003}} & \besttwo{\meanstd{0.873}{0.001}} & \meannostd{4.2}{0.0} & \meannostd{285.26}{0.26} \\
    GAVIS (ours) & \bestone{\meanstd{24.42}{0.14}} & \bestone{\meanstd{0.812}{0.003}} & \bestone{\meanstd{0.323}{0.004}} & \besttwo{\meanstd{0.831}{0.005}} & \besttwo{\meanstd{0.890}{0.003}} & \bestone{\meannostd{207.3}{0.7}} & \bestone{\meannostd{0.42}{0.00}} &       & \bestone{\meanstd{23.97}{0.07}} & \bestone{\meanstd{0.791}{0.001}} & \bestone{\meanstd{0.338}{0.002}} & \bestone{\meanstd{0.820}{0.003}} & \bestone{\meanstd{0.876}{0.001}} & \bestone{\meannostd{192.6}{0.7}} & \bestone{\meannostd{0.79}{0.01}} \\
    \bottomrule
    \end{tabular}%
  }
  \vspace{-9pt}
  \caption{\textbf{Quantitative results.} Active mapping performance on all datasets across all baselines and our method. Best results are in \textbf{bold}; second-best are \underline{underlined}. Here, $T_{UP}$ denotes the uncertainty preparation time, and UQ FPS is the frame rate (FPS) for quantifying uncertainty for each candidate view. See Sec.~\ref{sec:exp} for details.
   \vspace{-1pt}
}
\label{tab:am_blender}
  \vspace{-13pt}
\end{table*}

In this section, we design experiments to address the following key questions:  
(\textbf{Q1}) Can GAVIS accurately quantify uncertainty to enable effective active mapping?  
(\textbf{Q2}) Can GAVIS operate efficiently in practice?  
(\textbf{Q3}) How effective is each component of GAVIS? 
(\textbf{Q4}) Can GAVIS be applied as a post-hoc method to improve the performance of existing uncertainty quantification approaches?  

\subsection{Experiment Setups} \label{sec:exp_setup}
\paragraph{Environments.}
To demonstrate the advantages of our method across diverse scenes and robotic applications, we conduct experiments on three types of scenarios spanning 3 domains and 4 datasets:
(1) the standard NeRF Synthetic dataset~\cite{mildenhall2021nerf} for object reconstruction;
(2) a space dataset \cite{xue2024neural} consisting of the Hubble Space Telescope (HST)
and the International Space Station (ISS)
for space robotics; and
(3) indoor scenes, including 8 environments from the Habitat-Matterport 3D (HM3D)~\cite{ramakrishnan2021hm3d} and 8 from Gibson~\cite{xia2018gibson} datasets, for household robotics. 
We set the planning steps to 10 (NeRF-Synthetic), 40 (Gibson), and 80 (HM3D), determined as the minimal steps at which the strongest methods achieve reasonable reconstruction quality. Further details are provided in appendix Sec.~\ref{sec:appx_exp:setup}.

\paragraph{Baselines.}
We evaluate GAVIS against state-of-the-art (SOTA) uncertainty quantification approaches for radiance fields in active mapping, 
including 3DGS-based FisherRF~\cite{jiang2023fisherrf},
VIMC~\cite{lyu2024manifold},
and NeRF-based NVF~\cite{xue2024neural}. %
To ensure a fair comparison of uncertainty quantification methods, we use the same active mapping pipeline for all methods by following the setup in NVF~\cite{xue2024neural}. 
All methods use the same view sampler that uniformly draws collision-free poses from the scene without any additional heuristics to fairly evaluate the performance of the uncertainty quantification method.
Further details on baseline and training setup are provided in Sec.~\ref{sec:appx_exp:train}.

\paragraph{Metrics.} 
For each method, we evaluate performance after mapping is complete using PSNR, SSIM, and LPIPS~\cite{zhang2018perceptual}, as well as mesh-based metrics including completion ratio (CR)~\cite{sucar2021imap} and scene visual coverage (VIS)~\cite{xue2024neural}, to assess the effectiveness of active mapping.
We additionally report mesh metrics, including completion (Comp) and accuracy (Acc), in the appendix Sec.~\ref{sec:appx_quant}.
However, we note that Acc is not well aligned with the objective of active mapping under 3DGS, further discussion is provided in Appendix Sec.~\ref{sec:appx_exp:metric}.
To evaluate the computational efficiency of each method, we measure the runtime during active mapping.
Since planning time depends on the number of candidate views used for uncertainty quantification, 
we decompose the runtime into two components:  
(a) \textbf{Uncertainty Preparation Time ($T_{UP}$)}, the additional time required after training the radiance field, independent of the number of candidate views. 
For GAVIS and NVF, this corresponds to visibility field construction; 
for FisherRF, to estimate the Hessian matrix of model parameters;
and for VIMC, to the extra time compared to standard 3DGS training.  
(b) \textbf{Uncertainty Quantification Time (UQ FPS)}, the frame rate (FPS) for evaluating uncertainty for each candidate view. 
We also report the Area Under the Sparsification Error curve (AUSE)~\cite{ilg2018uncertainty,poggi2020uncertainty} for uncertainty quality. Noting that the standard depth-based variant (AUSE-D)~\cite{jiang2023fisherrf,lyu2024manifold,goli2023bayes} can be misaligned with active mapping goals, we additionally introduce a visibility-based variant (AUSE-V) that better correlates with active mapping performance. Details and analysis are provided in Sec.~\ref{sec:appx_exp:metric}.

\begin{table}[htbp]
  \centering
\vspace{-6pt}
  \resizebox{0.72\linewidth}{!}{
    \begin{tabular}{cccccc}
    \toprule
          & PSNR $\uparrow$ & SSIM $\uparrow$ & LPIPS $\downarrow$ & CR $\uparrow$ & VIS $\uparrow$ \\
    \midrule
    GAVIS & \meanstd{24.70}{0.08} & \meanstd{0.839}{0.001} & \meanstd{0.224}{0.001} & \multicolumn{1}{l}{\meanstd{0.748}{0.006}} & \meanstd{0.697}{0.005} \\
    Isotropic & \meanstd{23.97}{0.08} & \meanstd{0.827}{0.001} & \meanstd{0.231}{0.001} & \multicolumn{1}{l}{\meanstd{0.741}{0.004}} & \meanstd{0.671}{0.006} \\
    w/o DC & \meanstd{24.18}{0.08} & \meanstd{0.830}{0.001} & \meanstd{0.234}{0.001} & \meanstd{0.712}{0.006} & \meanstd{0.668}{0.007} \\
    Iso. w/o DC & \meanstd{23.38}{0.15} & \meanstd{0.819}{0.003} & \meanstd{0.240}{0.003} & \meanstd{0.691}{0.012} & \meanstd{0.625}{0.011} \\
    \bottomrule
    \end{tabular}%
    }
    \vspace{-10pt}
      \caption{\textbf{Ablation study.} We evaluate GAVIS for active mapping by isolating the effects of (i) anisotropic visibility ($\nu(\bm{d}; \bm{d_p})=1$) and (ii) the Visibility-Field Density Control module.
} \label{tab:ablation}
\end{table}%

\begin{table}[htbp]
  \centering
   \vspace{-19pt}
  \resizebox{0.72\linewidth}{!}{
    \begin{tabular}{cccccc}
    \toprule
          & PSNR $\uparrow$ & SSIM $\uparrow$ & LPIPS $\downarrow$ & CR $\uparrow$ & VIS $\uparrow$ \\
    \midrule
    Fisher & \meanstd{20.73}{0.18} & \meanstd{0.779}{0.003} & \meanstd{0.285}{0.003} & \meanstd{0.513}{0.011} & \meanstd{0.469}{0.012} \\
    VIMC  & \meanstd{20.14}{0.17} & \meanstd{0.758}{0.002} & \meanstd{0.300}{0.002} & \meanstd{0.519}{0.006} & \meanstd{0.475}{0.007} \\
    Fisher+GAVIS & \meanstd{24.70}{0.10} & \meanstd{0.842}{0.001} & \meanstd{0.220}{0.001} & \meanstd{0.748}{0.004} & \meanstd{0.699}{0.006} \\
    VIMC+GAVIS & \meanstd{24.21}{0.11} & \meanstd{0.833}{0.001} & \meanstd{0.227}{0.001} & \meanstd{0.719}{0.004} & \meanstd{0.672}{0.004} \\
    \bottomrule
    \end{tabular}%
}
  \vspace{-10pt}
  \caption{\textbf{GAVIS as post-hoc module.} Active mapping results showing that applying GAVIS post hoc improves the performance of baseline 3DGS uncertainty-quantification methods.
} \label{tab:posthoc}
\end{table}%

\begin{table}[htbp]
  \centering
   \vspace{-19pt}
  \resizebox{0.72\linewidth}{!}{
    \begin{tabular}{ccccc}
    \toprule
          & GAVIS (Ours) & NVF   & FisherRF & VIMC \\
    \midrule
    AUSE-D $\downarrow$ & \bestone{0.224} & \besttwo{0.381} & 0.463 & 0.504 \\
    AUSE-V $\downarrow$ & \bestone{0.176} & \besttwo{0.231} & 0.496 & 0.447 \\
    \bottomrule
    \end{tabular}%
}
  \vspace{-10pt}
\caption{\textbf{Uncertainty quantification.} Quality of uncertainty is evaluated using the depth-based AUSE (AUSE-D) and visibility-based variant (AUSE-V). See Sec.~\ref{sec:exp} for details. \vspace{-14pt}
} \label{tab:uq}
\end{table}%

\subsection{Core Results}

\begin{figure}
    \centering
    \includegraphics[width=\linewidth]{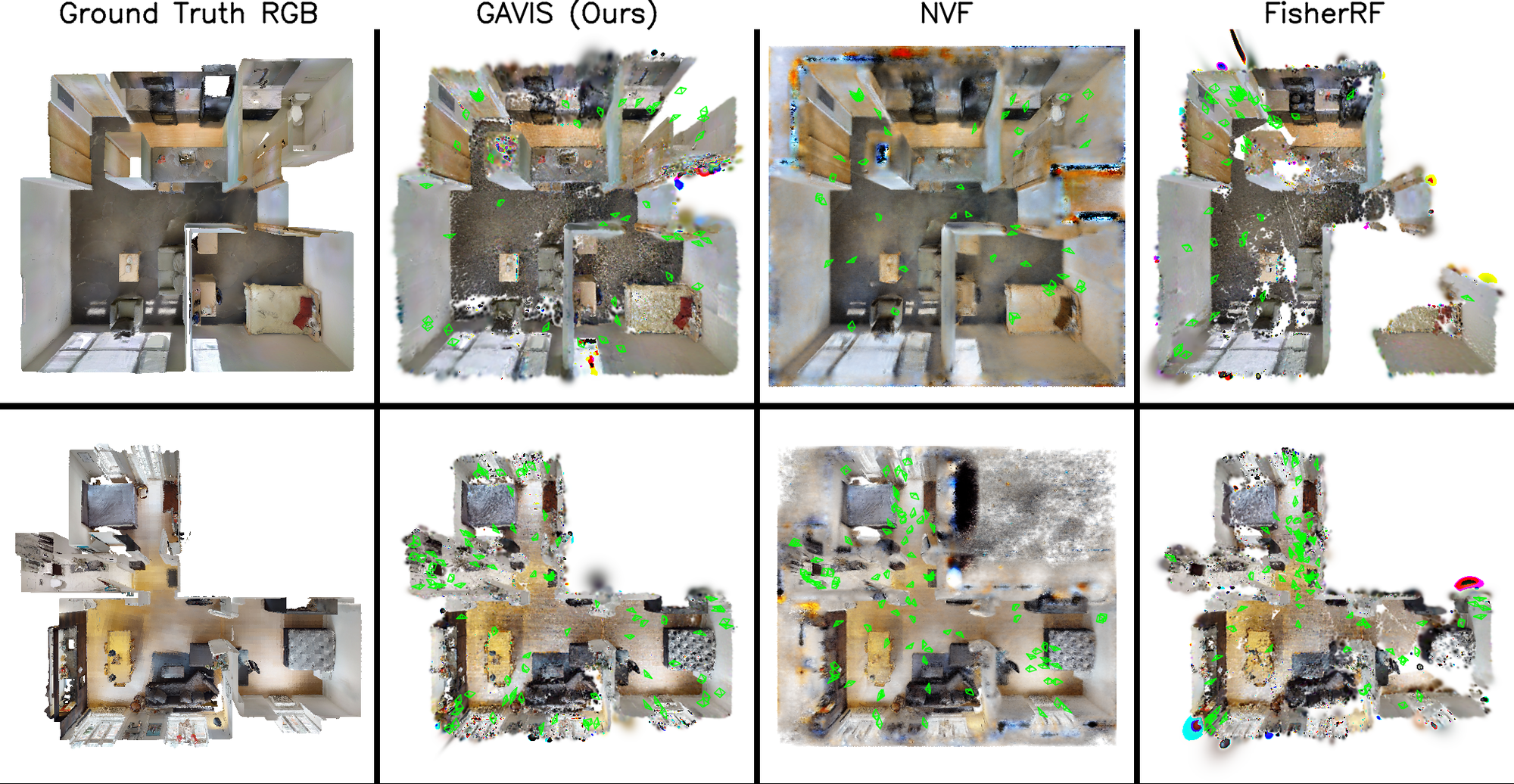}
    \vspace{-21pt}
    \caption{\textbf{Qualitative active mapping.} Reconstruction results and camera-view distributions (green frustums) from different methods’ active-mapping trajectories on Gibson scene (top) and HM3D scene (bottom). Full results are provided in Sec.~\ref{sec:appx_result}.}
    \label{fig:am_room}
    \vspace{-10pt}
\end{figure}

\begin{figure}
    \centering
    \includegraphics[width=\linewidth]{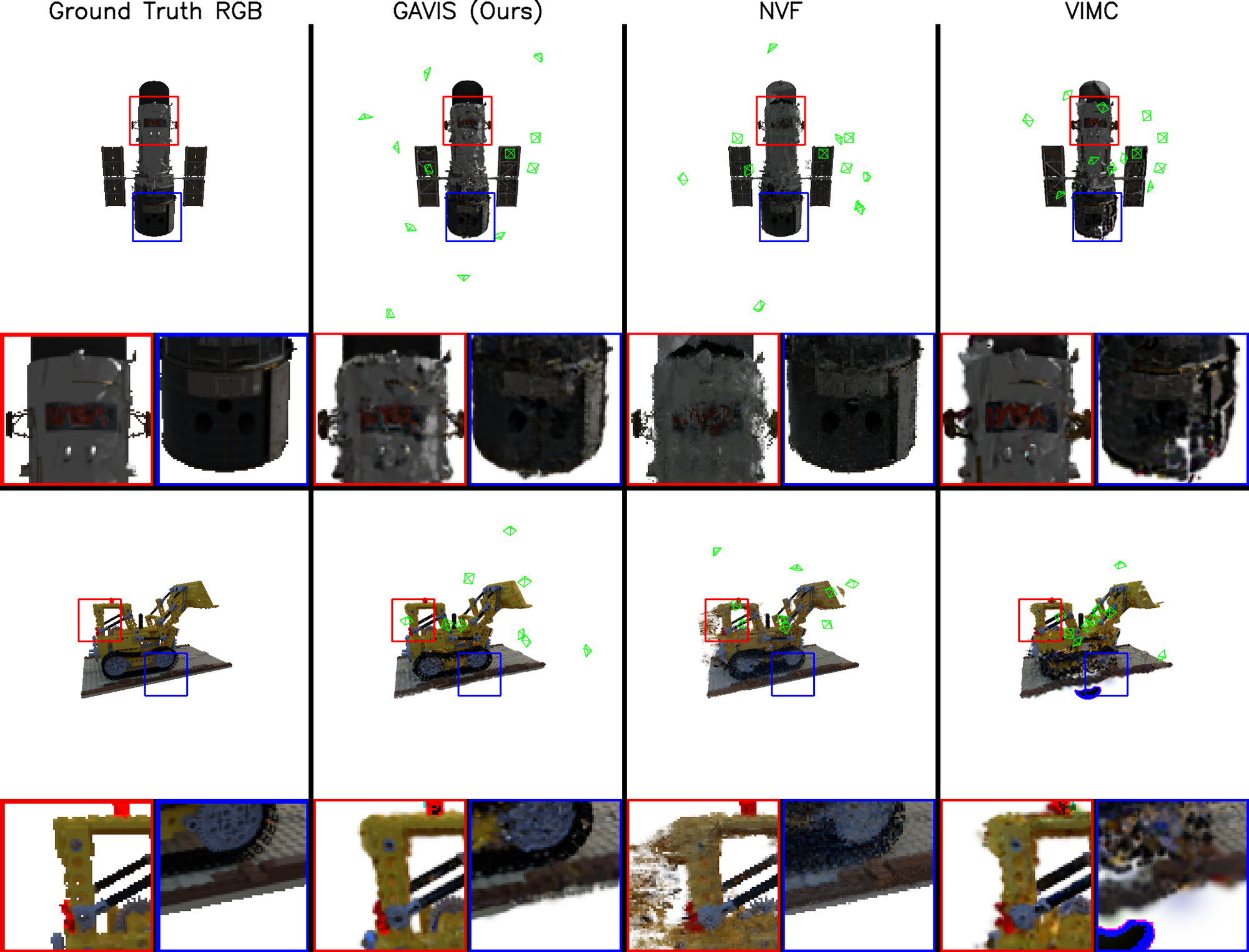}
    \vspace{-21pt}
    \caption{\textbf{Qualitative active mapping.} Reconstruction results and camera-view distributions (green frustums) from different methods’ active-mapping trajectories on HST scene (top) and Lego scene (bottom). Full results are provided in Sec.~\ref{sec:appx_result}.}
    \label{fig:am_blender}
    \vspace{-12pt}
\end{figure}

\begin{figure}
    \centering
    \includegraphics[width=\linewidth]{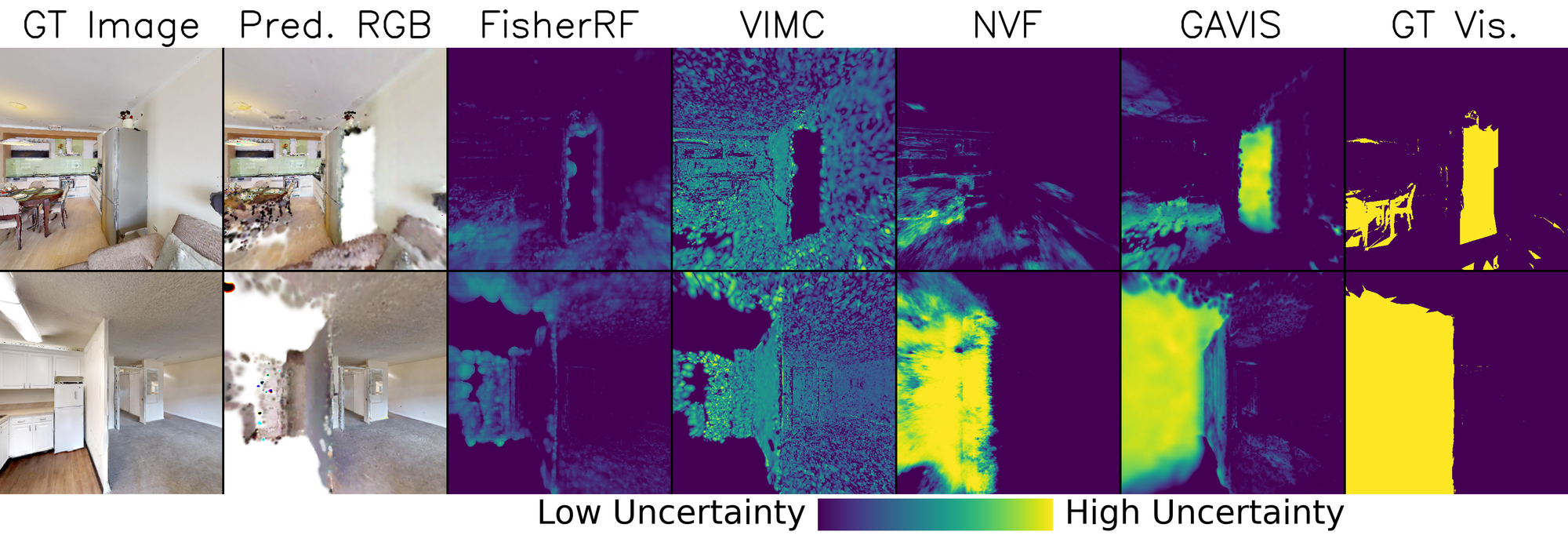}
    \vspace{-21pt}
    \caption{\textbf{Qualitative uncertainty estimation.} All methods are trained on the same set of views that only partially cover the scene, leaving some regions underexplored. GT Vis. indicates rasterized ground-truth mesh visibility (binary face labels), where brighter denotes higher uncertainty (invisible faces) and darker denotes lower uncertainty (visible faces). Our method accurately assigns high uncertainty to invisible regions, aligning with GT Vis.}
    \label{fig:en_room}
    \vspace{-16pt}
\end{figure}

\paragraph{GAVIS accurately quantifies uncertainty and enables effective active mapping.}
As shown in the quantitative results (Tab.~\ref{tab:am_blender}) and qualitative results (Fig.~\ref{fig:am_blender}, and Fig.~\ref{fig:am_room}), 
our method consistently outperforms all 3DGS-based baselines (FisherRF and VIMC) across all datasets and evaluation metrics. 
The improvement is particularly pronounced in challenging indoor scenes such as Gibson and HM3D, 
where GAVIS significantly surpasses FisherRF and VIMC. 
This is because these methods do not model visibility, 
highlighting that visibility modeling is crucial for active mapping in complex scenes, which is consistent with the findings in NVF~\cite{xue2024neural}. GAVIS outperforms NVF across all image-based metrics. %
This highlights the importance of analytical anisotropic visibility modeling for high-quality novel view synthesis. 
For visibility and mesh metrics, the gains are less pronounced because these metrics are inherently isotropic and direction-agnostic: a mesh face is marked visible once it is observed from any single direction, and a single depth observation may be sufficient to reconstruct the local geometry. Consequently, the benefits of anisotropic visibility, which encourages revisiting the same region from diverse directions, are not fully captured. 
To further demonstrate that GAVIS accurately quantifies uncertainty and enables effective exploration, particularly by assigning higher uncertainty to unseen regions, we present qualitative results of the quantified uncertainty maps. We train all methods with the same set of views covering only part of a room, leaving remaining areas underexplored to mimic a robot exploration process. 
As shown in Fig.~\ref{fig:en_room}, GAVIS successfully identifies unseen regions, such as areas behind doors or walls, consistent with the ground-truth visibility derived from the mesh (brighter indicates unseen during training, darker indicates seen). 
In contrast, other 3DGS-based baselines fail to detect these regions, assigning falsely low uncertainty and thus reducing their incentive to explore them. 
While NVF captures some underexplored regions (e.g., behind walls), its neural network-based approximation struggles to accurately detect smaller unseen areas (e.g., behind doors). 
Moreover, we emphasize that a reliable UQ method should assign high uncertainty to regions with low GT visibility; however, high uncertainty does not necessarily imply low GT visibility (e.g., when the test view direction significantly deviates from training views direction, see Sec.\ref{sec:appx_exp:metric} for details).
We additionally evaluate uncertainty quantification on 5 representative scenes across all datasets using AUSE-D and AUSE-V, as shown in Tab.~\ref{tab:uq}. GAVIS achieves the best results on both metrics, demonstrating the superior quality of its uncertainty estimates.

\paragraph{GAVIS is significantly more computationally efficient than existing methods.} We conduct extensive runtime evaluations to demonstrate the speed advantage of GAVIS over existing baselines. 
To ensure fair comparison, all methods are trained with the same views obtained from a representative trajectory that fully explores each scene, 
and runtime is measured at five evenly spaced steps along the trajectory, mimicking different levels of scene exploration. Experiments were run on a single NVIDIA A40 GPU. As shown in Tab.~\ref{tab:am_blender}, our method significantly outperforms all baselines in both uncertainty preparation and quantification time. 
Compared to our closest counterpart, NVF, which is the strongest baseline in active mapping performance, GAVIS achieves about \textbf{500$\times$} speedup in visibility field construction, 
thanks to our efficient spherical-harmonics-based construction and fast visibility query algorithm. 
It is also \textbf{30$\times$ faster} in uncertainty quantification. 
GAVIS further surpasses other 3DGS-based baselines: FisherRF requires gradient backpropagation through the 3DGS rasterizer, leading to high computational cost for uncertainty preparation and quantification; 
while VIMC must co-train a stochastic radiance field and, during inference, perform multiple rasterizations from each sample of the field parameters, resulting in substantially higher runtime.

\paragraph{Ablation studies show that each major component of GAVIS plays a crucial role.}
We ablate two key components of our method: (1) anisotropic visibility and (2) visibility field density control. 
Results averaged over all datasets are reported in Tab.~\ref{tab:ablation}. 
Removing either component causes a noticeable drop across all metrics, 
while jointly removing both reduces the model to a direct 3DGS extension of NVF, yielding the worst performance.
We include additional qualitative results in the appendix Fig.~\ref{fig:ablation_illustration} to further illustrate the importance of each component.

\paragraph{GAVIS can be applied post-hoc to improve the performance of existing active mapping methods.} We integrate GAVIS into the existing baselines FisherRF and VIMC as a post-hoc module to refine their field uncertainty estimation, assigning higher uncertainty to low-visibility regions. 
Details of the derivation and implementation are provided in appendix Sec.~\ref{sec:appx_posthoc}. 
Results averaged over all datasets are reported in Tab.~\ref{tab:posthoc}. GAVIS significantly improves the performance of both baseline methods, 
highlighting the crucial role of visibility in active mapping. 
Notably, FisherRF+GAVIS only marginally outperforms GAVIS alone, consistent with the observation in~\cite{xue2024neural} 
that visibility modeling is the dominant factor in effective active mapping.
Meanwhile, 
VIMC+GAVIS still underperforms GAVIS, as VIMC relies on sampling-based uncertainty estimation, 
where the noise introduced by sampling overshadows the benefit of visibility modeling.

\section{Conclusion}
\label{sec:conclusion}

In this work, we present Gaussian Splatting Anisotropic Visibility Field (GAVIS), a principled approach to quantifying uncertainty in 3D Gaussian Splatting through accurate visibility modeling. We propose an analytical method for computing the anisotropic visibility field in 3DGS using spherical harmonics, representing a substantial improvement over prior work that relies on training neural networks, and achieving significant gains in both efficiency and accuracy. Empirically, GAVIS consistently outperforms all baselines across diverse evaluation scenes. The proposed anisotropic visibility field is not limited to active mapping and can potentially be extended to other tasks that require visibility estimation in 3DGS.

\section*{Acknowledgments}
This work is supported by NSF grant 2101250.
We thank the anonymous reviewers for their valuable suggestions.

{
    \small
    \bibliographystyle{ieeenat_fullname}
    \bibliography{main}

\begin{thebibliography}{114}
\providecommand{\natexlab}[1]{#1}
\providecommand{\url}[1]{\texttt{#1}}
\expandafter\ifx\csname urlstyle\endcsname\relax
  \providecommand{\doi}[1]{doi: #1}\else
  \providecommand{\doi}{doi: \begingroup \urlstyle{rm}\Url}\fi

\bibitem[Abramowitz and Stegun(1948)]{abramowitz1948handbook}
Milton Abramowitz and Irene~A Stegun.
\newblock \emph{Handbook of mathematical functions with formulas, graphs, and
  mathematical tables}.
\newblock US Government printing office, 1948.

\bibitem[AG~Stuart et~al.(2025)AG~Stuart, Morton, Stavness, and
  Pound]{ag20253dgs}
Lewis AG~Stuart, Andrew Morton, Ian Stavness, and Michael~P Pound.
\newblock 3dgs-to-pc: 3d gaussian splatting to dense point clouds.
\newblock In \emph{Proceedings of the IEEE/CVF International Conference on
  Computer Vision}, pages 3730--3739, 2025.

\bibitem[Applications and (VTAD)(2019{\natexlab{a}})]{nasa_hubble_3d_model}
{NASA} Visualization~Technology Applications and Development (VTAD).
\newblock Hubble space telescope 3d model.
\newblock
  \url{https://science.nasa.gov/resource/hubble-space-telescope-3d-model-2/},
  2019{\natexlab{a}}.

\bibitem[Applications and (VTAD)(2019{\natexlab{b}})]{nasa_iss_3d_model}
{NASA} Visualization~Technology Applications and Development (VTAD).
\newblock International space station 3d model.
\newblock
  \url{https://science.nasa.gov/resource/international-space-station-3d-model/},
  2019{\natexlab{b}}.

\bibitem[Atanasov et~al.(2015)Atanasov, Le~Ny, Daniilidis, and
  Pappas]{atanasov2015decentralized}
Nikolay Atanasov, Jerome Le~Ny, Kostas Daniilidis, and George~J Pappas.
\newblock Decentralized active information acquisition: Theory and application
  to multi-robot slam.
\newblock In \emph{2015 IEEE International Conference on Robotics and
  Automation (ICRA)}, pages 4775--4782. IEEE, 2015.

\bibitem[Batinovic et~al.(2022)Batinovic, Ivanovic, Petrovic, and
  Bogdan]{batinovic2022shadowcasting}
Ana Batinovic, Antun Ivanovic, Tamara Petrovic, and Stjepan Bogdan.
\newblock A shadowcasting-based next-best-view planner for autonomous 3d
  exploration.
\newblock \emph{IEEE Robotics and Automation Letters}, 7\penalty0 (2):\penalty0
  2969--2976, 2022.

\bibitem[Bircher et~al.(2016)Bircher, Kamel, Alexis, Oleynikova, and
  Siegwart]{bircher2016receding}
Andreas Bircher, Mina Kamel, Kostas Alexis, Helen Oleynikova, and Roland
  Siegwart.
\newblock Receding horizon" next-best-view" planner for 3d exploration.
\newblock In \emph{2016 IEEE international conference on robotics and
  automation (ICRA)}, pages 1462--1468. IEEE, 2016.

\bibitem[Bircher et~al.(2018)Bircher, Kamel, Alexis, Oleynikova, and
  Siegwart]{bircher2018receding}
Andreas Bircher, Mina Kamel, Kostas Alexis, Helen Oleynikova, and Roland
  Siegwart.
\newblock Receding horizon path planning for 3d exploration and surface
  inspection.
\newblock \emph{Autonomous Robots}, 42\penalty0 (2):\penalty0 291--306, 2018.

\bibitem[Carlone et~al.(2014)Carlone, Du, Kaouk~Ng, Bona, and
  Indri]{carlone2014active}
Luca Carlone, Jingjing Du, Miguel Kaouk~Ng, Basilio Bona, and Marina Indri.
\newblock Active slam and exploration with particle filters using
  kullback-leibler divergence.
\newblock \emph{Journal of Intelligent \& Robotic Systems}, 75\penalty0
  (2):\penalty0 291--311, 2014.

\bibitem[Carrillo et~al.(2012)Carrillo, Reid, and
  Castellanos]{carrillo2012comparison}
Henry Carrillo, Ian Reid, and Jos{\'e}~A Castellanos.
\newblock On the comparison of uncertainty criteria for active slam.
\newblock In \emph{2012 IEEE International Conference on Robotics and
  Automation}, pages 2080--2087. IEEE, 2012.

\bibitem[Chen et~al.(2020)Chen, Martin, Huang, Wang, and
  Englot]{chen2020autonomous}
Fanfei Chen, John~D Martin, Yewei Huang, Jinkun Wang, and Brendan Englot.
\newblock Autonomous exploration under uncertainty via deep reinforcement
  learning on graphs.
\newblock In \emph{2020 IEEE/RSJ International Conference on Intelligent Robots
  and Systems (IROS)}, pages 6140--6147. IEEE, 2020.

\bibitem[Chen et~al.(2025)Chen, Zhan, Chen, Xu, Yan, Cai, and
  Xu]{chen2025activegamer}
Liyan Chen, Huangying Zhan, Kevin Chen, Xiangyu Xu, Qingan Yan, Changjiang Cai,
  and Yi Xu.
\newblock Activegamer: Active gaussian mapping through efficient rendering.
\newblock \emph{arXiv preprint arXiv:2501.06897}, 2025.

\bibitem[Chen et~al.(2022)Chen, Zhang, Li, Chen, Feng, Wang, and
  Wang]{chen2022hallucinated}
Xingyu Chen, Qi Zhang, Xiaoyu Li, Yue Chen, Ying Feng, Xuan Wang, and Jue Wang.
\newblock Hallucinated neural radiance fields in the wild.
\newblock In \emph{Proceedings of the IEEE/CVF Conference on Computer Vision
  and Pattern Recognition}, pages 12943--12952, 2022.

\bibitem[Connolly(1985)]{connolly1985determination}
Cl Connolly.
\newblock The determination of next best views.
\newblock In \emph{Proceedings. 1985 IEEE international conference on robotics
  and automation}, pages 432--435. IEEE, 1985.

\bibitem[Dai et~al.(2020)Dai, Papatheodorou, Funk, Tzoumanikas, and
  Leutenegger]{dai2020fast}
Anna Dai, Sotiris Papatheodorou, Nils Funk, Dimos Tzoumanikas, and Stefan
  Leutenegger.
\newblock Fast frontier-based information-driven autonomous exploration with an
  mav.
\newblock In \emph{2020 IEEE international conference on robotics and
  automation (ICRA)}, pages 9570--9576. IEEE, 2020.

\bibitem[Dang et~al.(2019)Dang, Mascarich, Khattak, Papachristos, and
  Alexis]{dang2019graph}
Tung Dang, Frank Mascarich, Shehryar Khattak, Christos Papachristos, and Kostas
  Alexis.
\newblock Graph-based path planning for autonomous robotic exploration in
  subterranean environments.
\newblock In \emph{2019 IEEE/RSJ International Conference on Intelligent Robots
  and Systems (IROS)}, pages 3105--3112. IEEE, 2019.

\bibitem[Deng et~al.(2022)Deng, Liu, Zhu, and Ramanan]{deng2022depth}
Kangle Deng, Andrew Liu, Jun-Yan Zhu, and Deva Ramanan.
\newblock Depth-supervised nerf: Fewer views and faster training for free.
\newblock In \emph{Proceedings of the IEEE/CVF conference on computer vision
  and pattern recognition}, pages 12882--12891, 2022.

\bibitem[Dornhege and Kleiner(2013)]{dornhege2013frontier}
Christian Dornhege and Alexander Kleiner.
\newblock A frontier-void-based approach for autonomous exploration in 3d.
\newblock \emph{Advanced Robotics}, 27\penalty0 (6):\penalty0 459--468, 2013.

\bibitem[Feng et~al.(2024)Feng, Zhan, Chen, Yan, Xu, Cai, Li, Zhu, and
  Xu]{feng2024naruto}
Ziyue Feng, Huangying Zhan, Zheng Chen, Qingan Yan, Xiangyu Xu, Changjiang Cai,
  Bing Li, Qilun Zhu, and Yi Xu.
\newblock Naruto: Neural active reconstruction from uncertain target
  observations.
\newblock In \emph{Proceedings of the IEEE/CVF Conference on Computer Vision
  and Pattern Recognition}, pages 21572--21583, 2024.

\bibitem[Fridovich-Keil et~al.(2022)Fridovich-Keil, Yu, Tancik, Chen, Recht,
  and Kanazawa]{fridovich2022plenoxels}
Sara Fridovich-Keil, Alex Yu, Matthew Tancik, Qinhong Chen, Benjamin Recht, and
  Angjoo Kanazawa.
\newblock Plenoxels: Radiance fields without neural networks.
\newblock In \emph{Proceedings of the IEEE/CVF conference on computer vision
  and pattern recognition}, pages 5501--5510, 2022.

\bibitem[Gal and Ghahramani(2016)]{gal2016dropout}
Yarin Gal and Zoubin Ghahramani.
\newblock Dropout as a bayesian approximation: Representing model uncertainty
  in deep learning.
\newblock In \emph{international conference on machine learning}, pages
  1050--1059. PMLR, 2016.

\bibitem[Gawlikowski et~al.(2023)Gawlikowski, Tassi, Ali, Lee, Humt, Feng,
  Kruspe, Triebel, Jung, Roscher, et~al.]{gawlikowski2023survey}
Jakob Gawlikowski, Cedrique Rovile~Njieutcheu Tassi, Mohsin Ali, Jongseok Lee,
  Matthias Humt, Jianxiang Feng, Anna Kruspe, Rudolph Triebel, Peter Jung,
  Ribana Roscher, et~al.
\newblock A survey of uncertainty in deep neural networks.
\newblock \emph{Artificial Intelligence Review}, 56\penalty0 (Suppl
  1):\penalty0 1513--1589, 2023.

\bibitem[Goli et~al.(2024)Goli, Reading, Sell{\'a}n, Jacobson, and
  Tagliasacchi]{goli2023bayes}
Lily Goli, Cody Reading, Silvia Sell{\'a}n, Alec Jacobson, and Andrea
  Tagliasacchi.
\newblock Bayes' rays: Uncertainty quantification for neural radiance fields.
\newblock In \emph{Proceedings of the IEEE/CVF Conference on Computer Vision
  and Pattern Recognition}, pages 20061--20070, 2024.

\bibitem[Gomez et~al.(2019)Gomez, Hernandez, and Barber]{gomez2019topological}
Clara Gomez, Alejandra~C Hernandez, and Ramon Barber.
\newblock Topological frontier-based exploration and map-building using
  semantic information.
\newblock \emph{Sensors}, 19\penalty0 (20):\penalty0 4595, 2019.

\bibitem[Guo et~al.(2017)Guo, Pleiss, Sun, and Weinberger]{guo2017calibration}
Chuan Guo, Geoff Pleiss, Yu Sun, and Kilian~Q Weinberger.
\newblock On calibration of modern neural networks.
\newblock In \emph{International conference on machine learning}, pages
  1321--1330. PMLR, 2017.

\bibitem[Holz et~al.(2010)Holz, Basilico, Amigoni, and
  Behnke]{holz2010evaluating}
Dirk Holz, Nicola Basilico, Francesco Amigoni, and Sven Behnke.
\newblock Evaluating the efficiency of frontier-based exploration strategies.
\newblock In \emph{ISR 2010 (41st International Symposium on Robotics) and
  ROBOTIK 2010 (6th German Conference on Robotics)}, pages 1--8. VDE, 2010.

\bibitem[Huang et~al.(2024)Huang, Li, Yang, Wang, and Liang]{huang20243d}
Xuan Huang, Hanhui Li, Zejun Yang, Zhisheng Wang, and Xiaodan Liang.
\newblock 3d visibility-aware generalizable neural radiance fields for
  interacting hands.
\newblock In \emph{Proceedings of the AAAI Conference on Artificial
  Intelligence}, pages 2400--2408, 2024.

\bibitem[Huber et~al.(2008)Huber, Bailey, Durrant-Whyte, and
  Hanebeck]{huber2008entropy}
Marco~F Huber, Tim Bailey, Hugh Durrant-Whyte, and Uwe~D Hanebeck.
\newblock On entropy approximation for gaussian mixture random vectors.
\newblock In \emph{2008 IEEE International Conference on Multisensor Fusion and
  Integration for Intelligent Systems}, pages 181--188. IEEE, 2008.

\bibitem[Ilg et~al.(2018)Ilg, Cicek, Galesso, Klein, Makansi, Hutter, and
  Brox]{ilg2018uncertainty}
Eddy Ilg, Ozgun Cicek, Silvio Galesso, Aaron Klein, Osama Makansi, Frank
  Hutter, and Thomas Brox.
\newblock Uncertainty estimates and multi-hypotheses networks for optical flow.
\newblock In \emph{Proceedings of the European Conference on Computer Vision
  (ECCV)}, pages 652--667, 2018.

\bibitem[Jackson(1999)]{jackson1999ced}
John~David Jackson.
\newblock \emph{Classical Electrodynamics}.
\newblock John Wiley \& Sons, Inc., Hoboken, NJ, 3rd edition, 1999.

\bibitem[Jiang et~al.(2024{\natexlab{a}})Jiang, Lei, Ashton, and
  Daniilidis]{jiang2024ag}
Wen Jiang, Boshu Lei, Katrina Ashton, and Kostas Daniilidis.
\newblock Ag-slam: Active gaussian splatting slam.
\newblock 2024{\natexlab{a}}.

\bibitem[Jiang et~al.(2024{\natexlab{b}})Jiang, Lei, and
  Daniilidis]{jiang2023fisherrf}
Wen Jiang, Boshu Lei, and Kostas Daniilidis.
\newblock Fisherrf: Active view selection and uncertainty quantification for
  radiance fields using fisher information.
\newblock \emph{ECCV}, 2024{\natexlab{b}}.

\bibitem[Jin et~al.(2023)Jin, Chen, R{\"u}ckin, and Popovi{\'c}]{jin2023neu}
Liren Jin, Xieyuanli Chen, Julius R{\"u}ckin, and Marija Popovi{\'c}.
\newblock Neu-nbv: Next best view planning using uncertainty estimation in
  image-based neural rendering.
\newblock In \emph{2023 IEEE/RSJ International Conference on Intelligent Robots
  and Systems (IROS)}, pages 11305--11312. IEEE, 2023.

\bibitem[Jin et~al.(2025)Jin, Zhong, Pan, Behley, Stachniss, and
  Popovi{\'c}]{jin2025activegs}
Liren Jin, Xingguang Zhong, Yue Pan, Jens Behley, Cyrill Stachniss, and Marija
  Popovi{\'c}.
\newblock Activegs: Active scene reconstruction using gaussian splatting.
\newblock \emph{IEEE Robotics and Automation Letters}, 2025.

\bibitem[Jin et~al.(2024)Jin, Gao, Wang, Wu, Lu, Xu, and Gao]{jin2024gs}
Rui Jin, Yuman Gao, Yingjian Wang, Yuze Wu, Haojian Lu, Chao Xu, and Fei Gao.
\newblock Gs-planner: A gaussian-splatting-based planning framework for active
  high-fidelity reconstruction.
\newblock In \emph{2024 IEEE/RSJ International Conference on Intelligent Robots
  and Systems (IROS)}, pages 11202--11209. IEEE, 2024.

\bibitem[Keetha et~al.(2024)Keetha, Karhade, Jatavallabhula, Yang, Scherer,
  Ramanan, and Luiten]{keetha2024splatam}
Nikhil Keetha, Jay Karhade, Krishna~Murthy Jatavallabhula, Gengshan Yang,
  Sebastian Scherer, Deva Ramanan, and Jonathon Luiten.
\newblock Splatam: Splat track \& map 3d gaussians for dense rgb-d slam.
\newblock In \emph{Proceedings of the IEEE/CVF Conference on Computer Vision
  and Pattern Recognition}, pages 21357--21366, 2024.

\bibitem[Keidar and Kaminka(2012)]{keidar2012robot}
Matan Keidar and Gal~A Kaminka.
\newblock Robot exploration with fast frontier detection: Theory and
  experiments.
\newblock In \emph{Proceedings of the 11th International Conference on
  Autonomous Agents and Multiagent Systems-Volume 1}, pages 113--120, 2012.

\bibitem[Keidar and Kaminka(2014)]{keidar2014efficient}
Matan Keidar and Gal~A Kaminka.
\newblock Efficient frontier detection for robot exploration.
\newblock \emph{The International Journal of Robotics Research}, 33\penalty0
  (2):\penalty0 215--236, 2014.

\bibitem[Kerbl et~al.(2023)Kerbl, Kopanas, Leimk{\"u}hler, and
  Drettakis]{kerbl20233d}
Bernhard Kerbl, Georgios Kopanas, Thomas Leimk{\"u}hler, and George Drettakis.
\newblock 3d gaussian splatting for real-time radiance field rendering.
\newblock \emph{ACM Trans. Graph.}, 42\penalty0 (4):\penalty0 139--1, 2023.

\bibitem[Kirsch and Gal(2022)]{kirsch2022unifying}
Andreas Kirsch and Yarin Gal.
\newblock Unifying approaches in active learning and active sampling via fisher
  information and information-theoretic quantities.
\newblock \emph{arXiv preprint arXiv:2208.00549}, 2022.

\bibitem[Klasson et~al.(2025)Klasson, Mereu, Kannala, and
  Solin]{klasson2025sources}
Marcus Klasson, Riccardo Mereu, Juho Kannala, and Arno Solin.
\newblock Sources of uncertainty in 3d scene reconstruction.
\newblock In \emph{European Conference on Computer Vision}, pages 271--289.
  Springer, 2025.

\bibitem[Kollar and Roy(2008)]{kollar2008trajectory}
Thomas Kollar and Nicholas Roy.
\newblock Trajectory optimization using reinforcement learning for map
  exploration.
\newblock \emph{The International Journal of Robotics Research}, 27\penalty0
  (2):\penalty0 175--196, 2008.

\bibitem[Kompis et~al.(2021)Kompis, Bartolomei, Mascaro, Teixeira, and
  Chli]{kompis2021informed}
Yves Kompis, Luca Bartolomei, Ruben Mascaro, Lucas Teixeira, and Margarita
  Chli.
\newblock Informed sampling exploration path planner for 3d reconstruction of
  large scenes.
\newblock \emph{IEEE Robotics and Automation Letters}, 6\penalty0 (4):\penalty0
  7893--7900, 2021.

\bibitem[Kopanas and Drettakis(2023)]{kopanas2023improving}
Georgios Kopanas and George Drettakis.
\newblock Improving nerf quality by progressive camera placement for
  free-viewpoint navigation.
\newblock 2023.

\bibitem[Kuang et~al.(2024)Kuang, Yan, Zhao, Zhou, and Zha]{kuang2024active}
Zijia Kuang, Zike Yan, Hao Zhao, Guyue Zhou, and Hongbin Zha.
\newblock Active neural mapping at scale.
\newblock In \emph{2024 IEEE/RSJ International Conference on Intelligent Robots
  and Systems (IROS)}, pages 7152--7159. IEEE, 2024.

\bibitem[Lee et~al.(2022)Lee, Chen, Wang, Liniger, Kumar, and
  Yu]{lee2022uncertainty}
Soomin Lee, Le Chen, Jiahao Wang, Alexander Liniger, Suryansh Kumar, and Fisher
  Yu.
\newblock Uncertainty guided policy for active robotic 3d reconstruction using
  neural radiance fields.
\newblock \emph{IEEE Robotics and Automation Letters}, 7\penalty0 (4):\penalty0
  12070--12077, 2022.

\bibitem[Lee et~al.(2024)Lee, Kang, Ha, and Yu]{lee2024bayesian}
Sibeak Lee, Kyeongsu Kang, Seongbo Ha, and Hyeonwoo Yu.
\newblock Bayesian nerf: Quantifying uncertainty with volume density for neural
  implicit fields.
\newblock \emph{arXiv preprint arXiv:2404.06727}, 2024.

\bibitem[Li and Cheung(2024)]{li2024variational}
Ruiqi Li and Yiu-ming Cheung.
\newblock Variational multi-scale representation for estimating uncertainty in
  3d gaussian splatting.
\newblock \emph{Advances in Neural Information Processing Systems},
  37:\penalty0 87934--87958, 2024.

\bibitem[Li et~al.()Li, Li, Tang, and Gu]{li2024bavsnerf}
You Li, Rui Li, Shengjun Tang, and Yanjie Gu.
\newblock Bavsnerf: Batch active view selection for neural radiance field using
  scene uncertainty.
\newblock In \emph{The First Workshop on Populating Empty Cities--Virtual
  Humans for Robotics and Autonomous Driving at CVPR 2024}.

\bibitem[Li et~al.(2025)Li, Kuang, Li, Hao, Yan, Zhou, and
  Zhang]{li2025activesplat}
Yuetao Li, Zijia Kuang, Ting Li, Qun Hao, Zike Yan, Guyue Zhou, and Shaohui
  Zhang.
\newblock Activesplat: High-fidelity scene reconstruction through active
  gaussian splatting.
\newblock \emph{IEEE Robotics and Automation Letters}, 2025.

\bibitem[Lluvia et~al.(2021)Lluvia, Lazkano, and Ansuategi]{lluvia2021active}
Iker Lluvia, Elena Lazkano, and Ander Ansuategi.
\newblock Active mapping and robot exploration: A survey.
\newblock \emph{Sensors}, 21\penalty0 (7):\penalty0 2445, 2021.

\bibitem[Lu et~al.(2020)Lu, Redondo, and Campoy]{lu2020optimal}
Liang Lu, Carlos Redondo, and Pascual Campoy.
\newblock Optimal frontier-based autonomous exploration in unconstructed
  environment using rgb-d sensor.
\newblock \emph{Sensors}, 20\penalty0 (22):\penalty0 6507, 2020.

\bibitem[Luo et~al.(2024)Luo, Chen, and Krishnapriyan]{luo2024enabling}
Shengjie Luo, Tianlang Chen, and Aditi~S Krishnapriyan.
\newblock Enabling efficient equivariant operations in the fourier basis via
  gaunt tensor products.
\newblock \emph{arXiv preprint arXiv:2401.10216}, 2024.

\bibitem[Lyu et~al.(2024)Lyu, Tewari, Habermann, Saito, Zollh{\"o}fer,
  Leimk{\"u}hler, and Theobalt]{lyu2024manifold}
Linjie Lyu, Ayush Tewari, Marc Habermann, Shunsuke Saito, Michael
  Zollh{\"o}fer, Thomas Leimk{\"u}hler, and Christian Theobalt.
\newblock Manifold sampling for differentiable uncertainty in radiance fields.
\newblock In \emph{SIGGRAPH Asia 2024 Conference Papers}, pages 1--11, 2024.

\bibitem[Matsuki et~al.(2024)Matsuki, Murai, Kelly, and
  Davison]{matsuki2024gaussian}
Hidenobu Matsuki, Riku Murai, Paul~HJ Kelly, and Andrew~J Davison.
\newblock Gaussian splatting slam.
\newblock In \emph{Proceedings of the IEEE/CVF Conference on Computer Vision
  and Pattern Recognition}, pages 18039--18048, 2024.

\bibitem[Meng et~al.(2017)Meng, Qin, Chen, Chen, Sun, Lin, and
  Ang]{meng2017two}
Zehui Meng, Hailong Qin, Ziyue Chen, Xudong Chen, Hao Sun, Feng Lin, and
  Marcelo~H Ang.
\newblock A two-stage optimized next-view planning framework for 3-d unknown
  environment exploration, and structural reconstruction.
\newblock \emph{IEEE Robotics and Automation Letters}, 2\penalty0 (3):\penalty0
  1680--1687, 2017.

\bibitem[Mildenhall et~al.(2021)Mildenhall, Srinivasan, Tancik, Barron,
  Ramamoorthi, and Ng]{mildenhall2021nerf}
Ben Mildenhall, Pratul~P Srinivasan, Matthew Tancik, Jonathan~T Barron, Ravi
  Ramamoorthi, and Ren Ng.
\newblock Nerf: Representing scenes as neural radiance fields for view
  synthesis.
\newblock \emph{Communications of the ACM}, 65\penalty0 (1):\penalty0 99--106,
  2021.

\bibitem[Mu et~al.(2016)Mu, Giamou, Paull, Agha-mohammadi, Leonard, and
  How]{mu2016information}
Beipeng Mu, Matthew Giamou, Liam Paull, Ali-akbar Agha-mohammadi, John Leonard,
  and Jonathan How.
\newblock Information-based active slam via topological feature graphs.
\newblock In \emph{2016 IEEE 55th Conference on decision and control (Cdc)},
  pages 5583--5590. IEEE, 2016.

\bibitem[M{\"u}ller et~al.(2022)M{\"u}ller, Evans, Schied, and
  Keller]{muller2022instant}
Thomas M{\"u}ller, Alex Evans, Christoph Schied, and Alexander Keller.
\newblock Instant neural graphics primitives with a multiresolution hash
  encoding.
\newblock \emph{ACM transactions on graphics (TOG)}, 41\penalty0 (4):\penalty0
  1--15, 2022.

\bibitem[Nakayama et~al.(2024)Nakayama, Uy, You, Li, and
  Guibas]{nakayama2024provnerf}
Kiyohiro Nakayama, Mikaela~Angelina Uy, Yang You, Ke Li, and Leonidas~J Guibas.
\newblock Provnerf: Modeling per point provenance in nerfs as a stochastic
  field.
\newblock \emph{Advances in Neural Information Processing Systems},
  37:\penalty0 99145--99160, 2024.

\bibitem[Niroui et~al.(2019)Niroui, Zhang, Kashino, and Nejat]{niroui2019deep}
Farzad Niroui, Kaicheng Zhang, Zendai Kashino, and Goldie Nejat.
\newblock Deep reinforcement learning robot for search and rescue applications:
  Exploration in unknown cluttered environments.
\newblock \emph{IEEE Robotics and Automation Letters}, 4\penalty0 (2):\penalty0
  610--617, 2019.

\bibitem[Olver(2010)]{olver2010nist}
Frank~WJ Olver.
\newblock \emph{NIST handbook of mathematical functions hardback and CD-ROM}.
\newblock Cambridge university press, 2010.

\bibitem[Ovadia et~al.(2019)Ovadia, Fertig, Ren, Nado, Sculley, Nowozin,
  Dillon, Lakshminarayanan, and Snoek]{ovadia2019can}
Yaniv Ovadia, Emily Fertig, Jie Ren, Zachary Nado, David Sculley, Sebastian
  Nowozin, Joshua Dillon, Balaji Lakshminarayanan, and Jasper Snoek.
\newblock Can you trust your model's uncertainty? evaluating predictive
  uncertainty under dataset shift.
\newblock \emph{Advances in neural information processing systems}, 32, 2019.

\bibitem[Pan et~al.(2022)Pan, Lai, Song, and Huang]{pan2022activenerf}
Xuran Pan, Zihang Lai, Shiji Song, and Gao Huang.
\newblock Activenerf: Learning where to see with uncertainty estimation.
\newblock In \emph{European Conference on Computer Vision}, pages 230--246.
  Springer, 2022.

\bibitem[Papachristos et~al.(2017)Papachristos, Khattak, and
  Alexis]{papachristos2017uncertainty}
Christos Papachristos, Shehryar Khattak, and Kostas Alexis.
\newblock Uncertainty-aware receding horizon exploration and mapping using
  aerial robots.
\newblock In \emph{2017 IEEE international conference on robotics and
  automation (ICRA)}, pages 4568--4575. IEEE, 2017.

\bibitem[Placed et~al.(2023)Placed, Strader, Carrillo, Atanasov, Indelman,
  Carlone, and Castellanos]{placed2023survey}
Julio~A Placed, Jared Strader, Henry Carrillo, Nikolay Atanasov, Vadim
  Indelman, Luca Carlone, and Jos{\'e}~A Castellanos.
\newblock A survey on active simultaneous localization and mapping: State of
  the art and new frontiers.
\newblock \emph{IEEE Transactions on Robotics}, 2023.

\bibitem[Poggi et~al.(2020)Poggi, Aleotti, Tosi, and
  Mattoccia]{poggi2020uncertainty}
Matteo Poggi, Filippo Aleotti, Fabio Tosi, and Stefano Mattoccia.
\newblock On the uncertainty of self-supervised monocular depth estimation.
\newblock In \emph{Proceedings of the IEEE/CVF conference on computer vision
  and pattern recognition}, pages 3227--3237, 2020.

\bibitem[Qiao et~al.(2018)Qiao, Fang, and Si]{qiao2018sample}
Wenchuan Qiao, Zheng Fang, and Bailu Si.
\newblock Sample-based frontier detection for autonomous robot exploration.
\newblock In \emph{2018 IEEE International Conference on Robotics and
  Biomimetics (ROBIO)}, pages 1165--1170. IEEE, 2018.

\bibitem[Ramakrishnan et~al.(2021)Ramakrishnan, Gokaslan, Wijmans, Maksymets,
  Clegg, Turner, Undersander, Galuba, Westbury, Chang, Savva, Zhao, and
  Batra]{ramakrishnan2021hm3d}
Santhosh~Kumar Ramakrishnan, Aaron Gokaslan, Erik Wijmans, Oleksandr Maksymets,
  Alexander Clegg, John~M Turner, Eric Undersander, Wojciech Galuba, Andrew
  Westbury, Angel~X Chang, Manolis Savva, Yili Zhao, and Dhruv Batra.
\newblock Habitat-matterport 3d dataset ({HM}3d): 1000 large-scale 3d
  environments for embodied {AI}.
\newblock In \emph{Thirty-fifth Conference on Neural Information Processing
  Systems Datasets and Benchmarks Track}, 2021.

\bibitem[Ran et~al.(2023)Ran, Zeng, He, Chen, Li, Chen, Lee, and Ye]{neurar}
Yunlong Ran, Jing Zeng, Shibo He, Jiming Chen, Lincheng Li, Yingfeng Chen,
  Gimhee Lee, and Qi Ye.
\newblock Neurar: Neural uncertainty for autonomous 3d reconstruction with
  implicit neural representations.
\newblock \emph{IEEE Robotics and Automation Letters}, 8\penalty0 (2):\penalty0
  1125--1132, 2023.

\bibitem[Respall et~al.(2021)Respall, Devitt, Fedorenko, and
  Klimchik]{respall2021fast}
Victor~Massagu{\'e} Respall, Dmitry Devitt, Roman Fedorenko, and Alexandr
  Klimchik.
\newblock Fast sampling-based next-best-view exploration algorithm for a mav.
\newblock In \emph{2021 IEEE International Conference on Robotics and
  Automation (ICRA)}, pages 89--95. IEEE, 2021.

\bibitem[Sakurai and Napolitano(2020)]{sakurai2020modern}
Jun~John Sakurai and Jim Napolitano.
\newblock \emph{Modern quantum mechanics}.
\newblock Cambridge University Press, 2020.

\bibitem[Savva et~al.(2019)Savva, Kadian, Maksymets, Zhao, Wijmans, Jain,
  Straub, Liu, Koltun, Malik, et~al.]{savva2019habitat}
Manolis Savva, Abhishek Kadian, Oleksandr Maksymets, Yili Zhao, Erik Wijmans,
  Bhavana Jain, Julian Straub, Jia Liu, Vladlen Koltun, Jitendra Malik, et~al.
\newblock Habitat: A platform for embodied ai research.
\newblock In \emph{Proceedings of the IEEE/CVF international conference on
  computer vision}, pages 9339--9347, 2019.

\bibitem[Selin et~al.(2019)Selin, Tiger, Duberg, Heintz, and
  Jensfelt]{selin2019efficient}
Magnus Selin, Mattias Tiger, Daniel Duberg, Fredrik Heintz, and Patric
  Jensfelt.
\newblock Efficient autonomous exploration planning of large-scale 3-d
  environments.
\newblock \emph{IEEE Robotics and Automation Letters}, 4\penalty0 (2):\penalty0
  1699--1706, 2019.

\bibitem[Senarathne and Wang(2016)]{senarathne2016towards}
PGCN Senarathne and Danwei Wang.
\newblock Towards autonomous 3d exploration using surface frontiers.
\newblock In \emph{2016 IEEE International Symposium on Safety, Security, and
  Rescue Robotics (SSRR)}, pages 34--41. IEEE, 2016.

\bibitem[Shen et~al.(2021)Shen, Ruiz, Agudo, and
  Moreno-Noguer]{shen2021stochastic}
Jianxiong Shen, Adria Ruiz, Antonio Agudo, and Francesc Moreno-Noguer.
\newblock Stochastic neural radiance fields: Quantifying uncertainty in
  implicit 3d representations.
\newblock In \emph{2021 International Conference on 3D Vision (3DV)}, pages
  972--981. IEEE, 2021.

\bibitem[Shen et~al.(2022)Shen, Agudo, Moreno-Noguer, and
  Ruiz]{shen2022conditional}
Jianxiong Shen, Antonio Agudo, Francesc Moreno-Noguer, and Adria Ruiz.
\newblock Conditional-flow nerf: Accurate 3d modelling with reliable
  uncertainty quantification.
\newblock In \emph{European Conference on Computer Vision}, pages 540--557.
  Springer, 2022.

\bibitem[Shen et~al.(2024)Shen, Ren, Ruiz, and
  Moreno-Noguer]{shen2024estimating}
Jianxiong Shen, Ruijie Ren, Adria Ruiz, and Francesc Moreno-Noguer.
\newblock Estimating 3d uncertainty field: Quantifying uncertainty for neural
  radiance fields.
\newblock In \emph{2024 IEEE International Conference on Robotics and
  Automation (ICRA)}, pages 2375--2381. IEEE, 2024.

\bibitem[Shen et~al.(2012)Shen, Michael, and Kumar]{shen2012stochastic}
Shaojie Shen, Nathan Michael, and Vijay Kumar.
\newblock Stochastic differential equation-based exploration algorithm for
  autonomous indoor 3d exploration with a micro-aerial vehicle.
\newblock \emph{The International Journal of Robotics Research}, 31\penalty0
  (12):\penalty0 1431--1444, 2012.

\bibitem[Shi et~al.(2021)Shi, Li, and Yu]{shi2021self}
Yujiao Shi, Hongdong Li, and Xin Yu.
\newblock Self-supervised visibility learning for novel view synthesis.
\newblock In \emph{Proceedings of the IEEE/CVF Conference on Computer Vision
  and Pattern Recognition}, pages 9675--9684, 2021.

\bibitem[Shih et~al.(2024)Shih, Ma, Boyice, Holynski, Cole, Curless, and
  Kontkanen]{shih2024extranerf}
Meng-Li Shih, Wei-Chiu Ma, Lorenzo Boyice, Aleksander Holynski, Forrester Cole,
  Brian Curless, and Janne Kontkanen.
\newblock Extranerf: Visibility-aware view extrapolation of neural radiance
  fields with diffusion models.
\newblock In \emph{Proceedings of the IEEE/CVF Conference on Computer Vision
  and Pattern Recognition}, pages 20385--20395, 2024.

\bibitem[Sim and Roy(2005)]{sim2005global}
Robert Sim and Nicholas Roy.
\newblock Global a-optimal robot exploration in slam.
\newblock In \emph{Proceedings of the 2005 IEEE international conference on
  robotics and automation}, pages 661--666. IEEE, 2005.

\bibitem[Somraj and Soundararajan(2023)]{somraj2023vip}
Nagabhushan Somraj and Rajiv Soundararajan.
\newblock Vip-nerf: Visibility prior for sparse input neural radiance fields.
\newblock In \emph{ACM SIGGRAPH 2023 conference proceedings}, pages 1--11,
  2023.

\bibitem[Srinivasan et~al.(2021)Srinivasan, Deng, Zhang, Tancik, Mildenhall,
  and Barron]{srinivasan2021nerv}
Pratul~P Srinivasan, Boyang Deng, Xiuming Zhang, Matthew Tancik, Ben
  Mildenhall, and Jonathan~T Barron.
\newblock Nerv: Neural reflectance and visibility fields for relighting and
  view synthesis.
\newblock In \emph{Proceedings of the IEEE/CVF conference on computer vision
  and pattern recognition}, pages 7495--7504, 2021.

\bibitem[Stachniss et~al.(2004)Stachniss, Hahnel, and
  Burgard]{stachniss2004exploration}
Cyrill Stachniss, Dirk Hahnel, and Wolfram Burgard.
\newblock Exploration with active loop-closing for fastslam.
\newblock In \emph{2004 IEEE/RSJ International Conference on Intelligent Robots
  and Systems (IROS)(IEEE Cat. No. 04CH37566)}, pages 1505--1510. IEEE, 2004.

\bibitem[Stachniss et~al.(2005)Stachniss, Grisetti, and
  Burgard]{stachniss2005information}
Cyrill Stachniss, Giorgio Grisetti, and Wolfram Burgard.
\newblock Information gain-based exploration using rao-blackwellized particle
  filters.
\newblock In \emph{Robotics: Science and systems}, pages 65--72, 2005.

\bibitem[Sucar et~al.(2021{\natexlab{a}})Sucar, Liu, Ortiz, and
  Davison]{Sucar:etal:ICCV2021}
Edgar Sucar, Shikun Liu, Joseph Ortiz, and Andrew~J Davison.
\newblock imap: Implicit mapping and positioning in real-time.
\newblock In \emph{Proceedings of the IEEE/CVF International Conference on
  Computer Vision}, pages 6229--6238, 2021{\natexlab{a}}.

\bibitem[Sucar et~al.(2021{\natexlab{b}})Sucar, Liu, Ortiz, and
  Davison]{sucar2021imap}
Edgar Sucar, Shikun Liu, Joseph Ortiz, and Andrew~J Davison.
\newblock imap: Implicit mapping and positioning in real-time.
\newblock In \emph{Proceedings of the IEEE/CVF international conference on
  computer vision}, pages 6229--6238, 2021{\natexlab{b}}.

\bibitem[S{\"u}nderhauf et~al.(2023)S{\"u}nderhauf, Abou-Chakra, and
  Miller]{sunderhauf2023density}
Niko S{\"u}nderhauf, Jad Abou-Chakra, and Dimity Miller.
\newblock Density-aware nerf ensembles: Quantifying predictive uncertainty in
  neural radiance fields.
\newblock In \emph{2023 IEEE International Conference on Robotics and
  Automation (ICRA)}, pages 9370--9376. IEEE, 2023.

\bibitem[Sutantyo et~al.(2013)Sutantyo, Levi, M{\"o}slinger, and
  Read]{sutantyo2013collective}
Donny Sutantyo, Paul Levi, Christoph M{\"o}slinger, and Mark Read.
\newblock Collective-adaptive l{\'e}vy flight for underwater multi-robot
  exploration.
\newblock In \emph{2013 IEEE International Conference on Mechatronics and
  Automation}, pages 456--462. IEEE, 2013.

\bibitem[Tao et~al.(2025)Tao, Ong, Murali, Spasojevic, Chaudhari, and
  Kumar]{tao2025rt}
Yuezhan Tao, Dexter Ong, Varun Murali, Igor Spasojevic, Pratik Chaudhari, and
  Vijay Kumar.
\newblock Rt-guide: Real-time gaussian splatting for information-driven
  exploration.
\newblock \emph{IEEE Robotics and Automation Letters}, 2025.

\bibitem[Taylor et~al.(1994)Taylor, Kuyatt, et~al.]{taylor1994guidelines}
Barry~N Taylor, Chris~E Kuyatt, et~al.
\newblock \emph{Guidelines for evaluating and expressing the uncertainty of
  NIST measurement results}.
\newblock US Department of Commerce, Technology Administration, National
  Institute of Standards and Technology, 1994.

\bibitem[Umari and Mukhopadhyay(2017)]{umari2017autonomous}
Hassan Umari and Shayok Mukhopadhyay.
\newblock Autonomous robotic exploration based on multiple rapidly-exploring
  randomized trees.
\newblock In \emph{2017 IEEE/RSJ International Conference on Intelligent Robots
  and Systems (IROS)}, pages 1396--1402. IEEE, 2017.

\bibitem[Valencia and Andrade-Cetto(2017)]{valencia2017active}
Rafael Valencia and Juan Andrade-Cetto.
\newblock Active pose slam.
\newblock In \emph{Mapping, planning and exploration with pose SLAM}, pages
  89--108. Springer, 2017.

\bibitem[Wang et~al.(2021)Wang, Feng, and Zhang]{wang2021rethinking}
Deng-Bao Wang, Lei Feng, and Min-Ling Zhang.
\newblock Rethinking calibration of deep neural networks: Do not be afraid of
  overconfidence.
\newblock \emph{Advances in Neural Information Processing Systems},
  34:\penalty0 11809--11820, 2021.

\bibitem[Warburg et~al.(2023)Warburg, Weber, Tancik, Holynski, and
  Kanazawa]{warburg2023nerfbusters}
Frederik Warburg, Ethan Weber, Matthew Tancik, Aleksander Holynski, and Angjoo
  Kanazawa.
\newblock Nerfbusters: Removing ghostly artifacts from casually captured nerfs.
\newblock In \emph{Proceedings of the IEEE/CVF International Conference on
  Computer Vision}, pages 18120--18130, 2023.

\bibitem[Wen et~al.(2020)Wen, Zhao, Yuan, Wang, Zhang, and
  Manfredi]{wen2020path}
Shuhuan Wen, Yanfang Zhao, Xiao Yuan, Zongtao Wang, Dan Zhang, and Luigi
  Manfredi.
\newblock Path planning for active slam based on deep reinforcement learning
  under unknown environments.
\newblock \emph{Intelligent Service Robotics}, 13\penalty0 (2):\penalty0
  263--272, 2020.

\bibitem[Whaite and Ferrie(1997)]{whaite1997autonomous}
Peter Whaite and Frank~P Ferrie.
\newblock Autonomous exploration: Driven by uncertainty.
\newblock \emph{IEEE Transactions on Pattern Analysis and Machine
  Intelligence}, 19\penalty0 (3):\penalty0 193--205, 1997.

\bibitem[Wigner(2012)]{wigner2012group}
Eugene Wigner.
\newblock \emph{Group theory: and its application to the quantum mechanics of
  atomic spectra}.
\newblock Elsevier, 2012.

\bibitem[Wilson et~al.(2025)Wilson, Almeida, Sun, Mahajan, Ghaffari, Ewen,
  Ghasemalizadeh, Kuo, and Sen]{wilson2025modeling}
Joey Wilson, Marcelino Almeida, Min Sun, Sachit Mahajan, Maani Ghaffari, Parker
  Ewen, Omid Ghasemalizadeh, Cheng-Hao Kuo, and Arnie Sen.
\newblock Modeling uncertainty in 3d gaussian splatting through continuous
  semantic splatting.
\newblock In \emph{2025 IEEE International Conference on Robotics and
  Automation (ICRA)}, pages 3284--3290. IEEE, 2025.

\bibitem[Xia et~al.(2018)Xia, Zamir, He, Sax, Malik, and
  Savarese]{xia2018gibson}
Fei Xia, Amir~R Zamir, Zhiyang He, Alexander Sax, Jitendra Malik, and Silvio
  Savarese.
\newblock Gibson env: Real-world perception for embodied agents.
\newblock In \emph{Proceedings of the IEEE conference on computer vision and
  pattern recognition}, pages 9068--9079, 2018.

\bibitem[Xin et~al.(2021)Xin, Zhou, An, Yan, Xu, Hu, and Yau]{xin2021fast}
Hanggao Xin, Zhiqian Zhou, Di An, Ling-Qi Yan, Kun Xu, Shi-Min Hu, and
  Shing-Tung Yau.
\newblock Fast and accurate spherical harmonics products.
\newblock \emph{ACM Trans. Graph.}, 40\penalty0 (6):\penalty0 280--1, 2021.

\bibitem[Xu et~al.(2025)Xu, Jin, Wu, Zhao, Zhang, Zhao, Gao, Gan, and
  Ding]{xu2025hgs}
Zijun Xu, Rui Jin, Ke Wu, Yi Zhao, Zhiwei Zhang, Jieru Zhao, Fei Gao, Zhongxue
  Gan, and Wenchao Ding.
\newblock Hgs-planner: Hierarchical planning framework for active scene
  reconstruction using 3d gaussian splatting.
\newblock In \emph{2025 IEEE International Conference on Robotics and
  Automation (ICRA)}, pages 14161--14167. IEEE, 2025.

\bibitem[Xue et~al.(2024)Xue, Dill, Mathur, Dellaert, Tsiotras, and
  Xu]{xue2024neural}
Shangjie Xue, Jesse Dill, Pranay Mathur, Frank Dellaert, Panagiotis Tsiotras,
  and Danfei Xu.
\newblock Neural visibility field for uncertainty-driven active mapping.
\newblock In \emph{Proceedings of the IEEE/CVF Conference on Computer Vision
  and Pattern Recognition}, pages 18122--18132, 2024.

\bibitem[Yamauchi(1997)]{yamauchi1997frontier}
Brian Yamauchi.
\newblock A frontier-based approach for autonomous exploration.
\newblock In \emph{Proceedings 1997 IEEE International Symposium on
  Computational Intelligence in Robotics and Automation CIRA'97.'Towards New
  Computational Principles for Robotics and Automation'}, pages 146--151. IEEE,
  1997.

\bibitem[Yamauchi(1998)]{yamauchi1998frontier}
Brian Yamauchi.
\newblock Frontier-based exploration using multiple robots.
\newblock In \emph{Proceedings of the second international conference on
  Autonomous agents}, pages 47--53, 1998.

\bibitem[Yan et~al.(2023)Yan, Liu, Quan, Chen, and Fu]{yan2023active}
Dongyu Yan, Jianheng Liu, Fengyu Quan, Haoyao Chen, and Mengmeng Fu.
\newblock Active implicit object reconstruction using uncertainty-guided
  next-best-view optimization.
\newblock \emph{IEEE Robotics and Automation Letters}, 2023.

\bibitem[Zeng et~al.(2025)Zeng, Ye, Liu, Xu, Li, Xu, Li, and
  Chen]{zeng2025multi}
Jing Zeng, Qi Ye, Tianle Liu, Yang Xu, Jin Li, Jinming Xu, Liang Li, and Jiming
  Chen.
\newblock Multi-robot autonomous 3d reconstruction using gaussian splatting
  with semantic guidance.
\newblock \emph{IEEE Robotics and Automation Letters}, 2025.

\bibitem[Zhang et~al.(2024)Zhang, Wang, Wang, Li, Qin, and
  Wang]{zhang2024gaussian}
Dongbin Zhang, Chuming Wang, Weitao Wang, Peihao Li, Minghan Qin, and Haoqian
  Wang.
\newblock Gaussian in the wild: 3d gaussian splatting for unconstrained image
  collections.
\newblock In \emph{European Conference on Computer Vision}, pages 341--359.
  Springer, 2024.

\bibitem[Zhang et~al.(2018)Zhang, Isola, Efros, Shechtman, and
  Wang]{zhang2018perceptual}
Richard Zhang, Phillip Isola, Alexei~A Efros, Eli Shechtman, and Oliver Wang.
\newblock The unreasonable effectiveness of deep features as a perceptual
  metric.
\newblock In \emph{Proceedings of the IEEE conference on computer vision and
  pattern recognition}, pages 586--595, 2018.

\bibitem[Zhang et~al.(2021)Zhang, Srinivasan, Deng, Debevec, Freeman, and
  Barron]{zhang2021nerfactor}
Xiuming Zhang, Pratul~P Srinivasan, Boyang Deng, Paul Debevec, William~T
  Freeman, and Jonathan~T Barron.
\newblock Nerfactor: Neural factorization of shape and reflectance under an
  unknown illumination.
\newblock \emph{ACM Transactions on Graphics (ToG)}, 40\penalty0 (6):\penalty0
  1--18, 2021.

\bibitem[Zhang and Fan(2025)]{zhang2025visibility}
Yikang Zhang and Rui Fan.
\newblock Visibility-aware densification for 3d gaussian splatting in dynamic
  urban scenes.
\newblock \emph{arXiv preprint arXiv:2510.09364}, 2025.

\bibitem[Zhang and Scaramuzza(2018)]{zhang2018perception}
Zichao Zhang and Davide Scaramuzza.
\newblock Perception-aware receding horizon navigation for mavs.
\newblock In \emph{2018 IEEE International Conference on Robotics and
  Automation (ICRA)}, pages 2534--2541. IEEE, 2018.

\bibitem[Zhu et~al.(2017)Zhu, Mottaghi, Kolve, Lim, Gupta, Fei-Fei, and
  Farhadi]{zhu2017target}
Yuke Zhu, Roozbeh Mottaghi, Eric Kolve, Joseph~J Lim, Abhinav Gupta, Li
  Fei-Fei, and Ali Farhadi.
\newblock Target-driven visual navigation in indoor scenes using deep
  reinforcement learning.
\newblock In \emph{2017 IEEE international conference on robotics and
  automation (ICRA)}, pages 3357--3364. IEEE, 2017.

\end{thebibliography}
}

\clearpage
\maketitlesupplementary

\section{Anisotropic
Visibility Field Details} \label{sec:appx_avf}

In this section, we provide details on anisotropic visibility field $V^{(i)}(\bm{d})$. We first derive the spherical harmonics (SH) representation of the directional visibility function $\nu(\bm{d};\bm{d_p})$ (Sec.~\ref{sec:appx_avf_vmf_coeff}), then discuss the SH representation of the visibility field $V^{(i)}(\bm{d})$ (Sec.~\ref{sec:appx_avf_sh},~\ref{sec:appx_avf_vis_bound}), and finally describe how $V^{(i)}(\bm{d})$ is efficiently constructed and queried during uncertainty-aware rasterization (Sec.~\ref{sec:appx_avf_construction}).

\begin{figure*}[t]
    \centering
    \includegraphics[width=\textwidth]{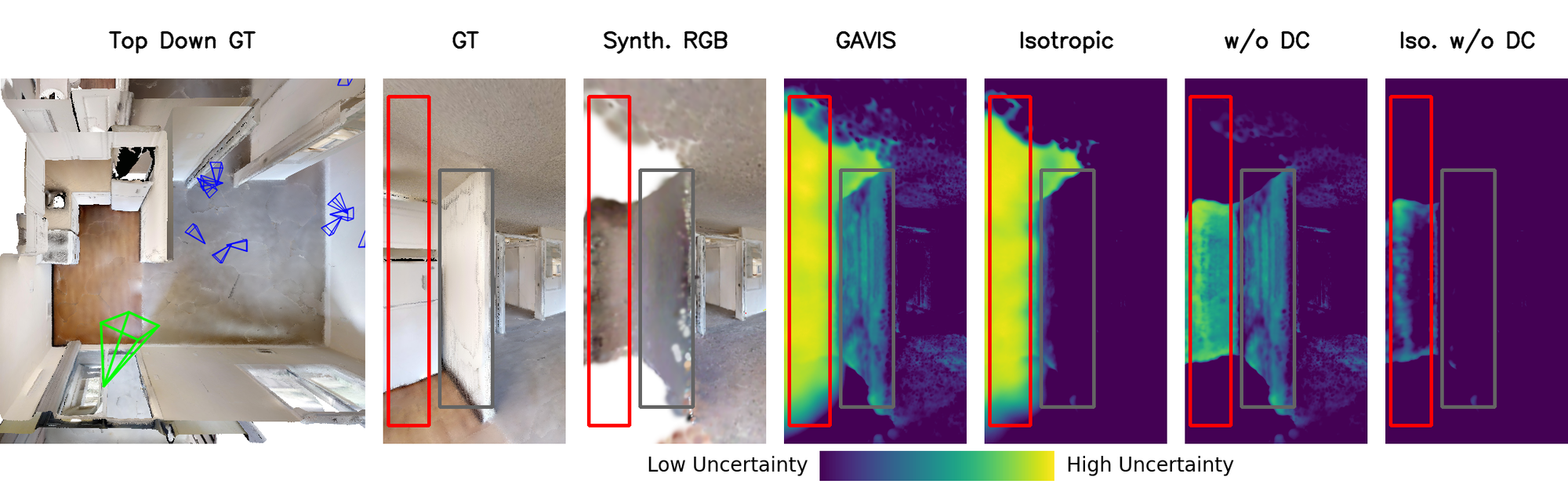}
    \caption{
    Illustration of the two major components of GAVIS: (1) anisotropic visibility (effects highlighted with gray boxes) and (2) visibility-field density control (effects highlighted with red boxes).  
From left to right:  
(i) top-down view of the scene with training views (blue frustums) covering only the right room and the queried view (green frustum) for uncertainty quantification;  
(ii) synthesized RGB image from a 3DGS model trained only on the partially covered region, leaving the rest of the scene unexplored;  
(iii) uncertainty estimated by full GAVIS;  
(iv) GAVIS with isotropic visibility;  
(v) GAVIS without visibility field density control;  
(vi) GAVIS with isotropic visibility and without density control. All methods are applied to the same learned 3DGS model.
    }
    \label{fig:ablation_illustration}
\end{figure*}

\subsection{Derivation of \eqref{eq:vmf_coeff}} \label{sec:appx_avf_vmf_coeff}

We represent $\nu(\bm{d};\bm{d_p})$ in the orthonormal basis of spherical harmonics $Y_\ell^m(\bm{d})$,
\begin{align} %
\begin{split}
\nu(\bm{d};\bm{d_p}) :=& \zeta \exp (\kappa \bm{d}\cdot \bm{d}_p) \\
=& \zeta \sum_{\ell=0}^\infty\sum_{m=-\ell}^\ell  a_{\ell m}\,Y_\ell^m(\bm{d}),
\end{split}
\end{align}
and derive an analytical expression for the coefficients $a_{\ell m}$.

Since $\exp(\kappa\,\bm{d}\!\cdot\!\bm{d_p})$ depends only on the dot‐product $x=\bm{d}\!\cdot\!\bm{d_p}$, we first expand in the one‐dimensional basis of Legendre polynomials $P_\ell(\bm{d}\!\cdot\!\bm{d_p})$ \cite{abramowitz1948handbook}:
\begin{equation}\label{eq:legendre}
\exp(\kappa\,\bm{d}\!\cdot\!\bm{d_p})
=\sum_{\ell=0}^\infty b_\ell\,P_\ell(\bm{d}\!\cdot\!\bm{d_p}),
\end{equation}

\noindent By comparing \eqref{eq:legendre} with the Legendre expansion of the exponential \cite{olver2010nist}:

\begin{equation*}
e^{\kappa x} \;=\; \sum_{\ell=0}^{\infty} (2\ell+1)\, i_{\ell}(\kappa)\, P_{\ell}(x).
\end{equation*}

\noindent Thus, we can identify the coefficients $b_\ell$ as
\begin{equation} \label{eq:bl}
b_\ell = (2\ell+1)\,i_\ell(\kappa).
\end{equation}
\noindent where $i_\ell$ is the modified spherical Bessel function of the first kind \cite{olver2010nist}.

Then we apply the addition theorem for spherical harmonics \cite{jackson1999ced}:
\begin{equation} \label{eq:add_theorem}
P_\ell\bigl(\bm{d}\!\cdot\!\bm{d_p}\bigr)
=\frac{4\pi}{2\ell+1}
\sum_{m=-\ell}^{\ell}
Y_\ell^m(\bm{d})\,Y_\ell^{m*}(\bm{d_p}).
\end{equation}

\noindent where $Y_\ell^{m*}(\bm{d_p})$ is the complex conjugate of $Y_\ell^m(\bm{d_p})$.

Substituting \eqref{eq:add_theorem} and \eqref{eq:bl} into \eqref{eq:legendre} gives
\begin{equation} \label{eq:v_full}
\nu(\bm{d};\bm{d_p})
=4\pi \zeta
\sum_{\ell=0}^\infty\sum_{m=-\ell}^\ell
i_\ell(\kappa)\;Y_\ell^{m}(\bm{d})\,Y_\ell^{m*}(\bm{d_p}).
\end{equation}

Comparing with \eqref{eq:SH_vMF}, we identify the coefficients
\begin{equation} \label{eq:appx:vmf_coeff}
a_{\ell m}
=4\pi\;i_\ell(\kappa)\;Y_\ell^{m*}(\bm{d_p}).
\end{equation}
\noindent where the modified spherical Bessel function of the first kind $i_\ell(\kappa)$ can be precomputed since $\kappa$ is a fixed hyperparameter (set to $\kappa = 1$).

Note that $\nu(\bm{d};\bm{d_p}):\mathbb{S}^2\to[0,1]$ is real-valued, so in practice we expand it in a real spherical harmonics basis for implementation.

\begin{equation} \label{eq:vis_real}
\nu(\bm{d};\bm{d_p})
=4\pi \zeta
\sum_{\ell=0}^\infty\sum_{m=-\ell}^\ell
i_\ell(\kappa)\;Y_{\ell m}(\bm{d})\,Y_{\ell m}(\bm{d_p}).
\end{equation}

\noindent where $Y_{\ell m}(\bm{d})$ is the real spherical harmonics function defined as
\begin{equation*}
Y_{\ell m}(\bm{d}) =
\begin{cases}
\sqrt{2} \, (-1)^m \, \mathrm{Im}\left[Y_\ell^{|m|}(\bm{d})\right], & m < 0 \\
Y_\ell^0(\bm{d}), & m = 0 \\
\sqrt{2} \, (-1)^m \, \mathrm{Re}\left[Y_\ell^{m}(\bm{d})\right], & m > 0
\end{cases}
\end{equation*}

\noindent and it could be verified that \eqref{eq:vis_real} is equivalent to \eqref{eq:v_full}.

Therefore, we obtain a closed-form expression for the coefficients representing the directional visibility function $\nu(\bm{d}; \bm{d_p})$ in spherical harmonics. 

\subsection{Details of SH Representation of \texorpdfstring{$V^{(i)}(\bm{d})$}{V(d)}} \label{sec:appx_avf_sh}

In this subsection, we provide further details on why computing the anisotropic visibility field $V^{(i)}(\bm{d})$ by directly combining \eqref{eq:SH_vMF} and \eqref{eq:vis_all} is impractical due to computational complexity.

We first rewrite \eqref{eq:vis_all} in a recursive form:
\begin{equation}    \label{eq:vis_all_rec}
V^{(i)}_{\mathcal{P} \cup \{\bm{p}\}  }(\bm{d}) = 
V^{(i)}_{\mathcal{P}}(\bm{d}) + V^{(i)}_{\bm{p}}(\bm{d}) - V^{(i)}_{\mathcal{P}}(\bm{d})V^{(i)}_{\bm{p}}(\bm{d})
\end{equation}
\noindent starting from $V^{(i)}_{\emptyset}(\bm{d})=0$, and then recursively computing the anisotropic visibility field over the entire training set.
Since spherical harmonics form a linear basis, the sum of two functions expressed in spherical harmonics can be obtained by simply adding their corresponding coefficients $a_{\ell m}$. However, their product poses significant challenges, particularly, the product of two spherical harmonics can be re-expanded as

\begin{align}
Y_{\ell_1}^{m_1}(\boldsymbol{d})\,Y_{\ell_2}^{m_2}(\boldsymbol{d}) 
&= \sum_{L = |\ell_1 - \ell_2|}^{\ell_1 + \ell_2}
   \sum_{M = -L}^{L}
   \sqrt{\frac{(2\ell_1+1)(2\ell_2+1)}{4\pi(2L+1)}} \nonumber
\\
&\quad\times
   C_{\ell_1 0\,\ell_2 0}^{L 0}\,
   C_{\ell_1 m_1\,\ell_2 m_2}^{L M}\,
   Y_{L}^{M}(\boldsymbol{d}) .
\end{align}
where $C_{\ell_1 m_1\,\ell_2 m_2}^{L M}$ denote the Clebsch\textendash Gordan coefficients~\cite{sakurai2020modern,wigner2012group}. As is shown in the above equation,
multiplying two spherical harmonics of degree $L$ produces a spherical harmonics expansion of degree up to $2L$. Since the multi-view aggregation in \eqref{eq:vis_all} requires performing the multiplication $|\mathcal{P}|$ times, the resulting degree can grow to $|\mathcal{P}|L$, requiring $O(|\mathcal{P}|^2 L^2)$ parameters to store. Since multiplying two spherical harmonics of degree $L$ can be done in $O(L^3)$ time \cite{luo2024enabling,xin2021fast}, computing the anisotropic visibility field over the entire training set $\mathcal{P}$ results in an overall complexity of $O(|\mathcal{P}|^4 L^3)$. Therefore, obtaining a closed-form expression for the spherical harmonics coefficients of the anisotropic visibility field is impractical.

\subsection{Derivation of \eqref{eq:vis_bound}} \label{sec:appx_avf_vis_bound}

Starting from \eqref{eq:vis_all}, we have

\begin{equation} \label{eq:appx:vis_full}
1 - V^{(i)}(\bm{d})
= \prod_{\bm{p} \in \mathcal{P}} (1 - V^{(i)}_{\bm{p}}(\bm{d}))
\end{equation}

Applying the arithmetic-geometric mean (AM-GM) inequality to the right-hand side of \eqref{eq:appx:vis_full} yields

\begin{equation} \label{eq:appx:vis_bound_agm}
\sqrt[|\mathcal{P}|]{\prod_{\bm{p} \in \mathcal{P}} (1 - V^{(i)}_{\bm{p}}(\bm{d}))}
\leq \frac{1}{|\mathcal{P}|}\sum_{\bm{p} \in \mathcal{P}} (1 - V^{(i)}_{\bm{p}}(\bm{d}))
\end{equation}

Therefore, combining \eqref{eq:appx:vis_full} and \eqref{eq:appx:vis_bound_agm} yields the lower bound on $V^{(i)}(\bm{d})$.
\begin{equation} \label{eq:appx:vis_bound_final}
V^{(i)}(\bm{d}) \;\ge\; 1 - \Bigl(1 - \tfrac{\sum_{\bm{p} \in \mathcal{P}} V^{(i)}_{\bm{p}}(\bm{d})}{|\mathcal{P}|} \Bigr)^{|\mathcal{P}|}.
\end{equation}

\noindent which is equivalent to \eqref{eq:vis_bound} since $\tilde{V}(\bm{d}) = \sum_{\bm{p} \in \mathcal{P}} V^{(i)}_{\bm{p}}(\bm{d})$.

This bound is tight when all $V^{(i)}_{\bm{p}}(\bm{d})$ are equal, i.e., when the visibility contributions from all camera views $\bm{p}$ are the same. In practice, this bound provides a computationally efficient way to estimate the overall visibility field $V^{(i)}(\bm{d})$ without having to compute the product of spherical harmonics of $V^{(i)}_{\bm{p}}(\bm{d})$ over all camera poses $\bm{p} \in \mathcal{P}$.

\subsection{Details of Visibility Field Construction and Query} \label{sec:appx_avf_construction}

We provide additional details on constructing the anisotropic visibility field and on efficient querying. 
Firstly, we describe our efficient algorithm for computing the single-view (isotropic) particle visibility $\Phi_{i,\bm{p}}\, T_{\bm{p}}(t_i^{\bm{p}})$ for each particle $i$ in training view $\bm{p}$ in Alg.~\ref{alg:pos_vis}, which builds on a modified 3DGS rasterizer.

\begin{algorithm}
\caption{Single-view Particle Visibility }\label{alg:pos_vis}
\begin{algorithmic}[1]
\Require
$\bm u_g\!\in\!\mathbb{R}^2$: image-space projection of the Gaussian center to training view $\bm{p}$; 
$\bm \Sigma_g\!\in\!\mathbb{R}^{2\times 2}$: image-space conic (precision) matrix; 
$z_g\!\in\!\mathbb{R}$: depth (for front-to-back sorting);
$o_g\!\in\![0,1]$: base opacity; 
$\epsilon_T$: transmittance cutoff (default $10^{-4}$);
\Ensure $\{v_g\}_{g=1}^N$: isotropic particle visibility with respect to training view $\bm{p}$ 
\State Initialize $v_g\gets0$, $c_g\gets0$ for all $g$
\For{each pixel $\bm x$ in the image}
  \State $T\gets 1$
  \State $\mathcal G \gets$ Gaussians overlapping $\bm x$ (sorted by increasing $z$)
  \For{each $g\in\mathcal G$}
    \State $\alpha\gets o_g\, \exp\!\left(-(\bm u_g-\bm x)^\top \bm \Sigma_g (\bm u_g-\bm x)\right)$
    \State $v_g\gets v_g+T$;\quad $c_g\gets c_g+1$
    \State $T\gets T(1-\alpha)$
    \If{$T<\epsilon_T$} \textbf{break} \EndIf
  \EndFor
\EndFor
\For{$g=1$ to $N$}
  \If{$c_g>0$}
    \State $v_g \gets v_g / c_g$ \Comment{Average when a particle overlaps multiple pixels.}
  \EndIf
\EndFor
\State \textbf{Return} $\{v_g\}_{g=1}^N$
\end{algorithmic}
\end{algorithm}

Note that the averaging step at L15 is necessary 
when 3D Gaussian particles span multiple pixels during rasterization. Since anisotropic visibility is modeled solely as a function of viewing direction, rather than the ray-particle intersection location, similar to view-dependent color in 3DGS~\cite{kerbl20233d}, different rays from the same camera can interact with different points on the same particle, while sharing the same viewing direction. Therefore, averaging their transmittance contributions provides a consistent way to aggregate them into a single per-view value.

Next, we construct the anisotropic visibility field $V^{(i)}(\bm{d})$ by computing the SH coefficients $\gamma_{\ell m}^{(i)}$ for each particle $i$, accumulating contributions from all training views $\mathcal{P}$ via \eqref{eq:vis_gamma} as outlined in Alg.~\ref{alg:anisotropic_visibility_sh}. The single-view visibility terms $\Phi_{i,\bm{p}}, T_{\bm{p}}(t_i^{\bm{p}})$ are obtained using Alg.~\ref{alg:pos_vis}.

\begin{algorithm}[t]
\caption{Anisotropic Visibility Field Construction}
\label{alg:anisotropic_visibility_sh}
\begin{algorithmic}[1]
\Require Training views $\mathcal{P}$ with camera centers $\{\bm{x}_{\bm{p}}\}$; 3DGS particles $\mathcal{G}$ with centers $\{\bm{x}_g\}$; SH degree $L$; concentration $\kappa$; scale $\zeta=\exp(-\kappa)$
\Ensure Coefficients $\bm{\gamma}\in\mathbb{C}^{|\mathcal{G}|\times (L+1)^2}$
\State Precompute $\{i_\ell(\kappa)\}_{\ell=0}^{L}$ \Comment{Modified spherical Bessel function of the first kind}
\State $a_\ell \gets 4\pi\,\zeta\,i_\ell(\kappa)$ \textbf{for} $\ell=0,\dots,L$
\State Initialize $\gamma_{\ell m}^{(g)} \gets 0 \quad \forall g\!\in\!\mathcal{G},\, 0\!\le\!\ell\!\le\!L,\, -\ell\!\le\!m\!\le\!\ell$
\vspace{0.25em}
\For{$\bm{p}\in\mathcal{P}$} \Comment{Accumulate per-view contributions}
  \State $\bm{v} \leftarrow \textsc{SingleViewParticleVisibility}(\bm{p}, \mathcal{G})$
  \Comment{$\bm{v}[g]=\Phi_{g,\bm{p}}\,T_{\bm{p}}(t_g^{\bm{p}})$ via Alg.~\ref{alg:pos_vis}}
  \ForAll{$g \in \mathcal{G}$}
    \State $\bm{d}^{(g)}_{\bm{p}} \gets \dfrac{\bm{x}_g - \bm{x}_{\bm{p}}}{\left\|\bm{x}_g - \bm{x}_{\bm{p}}\right\|}$
    \State $\gamma_{\ell m}^{(g)} \gets \gamma_{\ell m}^{(g)} + \bm{v}[g]\cdot a_\ell \cdot Y_{\ell m}(\bm{d}^{(g)}_{\bm{p}})$ \textbf{for} $\ell=0{:}L$, $m=-\ell{:}\ell$
  \EndFor
\EndFor
\State \textbf{return} $\{\gamma_{\ell m}^{(g)}\}$
\end{algorithmic}
\end{algorithm}

Finally, we describe the algorithm for querying the anisotropic visibility field at a direction $\bm{d}$ in Alg.~\ref{alg:appx:visibility_query}. Given the stored SH coefficients ${\gamma_{\ell m}^{(g)}}$ for each particle $g$, we evaluate the SH basis at the query direction $\bm{d}$ to obtain $\tilde{V}(\bm{d})$, and then compute the visibility $V^{(g)}(\bm{d})$ using the AM-GM lower-bound estimator in \eqref{eq:vis_bound}. Because we use $L=2$ in practice, each particle requires only 9 SH coefficients, and the query cost is independent of the number of training views $|\mathcal{P}|$, enabling (per particle) constant-time querying.

\begin{algorithm}[t]
\caption{Query Anisotropic Visibility Field}
\label{alg:appx:visibility_query}
\begin{algorithmic}[1]
\Require SH coefficients $\{\gamma_{\ell m}^{(g)}\}$ for particle $g$; query direction $\bm{d}$; SH degree $L$; number of training views $|\mathcal{P}|$
\Ensure Visibility $V^{(g)}(\bm{d})$ for particle $g$ in direction $\bm{d}$
\State $y[\ell, m] \gets \textsc{SphericalHarmonics}(\ell, m, \bm{d})$ \textbf{for} $\ell=0{:}L$, $m=-\ell{:}\ell$
\State $\tilde{V} \gets \sum_{\ell=0}^{L}\sum_{m=-\ell}^{\ell} \gamma_{\ell m}^{(g)}\,y[\ell, m]$ 
\State ${V}^{(g)} \gets 1- (1-\frac{\tilde{V}}{|\mathcal{P}|})^{|\mathcal{P}|}$ \Comment{AM--GM lower-bound estimator}
\State \textbf{return} ${V}^{(g)}$
\end{algorithmic}
\end{algorithm}

Moreover, our method is significantly memory efficient. The anisotropic visibility field requires $\sim28\%$ additional parameters over standard 3DGS, including SH coefficients and virtual particles (see next subsection). Consequently, our uncertainty quantification method incurs substantially lower memory overhead compared to other 3DGS-based approaches. For example, FisherRF~\cite{jiang2023fisherrf} introduces $100\%$ additional parameters to store per-parameter uncertainty, while VIMC~\cite{lyu2024manifold} requires at least $200\%$ overhead to represent the manifold of radiance field parameters.

\subsection{Details of Visibility Field Density Control} \label{sec:appx_avf_density}

We use \emph{virtual particles} to distinguish free space from underexplored regions. 
Given a trained 3DGS, we uniformly sample virtual particles within the scene. 
Each particle is initialized with zero opacity, identity rotation, and an isotropic scale~$s$. 
The total number of virtual particles is
$$
N_{\mathrm{vp}} =  \rho \cdot \bm{V}_{\mathrm{scene}} ,
$$
where $\bm{V}_{\mathrm{scene}}$ is the scene volume and $\rho=100$ is a density hyperparameter specifying the number of virtual particles per unit volume.
The scale $s$ is chosen such that the expected total volume of virtual particles per unit volume of the scene slightly exceeds~$1$, i.e.,
$$
\frac{4}{3}\pi s^3 \rho = 1 + \eta ,
$$
with $\eta = 0.5$ ensuring that the total particle volume slightly exceeds the scene volume, which encourages broad coverage of the space. Note that $s$ is substantially larger than the scale of the original 3DGS particles. This formulation allows a sparse set of virtual particles to approximate coverage of the entire scene, making it significantly more efficient than using a dense grid of particles.

We then combine virtual particles and original particles to compute the visibility of each virtual particle using the same single-view visibility algorithm as for the original particles, with virtual particles treated as transparent during visibility computation. The multi-view visibility of each virtual particle is then computed as

\begin{equation*}
1 - \prod_{\bm{p} \in \mathcal{P}} \left(1 - \Phi_{i,\bm{p}}\, T_{\bm{p}}(t_i^{\bm{p}})\right)
\end{equation*}

Finally, we prune virtual particles using a conservative visibility threshold $\epsilon_v = 0.95$. 
Particles with visibility greater than $\epsilon_v$ are treated as free space and discarded, while those with visibility below $\epsilon_v$ are retained as indicators of underexplored regions. 

When querying the anisotropic visibility field for virtual particles, we treat all retained virtual particles as invisible in every direction, which is equivalent to setting $\gamma_{\ell m}^{(i)} = 0$ for all $\ell,m$. 
For computational efficiency, however, we simply return $V^{(i)}(\bm{d}) = 0$ for all query directions~$\bm{d}$.

Note that, unlike density control in 3DGS training~\cite{kerbl20233d}, which is applied repeatedly throughout optimization, the visibility-field density control is a post-hoc procedure and needs to be executed only once after training. This makes it substantially faster in practice. Details of the algorithm are provided in Alg.~\ref{alg:virtual_particles_density_control}.

\begin{algorithm}[h]
\caption{Visibility Field Density Control}
\label{alg:virtual_particles_density_control}
\begin{algorithmic}[1]
\Require Training views $\mathcal{P}$; original 3DGS particles $\mathcal{G}$ with centers $\{\bm{x}_g\}$, rotations $\{\bm{R}_g\}$, scales $\{s_g\}$, opacities $\{o_g\}$; scene volume $\bm{V}_{\mathrm{scene}}$; virtual particle density $\rho$;
visibility threshold $\epsilon_v$; hyperparameter $\eta$ (default $0.5$)
\Ensure Augmented 3DGS $\widetilde{\mathcal{G}} = \mathcal{G} \cup \mathcal{G}_{\mathrm{virt}}^{\mathrm{keep}}$ for uncertainty quantification
\vspace{0.25em}
\State $N_{\mathrm{virt}} \gets \left\lfloor \rho \cdot \lvert \bm{V}_{\mathrm{scene}} \rvert \right\rfloor$ \Comment{Number of virtual particles}
\State $s \gets \sqrt[3]{\dfrac{3(1+\eta)}{4\pi\rho}}$ \Comment{Scale of virtual particles}
\State Sample virtual centers $\{\bm{x}_j\}_{j=1}^{N_{\mathrm{virt}}}$ uniformly in $\bm{V}_{\mathrm{scene}}$
\State $\mathcal{G}_{\mathrm{virt}} \gets \{(\bm{x}_j, \bm{I}_3, s, 0)\}_{j=1}^{N_{\mathrm{virt}}}$ 
\Comment{Virtual particles: identity rotation, scale $s$, zero opacity}
\State $\mathcal{G}_{\mathrm{all}} \gets \mathcal{G} \cup \mathcal{G}_{\mathrm{virt}}$
\vspace{0.25em}
\State $p_g \gets 1 \quad \forall g \in \mathcal{G}_{\mathrm{virt}}$ 
\Comment{Accumulator for $\prod_{\bm{p}} (1 - \Phi_{g,\bm{p}}T_{\bm{p}}(t_g^{\bm{p}}))$}
\vspace{0.25em}
\ForAll{$\bm{p} \in \mathcal{P}$} \Comment{Accumulate per-view contributions}
  \State $\bm{v}_{\bm{p}} \gets \textsc{SingleViewParticleVisibility}(\bm{p}, \mathcal{G}_{\mathrm{all}})$
  \Comment{$\bm{v}_{\bm{p}}[g] = \Phi_{g,\bm{p}}\,T_{\bm{p}}(t_g^{\bm{p}})$ via Alg.~\ref{alg:pos_vis}}
  \ForAll{$g \in \mathcal{G}_{\mathrm{virt}}$}
    \State $p_g \gets p_g \cdot (1 - \bm{v}_{\bm{p}}[g])$
  \EndFor
\EndFor
\vspace{0.25em}
\State $\mathcal{G}_{\mathrm{virt}}^{\mathrm{keep}} \gets \emptyset$
\ForAll{$g \in \mathcal{G}_{\mathrm{virt}}$}
  \State $\tilde{v}_g \gets 1 - p_g$ 
  \Comment{Multi-view visibility $1 - \prod_{\bm{p} \in \mathcal{P}} (1 - \Phi_{g,\bm{p}}T_{\bm{p}}(t_g^{\bm{p}}))$}
  \If{$\tilde{v}_g \le \epsilon_v$}
    \Comment{Low visibility $\Rightarrow$ underexplored region}
    \State $\mathcal{G}_{\mathrm{virt}}^{\mathrm{keep}} \gets \mathcal{G}_{\mathrm{virt}}^{\mathrm{keep}} \cup \{(\bm{x}_g, \bm{I}_3, s, 0)\}$
  \Else
    \Comment{High visibility $\Rightarrow$ free space, discard}
    \State \textbf{continue}
  \EndIf
\EndFor
\vspace{0.25em}
\State $\widetilde{\mathcal{G}} \gets \mathcal{G} \cup \mathcal{G}_{\mathrm{virt}}^{\mathrm{keep}}$
\State \textbf{return} $\widetilde{\mathcal{G}}$
\end{algorithmic}
\end{algorithm}

Finally, an illustration of the two major components of GAVIS: (1) anisotropic visibility and (2) visibility-field density control, is provided in Fig.~\ref{fig:ablation_illustration}.

\section{Details of Uncertainty-aware 3DGS Rasterization} \label{sec:appx_uvr}

In this section, we provide additional details on uncertainty-aware 3DGS rasterization, specifically how to quantify the uncertainty of the observed color along a ray direction $\bm{d}$ using the anisotropic visibility field $\{V^{(g)}(\bm{d})\}_{g \in \mathcal{G}}$. This section extends the uncertainty-aware volume rendering formulation introduced in NVF~\cite{xue2024neural}, originally developed for NeRF, to 3DGS.

The rasterization process for 3DGS (along a ray $\bm{r} (t) = \bm{o} + t \bm{d}$) is expressed as:
\begin{equation}
\hat{C}(\bm{r}) = \sum_i w_i\, \bm{c}_i(\bm{d}),
\label{eq:orig_vr}
\end{equation}
\noindent where
\[
w_i = \alpha_i \prod_{j=0}^{i-1} (1 - \alpha_j),
\]
and $\alpha_i = 1 - \exp(-\sigma_i \delta_i)$ is the opacity of the $i$-th particle, with $\sigma_i$ its density and $\delta_i$ the interval length.  
In 3DGS, effective $\alpha_i$ is computed by evaluating a 2D Gaussian with covariance $\Sigma_i$ at the pixel location, multiplied by the learned per-particle opacity $o_i$, where $\Sigma_i$ is obtained by projecting the 3D Gaussian onto the image plane~\cite{kerbl20233d}.  
The emitted color $\bm{c}_i(\bm{d})$ is obtained from the spherical harmonics coefficients of particle $i$, evaluated at direction $\bm{d}$.

To extend \eqref{eq:orig_vr} to uncertainty-aware rasterization, consider computing the posterior distribution of the observed color $\hat{C}(\bm{r})$ along direction $\bm{d}$, denoted as $p(\bm{z}_0)$. Each emitted color is modeled as a Gaussian random variable $\bm{c}_i(\bm{d}) \sim \mathcal{N}(\bm{\mu}_{\bm{c}_i}, \bm{Q}_{\bm{c}_i})$, where $\bm{\mu}_{\bm{c}_i}$ and $\bm{Q}_{\bm{c}_i}$ are respectively the mean and covariance. Since density $\sigma_i$ can be interpreted as the differential probability that a ray terminates at position $i$~\cite{mildenhall2021nerf}, the effective opacity $\alpha_i$ represents the probability of ray terminating at that position. The rasterization process can thus be expressed as a Bayesian Network~\cite{xue2024neural} via the recursion
\begin{equation}
p(\bm{z}_i) = \alpha_i\, p(\bm{c}_i) + (1-\alpha_i)\, p(\bm{z}_{i+1}),
\label{eq:recursive}
\end{equation}
\noindent where $p(\bm{z}_i)$ denotes the distribution of the observed color if the camera were placed at position $i$ along the ray. Equation~\eqref{eq:recursive} states that with probability $\alpha_i$ a ray terminates at particle $i$, in which case the observed color is $\bm{c}_i$ ; otherwise, with probability $1-\alpha_i$, the observed color is inherited from the next position along the ray. The recursion terminates at the last particle, whose emitted color distribution is Gaussian.

By applying the recursive expression in \eqref{eq:recursive} from the last particle to the first, we can compute the posterior of the observed color at the camera position $\bm{z}_0$, which is the output of the rasterization process. 
\begin{equation} \label{eq:gmm_naive}
 p(\bm{z}_0) = \sum_i w_i \mathcal{N}(\bm{\mu}_{\bm{c}_i}, \bm{Q}_{\bm{c}_i}), 
\end{equation} 
\noindent which is a Gaussian mixture model (GMM) with $N$ (number of particles along the ray) components.

As noted in NVF~\cite{xue2024neural}, directly quantifying uncertainty from the GMM in~\eqref{eq:gmm_naive} is limited because the predicted color variance $\bm{Q}_{\bm{c}_i}$ is often inaccurate due to approximations in the uncertainty estimation process. To address this issue, NVF~\cite{xue2024neural} introduces a (isotropic) visibility-modulated variance correction which ensures particles with lower visibility exhibit higher uncertainty.  

We extend this correction to anisotropic visibility and 3DGS. Let $V_i$ be a binary random variable indicating whether particle $i$ is visible along the ray, with visibility probability given by the visibility field, $P(V_i = 1) = V^{(i)}(\bm{d})$. The distribution of the emitted color is therefore modified to
\begin{equation}
p(\bm{c}_i \mid V_i) =
\begin{cases}
\mathcal{N}(\bm{\mu}_{\bm{c}_i},\, \bm{Q}_{\bm{c}_i}), & \text{if } V_i = 1, \\[4pt]
\mathcal{N}(\bm{\mu}_0,\, \bm{Q}_0), & \text{otherwise},
\end{cases}
\label{eq:pcv}
\end{equation}
\noindent where $\mathcal{N}(\bm{\mu}_0, \bm{Q}_0)$ is a prior with large variance $\bm{Q}_0$ to reflect the high uncertainty of invisible particles.

Opacity is also visibility-modulated, since opacity predictions are unreliable for particles that are rarely or never observed. Similar to~\cite{xue2024neural}, the compensated opacity for 3DGS is
\begin{equation} \label{eq:alpha_comp}
\alpha_i^* = (v_i + \beta (1 - v_i))\, \alpha_i 
+ o_0 (1 - \beta)\, (1 - v_i),
\end{equation}
\noindent where $v_i := V^{(i)}(\bm{d})$ for shorthand, $o_0$ is the prior opacity for invisible particles, and $\beta$ is a hyperparameter representing the reliability of the opacity prediction for invisible regions. The first term in~\eqref{eq:alpha_comp} captures the case where the particle is visible or the opacity prediction is reliable. The second term introduces a correction when the particle is invisible and the predicted opacity is unreliable, by substituting the predicted opacity in place of a prior opacity $o_0$. Note that here we use the per-particle opacity $o_0$ instead of the effective opacity that incorporates the Gaussian decay as a function of the distance between the particle center and the pixel location.

Therefore, the visibility-modulated uncertainty-aware 3DGS rasterization process can be expressed as the following Gaussian mixture model:
\begin{equation}
p(\bm{z}_0) = 
\sum_i w_i^*\, v_i\, \mathcal{N}\!\big(\bm{\mu}_{\bm{c}_i},\, \bm{Q}_{\bm{c}_i}\big)
\;+\;
\mathcal{N}\!\big(\bm{\mu}_{0},\, \bm{Q}_{0}\big)\,\sum_i w_i^* (1 - v_i),
\label{eq:appx:gmm}
\end{equation}
\noindent where $w_i^* = \alpha_i^* \prod_{j=1}^{i-1} (1 - \alpha_j^*)$ denotes the visibility-compensated weights.

The entropy of the GMM in~\eqref{eq:appx:gmm} is then used to quantify the uncertainty of the observed color $\bm{z}_0$. Because the entropy of a Gaussian mixture lacks a closed-form expression, the following upper bound~\cite{huber2008entropy}, also adopted in NVF~\cite{xue2024neural}, is used as an accurate estimator of the entropy:
\begin{equation}
\mathcal{H}(\bm{z}_0)
\;\le\;
\sum_{i=0}^{N} 
\bar{w}_i \left(
 - \log (\bar{w}_i)
 + \frac{1}{2} \log\!\big( (2\pi e)^D\, |\bm{Q}_i| \big)
\right),
\label{eq:huber}
\end{equation}
\noindent where we denote $\bar{w}_i:=w_i^* v_i$ for $i\geq 1$, and $\bar{w}_0:=\sum_{i=1}^{N} w_i^* (1 - v_i)$, and $\bm{Q}_i := \bm{Q}_{\bm{c}_i}$ for $i\geq 1$ and $\bm{Q}_0$ for $i=0$. Here, $D = 3$ corresponds to the RGB color channels.

It is worth noting that the visibility field $v_i$ is estimated using the lower bound in~\eqref{eq:appx:vis_bound_final}. 
This this approximation underestimates visibility and therefore overestimates how much of the mixture should be attributed to invisible regions.
Consequently, the RHS in~\eqref{eq:huber} remains a valid upper bound on the entropy computed with exact visibility, since the inequality is preserved under the assumption $\bm{Q}_0 \succeq \bm{Q}_{\bm{c}_i}$ for all $i$. This condition is enforced by construction, as $\bm{Q}_0$ is chosen to be a large-variance prior.

As demonstrated in~\cite{xue2024neural}, when using visibility-modulated uncertainty-aware volume rendering, setting $\bm{Q}_{\bm{c}_i}$ to a constant already yields strong performance in active mapping, since the visibility field effectively captures uncertainty in unseen regions, which is the dominant factor for active mapping. Incorporating predicted variances $\bm{Q}_{\bm{c}_i}$ yields only marginal performance gains.

Motivated by this observation, our base GAVIS model adopts a constant variance $\bm{Q}_{\bm{c}_i} = \sigma_c^2 \mathbf{I}$, which offers a favorable trade-off between performance and computational cost. In addition, existing uncertainty quantification methods for 3DGS, including FisherRF~\cite{jiang2023fisherrf} and VIMC~\cite{lyu2024manifold}, could be used to predict $\bm{Q}_{\bm{c}_i}$, leading to two variants of the base GAVIS model: (1) FisherRF+GAVIS and (2) VIMC+GAVIS. Further details on how these methods are used to obtain $\bm{Q}_{\bm{c}_i}$ are provided in the Sec.~\ref{sec:appx_posthoc}, and quantitative experiments of the two variants are included in the Sec.~\ref{sec:appx_result}, demonstrating that GAVIS can serve as a post-hoc module that significantly improves the performance of existing methods.

Finally, we describe the algorithm for 3DGS uncertainty quantification using the visibility field at a synthesis view $\bm{p}$ in Alg.~\ref{alg:uq}. This procedure is efficiently implemented using a modified 3DGS rasterizer, enabling real-time uncertainty quantification at around 200 FPS.

\begin{algorithm}
\caption{3DGS Uncertainty Quantification with Visibility Field}\label{alg:uq}
\begin{algorithmic}[1]
\Require
$\bm u_g\!\in\!\mathbb{R}^2$: image-space projection of the Gaussian center to synthesis view $\bm{p}$; 
$\bm \Sigma_g\!\in\!\mathbb{R}^{2\times 2}$: image-space conic (precision) matrix; 
$z_g\!\in\!\mathbb{R}$: depth (for front-to-back sorting);
$o_g\!\in\![0,1]$: base opacity; 
$v_g\!\in\![0,1]$: queried visibility field value for Gaussian $g$ along direction $\bm{d}_p$; 
$\bm{Q}_{\bm{c}_g}$: predicted color covariance for Gaussian $g$ along direction $\bm{d}_p$; 
$\epsilon_T$: transmittance cutoff (default $10^{-4}$);
$\beta$, $o_0$, $\bm{Q}_0$: hyperparameters for visibility compensation (default $\beta=0.5$, $o_0=0.15$, $\bm{Q}_0=\bm{I}_3$).
\Ensure Entropy map $\mathcal{H}\!\in\!\mathbb{R}^{H \times W}$ for synthesis view $\bm p$.
\State Initialize $\bm{H}[\bm{x}]\gets0$ for all pixels $\bm{x}$ in the image
\For{each pixel $\bm x$ in the image}
  \State $T\gets 1$ \Comment{current transmittance}
  \State $\bar{w}_0 \gets 0$ \Comment{accumulated weight of the invisible component}
  \State $\mathcal G \gets$ Gaussians overlapping $\bm x$ (sorted by increasing $z$)
  \For{each $g\in\mathcal G$}
    \State $\alpha\gets o_g\, \exp\!\left(-(\bm u_g-\bm x)^\top \bm \Sigma_g (\bm u_g-\bm x)\right)$
    \State $\alpha^*\gets (v_g + \beta (1 - v_g))\, \alpha 
+ o_0\, (1 - \beta)\, (1 - v_g)$ \Comment{Eq.~\eqref{eq:alpha_comp}}
    \State $w^* \gets \alpha^* T$

     \State $h \gets -\log (w^* v_g)  + \tfrac12 \log|\bm{Q}_{\bm{c}_g}| + \tfrac{3}{2}\log(2\pi e)$
     \State $\bm{H}[\bm x] \gets \bm{H}[\bm x] + w^*\, v_g\, h$ \Comment{Eq.~\eqref{eq:huber}}   %
    \State $\bar{w}_0 \gets \bar{w}_0 + w^* (1 - v_g)$
    \State $T\gets T(1-\alpha^*)$
    \If{$T<\epsilon_T$} \textbf{break} \EndIf
  \EndFor
  \State $h \gets -\log \bar{w}_0  + \tfrac12 \log|\bm{Q}_0| + \tfrac{3}{2}\log(2\pi e)$
  \State $\bm{H}[\bm x] \gets \bm{H}[\bm x] + \bar{w}_0\, h$ \Comment{Add the contribution of the invisible component}
\EndFor
\State \textbf{Return} $\bm{H}$
\end{algorithmic}
\end{algorithm}

\section{Active Mapping Details} \label{sec:appx_am}

In this section, we provide additional details on our active mapping method, which follows a formulation similar to that in~\cite{xue2024neural}.

The goal of active mapping or next-best-view planning is to select the camera pose $\bm{p}$ that maximizes the expected information gain (EIG) about the scene, which can be written as:
\begin{equation}
I(\bm{Z}; \bm{\theta}\mid \bm{p})
=
\mathcal{H}[\bm{Z}\mid \bm{p}]
-
\mathcal{H}[\bm{Z}_{\bm{p}} \mid \bm{\theta}, \bm{p}],
\end{equation}
where $I(\cdot)$ denotes mutual information, $\mathcal{H}[\cdot]$ is entropy, $\bm{Z}$ is the random variable corresponding to the rendered observation, and $\bm{\theta}$ represents the radiance-field parameters.

We assume a Gaussian observation model,
$
p(\bm{Z} \mid \bm{p}, \bm{\theta})
=
\mathcal{N}\!\left(f(\bm{p}; \bm{\theta}),\, \bm{\Sigma}_z\right),
$
where $f(\bm{p}; \bm{\theta})$ is the rendering function and $\bm{\Sigma}_z$ is the observation noise covariance. 
Assuming $\bm{\Sigma}_z$ is constant and isotropic, i.e.,
$\bm{\Sigma}_z = \sigma_z^2 \mathbf{I}$
then the entropy term $\mathcal{H}[\bm{Z}\mid \bm{\theta}, \bm{p}]$ becomes constant and can be omitted during optimization. 
Hence, maximizing EIG reduces to maximizing the entropy:
\begin{equation}
\bm{p}^* 
= 
\argmax_{\bm{p}}
\; \mathcal{H}\!\left(\bm{Z}_{\bm{p}}\right),
\label{eq:appx:i_gain}
\end{equation}
where we use $\mathcal{H}\!\left(\bm{Z}_{\bm{p}}\right)$ as shorthand for $\mathcal{H}[\bm{Z}_{\bm{p}} \mid \bm{p}]$, which could be obtained from the posterior of the rendered observation. This objective is commonly adopted in radiance-field-based active mapping~\cite{lee2022uncertainty,yan2023active,xue2024neural,lyu2024manifold}. Note that uncertainties arising from robot localization or dynamics are not modeled here, as the primary focus of these methods is the uncertainty of the radiance field itself. Moreover, the action $\,\bm{\tau}\,$ is restricted to the next camera pose, and the EIG of an entire trajectory is not considered, since efficiently computing the joint entropy over a sequence of observations remains an open challenge~\cite{kirsch2022unifying,placed2023survey}. A simplified workaround assumes independence across views~\cite{jiang2024ag,tao2025rt}, though this approximation is not the focus of this work. Extending the GAVIS framework to support trajectory-level EIG is a promising direction for future research, but it lies beyond the scope of this paper.

We adopt the spatial correlation correction term from NVF~\cite{xue2024neural} to recover the joint image-level entropy from per-pixel entropies:
\begin{equation}
\mathcal{H}(\bm{Z_p}) = \sum_{m,n} \Big( \mathcal{H}(\bm{Z_p}^{mn}) - f_{\rm corr}\!\big(\mathcal{H}(\bm{Z_p}^{mn});\, d_{\bm{p}}^{mn}\big) \Big),
\label{eq:id}
\end{equation}
\noindent where $\bm{Z_p}^{mn}$ denotes the observed color associated with pixel $(m,n)$, $d_{\bm{p}}^{mn}$ is its expected depth, and $f_{\rm corr}(\cdot)$ is a depth-dependent correction term that compensates for spatial correlations between neighboring pixels. Additional details can be found in~\cite{xue2024neural}.

Finally, we describe the active mapping algorithm with GAVIS in Alg.~\ref{alg:active_mapping}. The algorithm iteratively trains the 3DGS model, constructs the visibility field, and selects the next-best view based on the expected information gain computed from the visibility-aware 3DGS uncertainty quantification.

\begin{algorithm}
\caption{Active Mapping with GAVIS}\label{alg:active_mapping}
\begin{algorithmic}[1]

\Require 
$\mathcal{P}$: initial poses;
$\bm{Z}$: initial images;
$T$: number of active mapping planning steps.
\Ensure
$\mathcal{G}$: the trained 3DGS model after active mapping.

\For{$i=1$ to $T$}
\State $\mathcal{G} \gets$ train3DGS($\mathcal{P}, \bm{Z}$) \Comment{train 3DGS}
\State $\{\gamma_{\ell m}^{(g)}\} \gets$ VFConstruction($\mathcal{G}$, $\mathcal{P}$) \Comment{Alg.~\ref{alg:anisotropic_visibility_sh}}
\State $\widetilde{\mathcal{G}} \gets$ VFDensityControl($\mathcal{G}$, $\mathcal{P}$) \Comment{Alg.~\ref{alg:virtual_particles_density_control}}

\State $\mathcal{P}_{c} \gets\text{samplePoses}(\mathcal{G}) $ \Comment{sample candidate poses} %
\For{$\bm{p}$ in $\mathcal{P}_{c}$}
\State $v_g$ $\gets$ VFQuery($\widetilde{\mathcal{G}}$, $\{\gamma_{\ell m}^{(g)}\}$, $\bm{p}$) \Comment{Alg.~\ref{alg:appx:visibility_query}}
\State $\bm{Z_p}$ $\gets$ UQVF($\widetilde{\mathcal{G}}$, $v_g$ , $\bm{p}$) \Comment{Alg.~\ref{alg:uq}}

\EndFor
\State $\bm{p}_i \gets \argmax\limits_{\bm{p} \in \mathcal{P}_{c}}\mathcal{H} (\bm{Z_p}) $
\State $\mathcal{P} \gets \{\bm{p}_i\} \cup \mathcal{P}$
\State $\bm{Z} \gets \text{takeImageAt($\{\bm{p}_i\}$)} \cup \bm{Z}$ \Comment{update training set} %

\EndFor

\State \textbf{return} $\mathcal{G}$
\end{algorithmic}
\end{algorithm}

\section{Applying GAVIS as a Post-hoc Module} \label{sec:appx_posthoc}

In this section, we provide additional details on applying GAVIS as a post-hoc module to enhance existing uncertainty quantification methods, including FisherRF~\cite{jiang2023fisherrf} and VIMC~\cite{lyu2024manifold}. 
The key idea is to use the uncertainty estimated by FisherRF or VIMC to refine the variance of emitted color  $\bm{Q}_{\bm{c}_i}$ in \eqref{eq:gmm} along ray direction $\bm{d}$, whereas in base GAVIS this quantity is treated as a constant hyperparameter for all particles.
We first describe how to integrate GAVIS with FisherRF and VIMC, and then discuss the relationship between GAVIS and these methods.

\subsection{Integration with FisherRF} \label{sec:appx_posthoc_fisherrf}

FisherRF~\cite{jiang2023fisherrf} approximates the expected information gain of a candidate camera pose by maximizing
\begin{equation}
    \argmax_{\bm{p}} \;
    \tr\!\left(
        \mathbf{H}''[\bm{Z_p} \mid \bm{p}, \bm{\theta}^*] \,
        \mathbf{H}''[\bm{\theta}^* \mid D^{\text{train}}]^{-1}
    \right),
    \label{eq:fisher-obj}
\end{equation}
where the observation $\bm{Z_p}$ is the rendered image for pose $\bm{p}$, $\bm{\theta}^*$ are the estimated 3DGS parameters, and the training set is $D^{\text{train}} := \{(\bm{p}, \bm{Z}_{\bm{p}})\}_{\bm{p} \in \mathcal{P}}$. The observed information is given by:
\begin{equation}
    \mathbf{H}''[\bm{\theta}^* \mid D^{\text{train}}]
    =
    \sum_{(\bm{p}, \bm{Z}_{\bm{p}}) \in D^{\text{train}}}
    \mathbf{H}''[\bm{Z}_{\bm{p}} \mid \bm{p}, \bm{\theta}^*],
\end{equation}
where $\mathbf{H}''[\bm{Z}_{\bm{p}} \mid \bm{p}, \bm{\theta}^*]$ is the Hessian of the negative log-likelihood under the Gaussian observation model
\begin{equation} \small
    -\log p(\bm{Z}_{\bm{p}} \mid \bm{p}, \bm{\theta})
    =
    \frac{1}{2 \sigma_z^2}
    \bigl(\bm{Z}_{\bm{p}} - f(\bm{p}; \bm{\theta})\bigr)^\top
    \bigl(\bm{Z}_{\bm{p}} - f(\bm{p}; \bm{\theta})\bigr)
    + C,
\end{equation}
with rendering model $f(\bm{p}; \bm{\theta})$ and observation noise variance $\sigma_z^2$. FisherRF adopts the Laplace approximation~\cite{kirsch2022unifying}, leading to:

\begin{equation}
    \mathbf{H}''[\bm{Z} | \bm{p}, \bm{\theta}^*] \approx \frac{1}{\sigma_z^2}\text{diag} \left( \nabla_{\bm{\theta}} f(\bm{p}; \bm{\theta})^T \nabla_{\bm{\theta}} f(\bm{p}; \bm{\theta}) \right) + \lambda \bm{I}
\end{equation}

Thus the posterior covariance of $\bm{\theta}$ is given by:
\begin{equation}
\Sigma_{\bm{\theta}} \approx \mathbf{H}''[\bm{\theta}^*| D^{\text{train}}]^{-1}
\end{equation}

\noindent Further details on these derivations can be found in \cite{jiang2023fisherrf,goli2023bayes,kirsch2022unifying}.

We then propagate this parameter uncertainty to the emitted color of each particle. Let $\theta_{\ell,m,k}^{(i)}$ denote the color coefficient of the $i$-th particle in channel $k$, parameterized in the spherical harmonics basis. The emitted color in channel $k$ along direction $\bm{d}$ is given by
\begin{equation} \label{eq:rgb_sh}
    c_{i,k}(\bm{d})
    =
    \sum_{\ell=0}^{L} \sum_{m=-\ell}^{\ell}
    \theta_{\ell,m,k}^{(i)} \,
    Y_{\ell,m}(\bm{d}),
\end{equation}
which is linear in the coefficients $\theta_{\ell,m,k}^{(i)}$. 
The variance of the emitted color in channel $k$ is obtained by linear uncertainty propagation:
\begin{equation}
    Q_{c_{i,k}}(\bm{d})
    =
    \sum_{\ell=0}^{L} \sum_{m=-\ell}^{\ell}
    Y_{\ell,m}(\bm{d})^2 \,
    \Sigma_{\ell,m,k}^{(i)},
\end{equation}
where $\Sigma_{\ell,m,k}^{(i)}$ is the variance of the coefficient $\theta_{\ell,m,k}^{(i)}$ extracted from the diagonal of $\Sigma_{\bm{\theta}}$, as in FisherRF~\cite{jiang2023fisherrf} all parameters are assumed independent and the covariance matrix $\Sigma_{\bm{\theta}}$ is therefore diagonal.
Since each color channel is also modeled independently, the covariance of the emitted color vector $\bm{Q}_{\bm{c}_i}$ is diagonal with diagonal entries $Q_{c_{i,k}}$.

\subsection{Integration with VIMC} \label{sec:appx_posthoc_vimc}

VIMC~\cite{lyu2024manifold} estimates the distribution of $\bm{\theta}$ from a learned low-dimensional manifold of the radiance field parameters.
We draw $J$ parameter samples (with $J=2$ by default in VIMC). 
For each sampled parameter $\bm{\theta}_j$, we compute the emitted color $c_{i,k,j}(\bm{d})$ following~\eqref{eq:rgb_sh}, and estimate the per-channel uncertainty by taking the sample variance across the $J$ samples:
\begin{equation}
    Q_{c_{i,k}}(\bm{d})
    =
    \frac{1}{J-1} \sum_{j=1}^J \left(c_{i,k,j}(\bm{d}) - \bar{c}_{i,k}(\bm{d})\right)^2,
\end{equation}
where $\bar{c}_{i,k}(\bm{d}) = \frac{1}{J} \sum_{j=1}^J c_{i,k,j}(\bm{d})$ is the sample mean. 
Under the assumption that the three color channels are independent, the resulting emitted-color covariance $\bm{Q}_{\bm{c}_i}$ similarly reduces to a diagonal matrix whose diagonal entries are given by $Q_{c_{i,k}}(\bm{d})$.

\subsection{Discussion} \label{sec:appx_posthoc_discussion}

In principle, an accurate radiance-field uncertainty estimation method should assign high uncertainty to unseen regions, as these areas do not contribute to the training loss. 
However, existing approaches struggle to capture this behavior. 

In FisherRF, the diagonal Laplace approximation considers only the diagonal entries of the Hessian. 
This ignores the strong correlations among parameters of the same Gaussian particle (e.g., RGB spherical-harmonics coefficients within a color channel), causing the approximation to underestimate uncertainty, especially for viewing directions far from the training views. 

In VIMC, uncertainty is approximated from Monte Carlo samples drawn from a learned low-dimensional manifold. 
With only a small number of samples, variance estimates are noisy and unreliable, while increasing the sample count significantly raises computational cost, making it unsuitable for real-time applications.

Integrating GAVIS provides an efficient way to compensate for these limitations. 
Since GAVIS reliably assigns higher uncertainty to unseen regions, it improves the accuracy of uncertainty quantification and leads to better active-mapping performance (see Sec.~\ref{sec:appx_result}).

\section{Experimental Details} \label{sec:appx_exp}

\renewcommand\meanstd[2]{%
  \ensuremath{#1{\mskip2mu{\scriptstyle\pm}\mskip1mu{\scriptscriptstyle #2}}}%
}

\subsection{Environment Setup} \label{sec:appx_exp:setup}
We evaluate our method on four datasets: (1) the standard NeRF Synthetic dataset~\cite{mildenhall2021nerf}; (2) a space dataset consisting of the Hubble Space Telescope (HST)~\cite{nasa_hubble_3d_model} and the International Space Station (ISS) ~\cite{nasa_iss_3d_model} for space robotics scenarios; (3) eight indoor environments from the Habitat-Matterport 3D (HM3D)~\cite{ramakrishnan2021hm3d} dataset 
\begin{itemize}
    \item \texttt{00208-SQqGpSHzfSr}
    \item \texttt{00299-bdp1XNEdvmW}
    \item \texttt{00321-JWWJBQWHv64}
    \item \texttt{00323-yHLr6bvWsVm}
    \item \texttt{00415-rBmEe6ab5VP}
    \item \texttt{00441-4MRLu1yET6a}
    \item \texttt{00446-tL6i2PtktSh}
    \item \texttt{00670-mDdyQ6azhVD}
\end{itemize}
\noindent and (4) eight scenes from the Gibson~\cite{xia2018gibson} dataset (\texttt{annawan}, \texttt{bremerton}, \texttt{creede},
\texttt{eagerville}, \texttt{eastville}, \texttt{helix}, \texttt{hometown}, \texttt{quantico}). NeRF Synthetic and the space dataset are simulated in Blender, whereas HM3D and Gibson scenes are simulated in Habitat-Sim~\cite{savva2019habitat}. All images are rendered at a resolution of $512 \times 512$ pixels, with both horizontal and vertical fields of view set to $90^\circ$. 
For a fair comparison, all methods are initialized from the same set of closely spaced initial views that cover only a portion of the scene, mimicking the starting condition of a robotic active mapping process. The agent begins with $3$ initial views for NeRF Synthetic and the space dataset, and with $5$ spherically sampled views from a fixed starting position for HM3D and Gibson. We use $10$ active-mapping steps for NeRF Synthetic and the space dataset, $40$ steps for Gibson, and $80$ steps for HM3D, chosen as the minimal steps at which the strongest methods achieve acceptable reconstruction quality ($\text{PSNR} > 24$). Each scene is evaluated over three runs with different random seeds, and the mean and propagated standard error due to seed stochasticity \cite{taylor1994guidelines} of all metrics are reported. All experiments for each scene and each method are performed on a single NVIDIA A40 GPU.

\subsection{Active Mapping Setup}  \label{sec:appx_exp:am}
To ensure a fair comparison of uncertainty quantification methods for active mapping, all methods are evaluated using the same active mapping pipeline, differing only in the uncertainty quantification module, following the setup in NVF~\cite{xue2024neural}. In particular, each method uses the same minimally biased candidate view sampler, which samples only collision-free candidate poses without additional heuristics. This design reflects the principle that an optimal active mapping policy should, in principle, be achievable by relying solely on accurate uncertainty quantification, without the need for heuristic guidance. For the NeRF Synthetic and space datasets, candidate poses are sampled from $\mathrm{SE}(3)$ within the scene bounding box. For the Gibson and HM3D datasets, candidate poses are sampled from $\mathrm{SE}(2)$ on a plane at positions known to be collision-free, with a fixed height of $1.5\,\mathrm{m}$, mimicking the active mapping process of a mobile robot.

\subsection{Training Details} \label{sec:appx_exp:train}
For a fair comparison, we adopt the standard 3DGS training setup from~\cite{kerbl20233d} with minimal differences across all 3DGS-based methods. Since FisherRF and GAVIS are post-hoc methods, they share the exact same 3DGS training setup. VIMC uses the same setup as well, with the only difference being an additional loss term used to learn the low-dimensional manifold in the space of 3DGS parameters for uncertainty quantification. NVF is trained following the same configuration as in~\cite{xue2024neural}. 
For the NeRF Synthetic and space datasets, 3DGS is trained for $3500$ iterations at each planning step. For the Gibson and HM3D indoor datasets, 3DGS is trained for $7500$ iterations at each planning step, and depth supervision from~\cite{matsuki2024gaussian} is incorporated to accelerate convergence for all 3DGS-based methods, using simulator-rendered depth images. For a fair comparison, a similar depth loss~\cite{deng2022depth} is also applied when training NVF.
For shorter-horizon datasets (NeRF Synthetic, space, and Gibson), 3DGS is initialized using a subsampled sparse point cloud constructed from all previously observed depth images, to reduce training time. For the longer-horizon HM3D dataset, 3DGS at each planning step is initialized by merging the previous 3DGS model with a subsampled sparse point cloud derived from the most recent depth image.

\subsection{Metrics Details} \label{sec:appx_exp:metric}
\paragraph{Visual Coverage.}
For visual coverage evaluation, we adopt the ground-truth visibility (VIS) metric from~\cite{xue2024neural}. This metric assigns each face in the ground-truth mesh a binary visibility value, all faces are initialized with value 0, and then the value of a face is set to $1$ (visible) if it is observed by the training views without occlusion. The Vis score is computed as the ratio of the total area of visible faces to the total mesh surface area.

We also visualize the GT visibility map (used in Figs.~\ref{fig:en_room}, \ref{fig:enobjs}, and~\ref{fig:enroom}) by rendering the visibility-annotated mesh from each candidate query pose. Similar to the uncertainty map, darker regions indicate visible areas (value 1), while brighter regions indicate invisible areas (value 0). Ideally, an uncertainty quantification method tailored for active mapping should produce uncertainty maps that correlate with the GT visibility map, assigning high uncertainty (bright) to invisible regions and low uncertainty (dark) to visible regions.

We note that ground-truth (GT) visibility is inherently an isotropic metric that does not account for the view-dependent nature of visibility. In particular, observing a face from a single direction, regardless of viewing distance, may already be considered fully visible. As a result, such metrics do not favor uncertainty quantification methods that encourage revisiting regions from diverse viewpoints. Designing evaluation metrics for visual coverage that account for viewpoint diversity and better align with the active mapping's objective of acquiring observations from diverse directions remains an open problem.

\paragraph{Mesh Metrics.}
To assess reconstruction completeness, we also include the completion ratio defined in~\cite{Sucar:etal:ICCV2021}. Ground-truth 3D points are sampled from the original scene meshes. Predicted points are extracted from a trained 3DGS model using the approach described in~\cite{ag20253dgs}. For consistency with prior work~\cite{xue2024neural}, we use a completion ratio threshold of $0.01$ for the NeRF Synthetic and space datasets. For HM3D and Gibson environments, we adopt a threshold of $0.05$.

We additionally report mesh-based metrics, including completion (Comp) and accuracy (Acc). However, we adopt completion ratio (CR) as the primary mesh metric, as it is widely used, easier to interpret, and more robust to outliers compared to Comp~\cite{jin2023neu,li2025activesplat}. Moreover, Acc is not well aligned with the objective of active mapping under 3DGS, despite its use in NeRF-based active mapping~\cite{xue2024neural}. This is because 3DGS cannot distinguish empty space from underexplored regions, causing reconstructed point clouds to reflect only explored areas. As a result, Acc is biased toward policies that oversample a limited region while penalizing broader exploration in 3DGS active mapping.
Empirically, Acc shows minimal improvement with additional views and exhibits no statistically significant differences across methods; accordingly, it is rarely used in 3DGS-based active mapping~\cite{jiang2023fisherrf,lyu2024manifold}. As shown in Tab.~\ref{tab:appx:am}, Acc varies little across methods and shows weak correlation with other metrics.

\paragraph{Uncertainty Quantification.}
To quantitatively evaluate the quality of uncertainty quantification, we adopt the Area Under the Sparsification Error curve (AUSE) metric~\cite{ilg2018uncertainty,poggi2020uncertainty}, which measures the correlation between predicted uncertainty and actual error. 
However, the depth-based AUSE metric (AUSE-D) used in~\cite{jiang2023fisherrf,lyu2024manifold,goli2023bayes}, which evaluates the correlation between predicted uncertainty and depth error computed from the predicted depth and the ground-truth mesh, can be misaligned with the objective of active mapping. 
In particular, radiance fields can produce accurate depth predictions in unobserved regions via plausible imagination. As a result, generic uncertainty quantification methods, which are not designed for active mapping and tend to assign low uncertainty to these regions, can still achieve strong AUSE-D scores.
In contrast, effective active mapping requires high uncertainty in unobserved regions to encourage exploration. 
Therefore, a strong AUSE-D score does not necessarily indicate good active mapping performance.

We note that alternative metrics for evaluating uncertainty quantification exist, including negative log-likelihood (NLL) and expected calibration error (ECE)~\cite{guo2017calibration}. However, these metrics are less commonly used in the context of radiance fields, since they emphasize the quality of a fully specified and calibrated predictive distribution, which is not necessary for downstream tasks such as active mapping and artifact removal~\cite{goli2023bayes}. Moreover, it is unclear how to fairly compare these metrics across different methods, for example, GMM-based approaches (GAVIS\&NVF), methods assuming a Gaussian distribution (VIMC), and methods that do not explicitly predict a distribution (FisherRF). Furthermore, similar to AUSE-D, these metrics can also be misaligned with the objective of active mapping.

Therefore, designing uncertainty quantification metrics tailored to active mapping remains an open problem. Toward this direction, we introduce a ground-truth visibility-based variant, denoted as AUSE-V, which measures the correlation between the predicted uncertainty map and the ground-truth visibility map. 
The ground-truth visibility map is obtained by assigning binary face-level values on the ground-truth mesh and rendering them using a rasterizer. AUSE-V better captures the requirement of active mapping for high uncertainty in unobserved regions, and thus exhibits stronger correlation with both active mapping performance and qualitative behavior. Further results are provided in Sec.~\ref{sec:appx_quant}.

However, as discussed above, GT visibility is inherently isotropic and does not account for the view-dependent nature of visibility, so it may not fully capture the benefits of anisotropic visibility modeling. Moreover, a reliable uncertainty quantification method for active mapping should assign high uncertainty to regions with low GT visibility, but need not always assign low uncertainty to regions with high GT visibility. High uncertainty in observed regions may arise when the test view direction deviates substantially from the training views direction or when model capacity is limited. As a result, predictions can still be inaccurate even in observed regions. Therefore, although AUSE-V is more informative than AUSE-D, it may not fully capture uncertainty quality for active mapping.

\section{Additional Results} \label{sec:appx_result}

\subsection{Additional Quantitative Results} \label{sec:appx_quant}

We present the complete active mapping results for all methods, including the baselines FisherRF, VIMC, and NVF, as well as our proposed methods GAVIS, FisherRF+GAVIS, and VIMC+GAVIS, with all evaluation metrics reported for each dataset in Tab.~\ref{tab:appx:am}, with additional Comp and Acc metrics included.

As shown in Tab.~\ref{tab:appx:am}, our method significantly outperforms all 3DGS baselines in all metrics excluding Acc. Moreover, integrating GAVIS as a post-hoc module consistently yields substantial improvements over each corresponding baseline across all datasets and all metrics excluding Acc. Our method also surpasses NVF in active mapping performance while requiring substantially less computation time. GAVIS outperforms NVF on all image-based metrics (PSNR, SSIM, and LPIPS) across all datasets. 
On the primary mesh metric (CR) and visibility metric (VIS), GAVIS outperforms NVF on all datasets except Gibson. Since these metrics are inherently isotropic and direction agnostic, e.g., observing a region from a single direction may already provide high GT mesh visibility and low point-cloud error (especially when depth loss is used), the performance gains are less pronounced on these metrics.

Among the variants of our method (GAVIS, FisherRF+GAVIS, and VIMC+GAVIS), FisherRF+GAVIS overall outperforms the base GAVIS model on most datasets except the space dataset. This suggests that more accurately modeling $\bm{Q}_{\bm{c}_i}$ can further improve performance. The advantage is less pronounced in the space dataset, since color variance tends to be more monotonic across these scenes. 
VIMC+GAVIS outperforms the base GAVIS model only on the NeRF Synthetic dataset and not on the others. This is because VIMC relies on sampling-based uncertainty estimation, and with only 2 samples (the default setting in~\cite{lyu2024manifold}), the sampling noise tends to overshadow the benefits introduced by visibility modeling.

Although performance gains can be achieved by more accurately modeling $\bm{Q}_{\bm{c}_i}$, using a constant covariance already yields superior performance, as visibility is the dominant factor in active mapping, which is consistent with the findings in~\cite{xue2024neural}. Since obtaining more accurate $\bm{Q}_{\bm{c}_i}$ with FisherRF+GAVIS or VIMC+GAVIS requires additional processing with substantially higher computational cost, the base GAVIS model provides a more favorable trade-off between performance and efficiency in practice.

We include additional uncertainty quantification results using AUSE-D and AUSE-V in Tab.~\ref{tab:appx:uq}, covering all baseline methods and ablations. GAVIS achieves the best performance in AUSE-V among all baselines and ablations that disable individual components, which is consistent with its superior active mapping performance and qualitative results. GAVIS also attains the best AUSE-D score among all baseline methods. However, it does not outperform the isotropic ablation that disables anisotropic visibility by setting $\nu(\bm{d}; \bm{d_p}) = 1$. This discrepancy arises from the misalignment between AUSE-D and the objective of active mapping, as discussed in Sec.~\ref{sec:appx_exp}. 
A simple example can be constructed from a scenario similar to Fig.~\ref{fig:ablation_illustration}, where the 3DGS model produces plausible imaginations behind a wall, leading to relatively accurate depth predictions in those regions and consequently improving the overall AUSE-D score, even though the model incorrectly assigns low uncertainty to unobserved areas without considering anisotropic visibility.

\begin{table}[h]
  \centering
  \resizebox{\linewidth}{!}{
    \begin{tabular}{ccccc||ccc}
    \toprule
          & GAVIS & NVF   & FisherRF & VIMC  & Iso.  & w/o DC & Iso. w/o DC \\
    \midrule
    AUSE-D $\downarrow$ & 0.224 & 0.381 & 0.463 & 0.504 & 0.205 & 0.515 & 0.413 \\
    AUSE-V $\downarrow$ & 0.176 & 0.231 & 0.496 & 0.447 & 0.292 & 0.480 & 0.514 \\
    \bottomrule
    \end{tabular}%
}
\caption{Additional uncertainty quantification results
} \label{tab:appx:uq}
\end{table}%

We conduct additional ablation studies on the SH degree $L$, anisotropy parameter $\kappa$, and the density of virtual particles $\rho$ in visibility field density control, compared against the default GAVIS settings ($L=2$, $\kappa=1$, $\rho=100$). The results in Tab.~\ref{tab:appx:ablation}, averaged over all datasets, indicate that GAVIS is relatively robust to these hyperparameters. Specifically,
for $\kappa$, the default $\kappa=1$ performs best overall. Smaller values modestly reduce rendering quality, while overly large values (e.g., $\kappa=10$, a sharper $\nu(\bm{d}; \bm{d_p})$) degrade visual coverage.
For SH degree $L$, $L=3$ yields a slight improvement over $L=2$, but gains plateau at higher degrees; we retain $L=2$ as the default for efficiency.
For virtual particle density $\rho$, higher values consistently improve performance with diminishing returns, and $\rho=100$ provides a practical balance between performance and computational cost.

\subsection{Additional Qualitative Results} \label{sec:appx_qual}

In this subsection, we provide additional comprehensive qualitative results across a wide variety of scenes, demonstrating the generality of GAVIS compared to baseline methods for uncertainty quantification and active mapping.

\begin{figure*}[t]
    \centering
    \includegraphics[width=\textwidth]{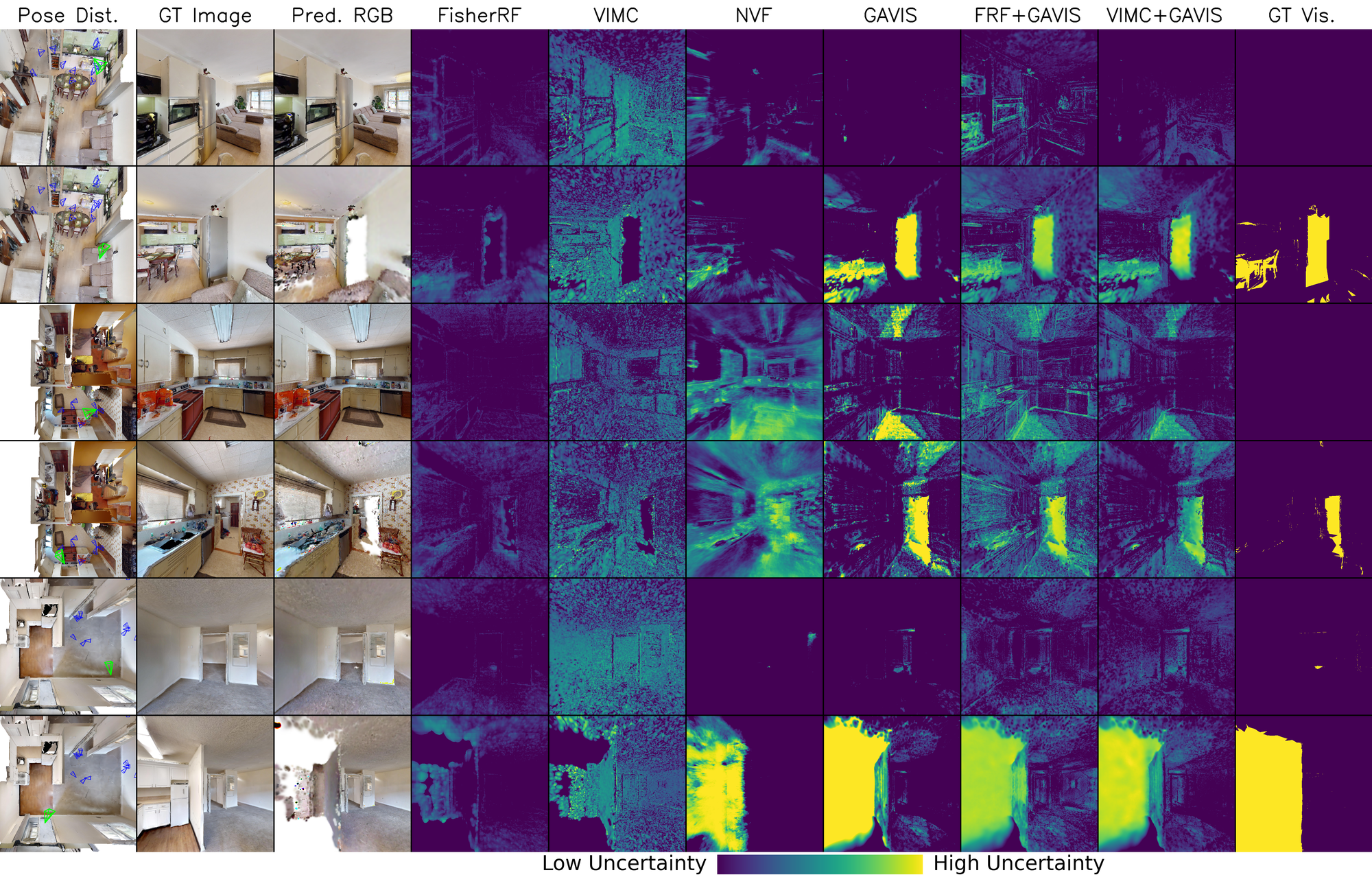}
    \caption{
    \textbf{Qualitative results for uncertainty quantification} in indoor scenes (Gibson and HM3D).  
From left to right:  
(1) Top-down view of the scene with training views (blue frustums) that cover only part of the room, and the queried view for uncertainty evaluation (green frustum);  
(2) Ground-truth RGB image rendered from the scene mesh;
(3) Synthesized RGB image from a 3DGS model trained on the partial-view dataset;  
Uncertainty maps produced by  
(4) FisherRF,  
(5) VIMC,  
(6) NVF,  
(7) GAVIS,  
(8) FisherRF+GAVIS,  
(9) VIMC+GAVIS;  
(10) Ground-truth visibility map used as reference, where bright regions indicate invisible areas and dark regions indicate visible areas.  
An accurate uncertainty quantification method should closely align with the GT visibility by reliably assigning high uncertainty to invisible regions.
All methods are trained using the same set of training views.
}
    \label{fig:enroom}
\end{figure*}

\begin{figure*}[t]
    \centering
    \includegraphics[width=\textwidth]{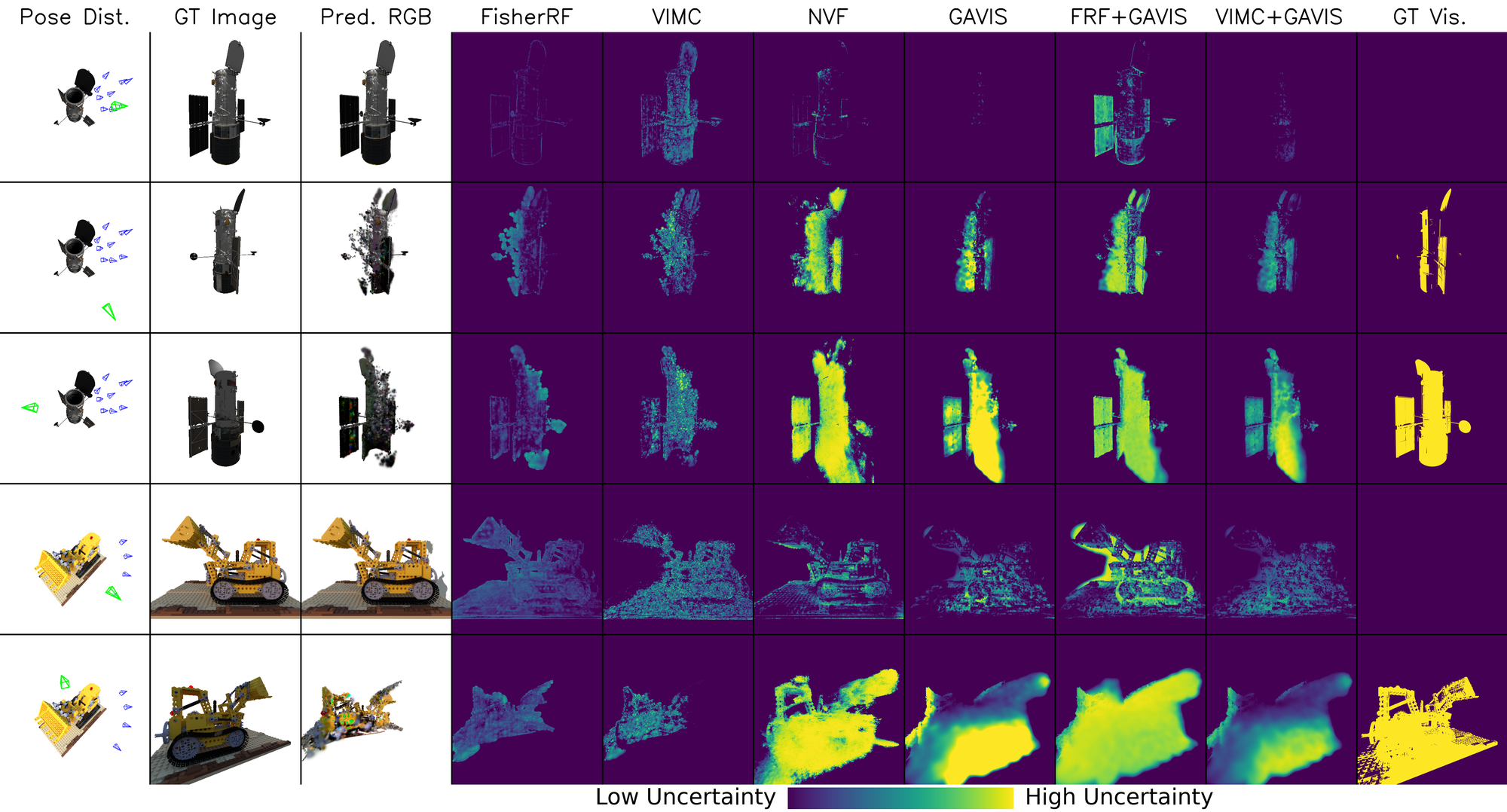}
    \caption{
    \textbf{Qualitative results for uncertainty quantification} in HST (from space dataset) and Lego (from NeRF Synthetic dataset) scenes.  
From left to right:  
(1) Isometric view of the object with training views (blue frustums) that cover only part of the object geometry, and the queried view for uncertainty evaluation (green frustum); 
(2) Ground-truth RGB image rendered from the scene mesh;
(3) Synthesized RGB image from a 3DGS model trained on the partial-view dataset;  
Uncertainty maps produced by  
(4) FisherRF,  
(5) VIMC,  
(6) NVF,  
(7) GAVIS,  
(8) FisherRF+GAVIS,  
(9) VIMC+GAVIS;  
(10) Ground-truth visibility map used as reference, where bright regions indicate invisible areas and dark regions indicate visible areas.  
An accurate uncertainty quantification method should closely align with the GT visibility by reliably assigning high uncertainty to invisible regions.
All methods are trained using the same set of training views.
}

    \label{fig:enobjs}
\end{figure*}

Fig.~\ref{fig:enroom}--\ref{fig:enobjs} visualize uncertainty quantification results. For each scene, all methods are trained on the same set of views, which only partially observe the underlying scene. 
To enable direct evaluation of uncertainty quantification, we generate ground-truth visibility (GT VIS) maps by marking mesh faces that are invisible from the training views with a bright color (and visible faces with a dark color), and rendering this visibility-annotated mesh from the same queried viewpoints. An accurate uncertainty quantification algorithm should assign high uncertainty to invisible regions, consistent with the GT visibility.

As shown in Figs.~\ref{fig:enroom}--\ref{fig:enobjs}, GAVIS provides uncertainty estimation that aligns most closely with the GT VIS maps, thanks to its analytical anisotropic visibility modeling. In contrast, NVF relies on a black-box neural network approximation of isotropic visibility, leading to a noisy and less accurate uncertainty map. All visibility-aware methods significantly outperform baselines that ignore visibility, which fail to assign high uncertainty to invisible regions. Qualitative results for FisherRF+GAVIS and VIMC+GAVIS are also presented, demonstrating significant improvements over the original method by reliably assigning high uncertainty to invisible regions.

\begin{figure*}[t]
    \centering
    \includegraphics[width=\textwidth]{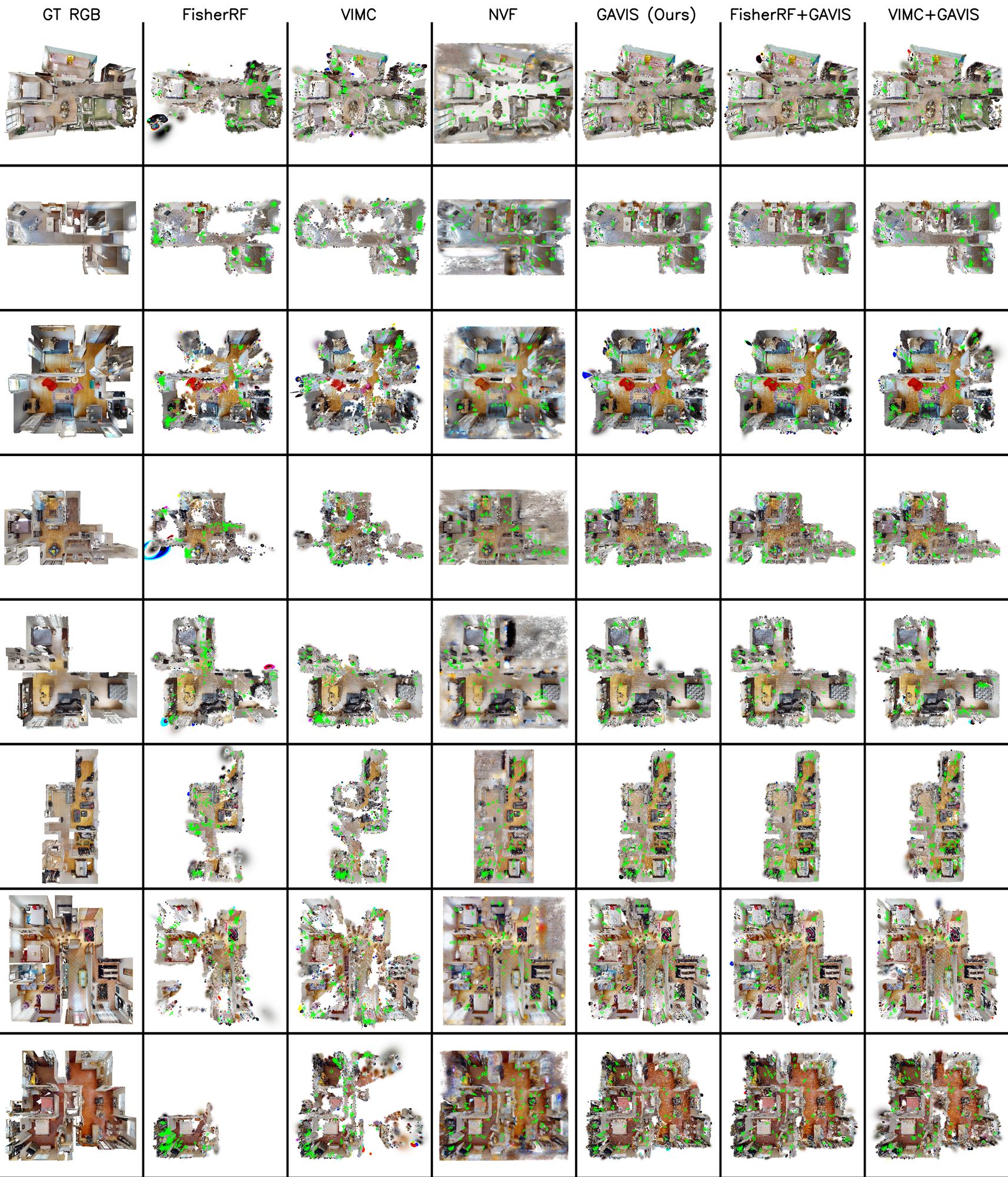}
    \begin{minipage}{\textwidth}
    \caption{
    \textbf{Qualitative results for active mapping} in all HM3D scenes  
From left to right:  
(1) Ground-truth top-down view;  
Reconstruction results from  
(2) FisherRF,  
(3) VIMC,  
(4) NVF,  
(5) GAVIS,  
(6) FisherRF+GAVIS,  
(7) VIMC+GAVIS.
Planned camera poses are shown as green frustums.
    }
        \label{fig:hm3d}
    \end{minipage}

\end{figure*}

\begin{figure*}[t]
    \centering
    \includegraphics[width=\textwidth]{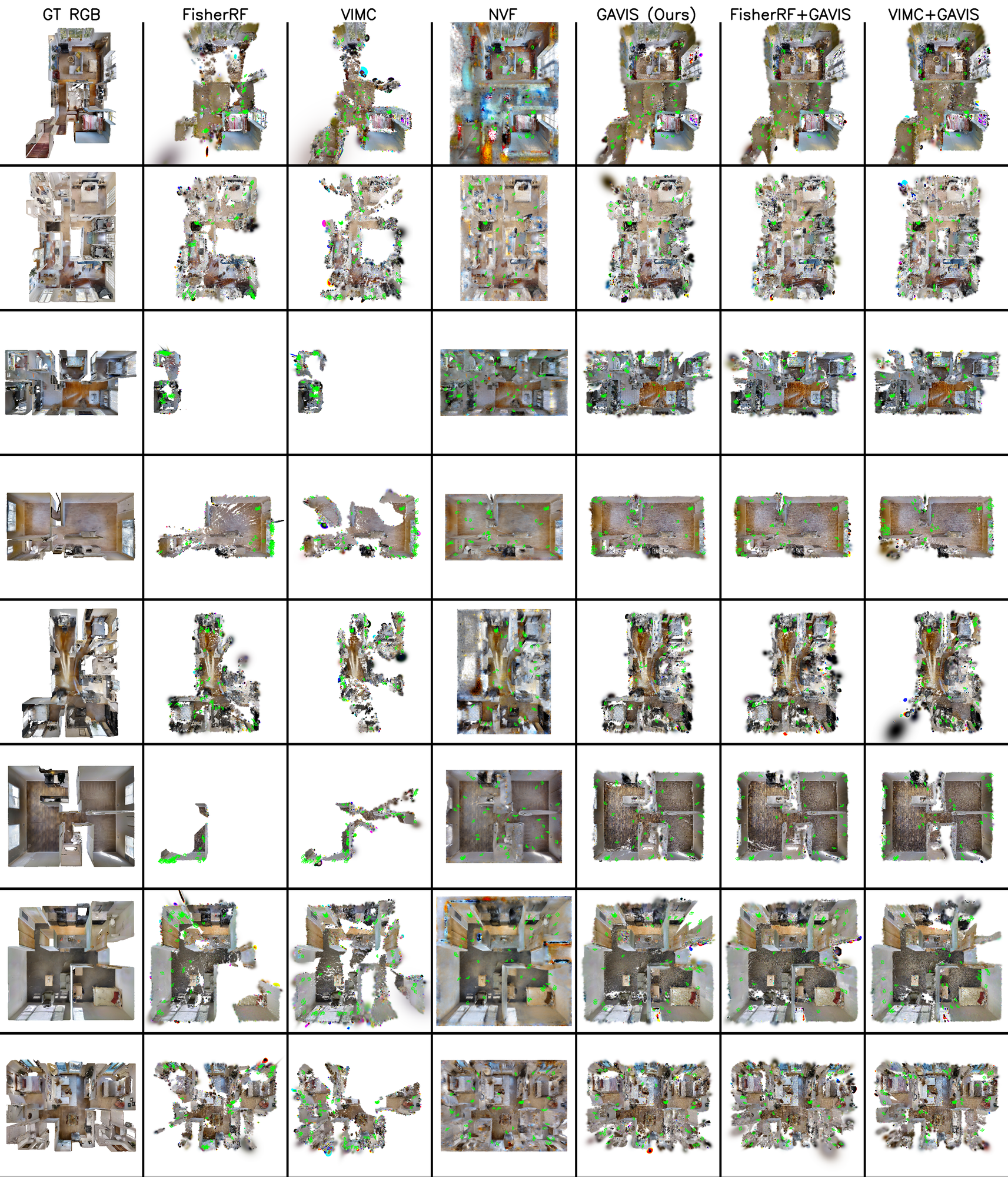}

    \begin{minipage}{\textwidth}
    \caption{
    \textbf{Qualitative results for active mapping} in all Gibson scenes  
From left to right:  
(1) Ground-truth top-down view;  
Reconstruction results from  
(2) FisherRF,  
(3) VIMC,  
(4) NVF,  
(5) GAVIS,  
(6) FisherRF+GAVIS,  
(7) VIMC+GAVIS.
Planned camera poses are shown as green frustums.
    }
        \label{fig:gibson}
    \end{minipage}

\end{figure*}

Finally, additional active-mapping reconstruction results for all Gibson and HM3D scenes are shown in Figs.~\ref{fig:hm3d} and \ref{fig:gibson}, demonstrating that our method leverages accurate uncertainty quantification to guide more effective exploration, resulting in broader scene coverage and higher-quality reconstructions.

\begin{table*}[htbp]
  \centering
  \resizebox{\linewidth}{!}{%
        \begin{tabular}{ccccccccc}
    \toprule
    Dataset & Method & PSNR $\uparrow$ & SSIM $\uparrow$ & LPIPS $\downarrow$ & Acc $\downarrow$ & Comp $\downarrow$ & CR $\uparrow$ & VIS $\uparrow$ \\
    \midrule
    \multirow{6}[2]{*}{\makecell[c]{NeRF \\ Synth.}} & FisherRF & \meanstd{22.34}{0.31} & \meanstd{0.870}{0.004} & \meanstd{0.119}{0.003} & \meanstd{0.014}{2e-4} & \meanstd{0.014}{0.001} & \meanstd{0.626}{0.011} & \meanstd{0.376}{0.010} \\
          & VIMC  & \meanstd{23.14}{0.25} & \meanstd{0.880}{0.003} & \meanstd{0.107}{0.003} & \bestone{\meanstd{0.013}{4e-4}} & \meanstd{0.015}{0.001} & \meanstd{0.651}{0.010} & \meanstd{0.407}{0.011} \\
          & NVF   & \meanstd{22.59}{0.26} & \meanstd{0.859}{0.005} & \meanstd{0.147}{0.004} & \meanstd{0.025}{0.002} & \meanstd{0.020}{0.001} & \meanstd{0.549}{0.014} & \meanstd{0.431}{0.005} \\
          & GAVIS (ours) & \multicolumn{1}{l}{\bestthree{\meanstd{24.26}{0.25}}} & \multicolumn{1}{l}{\bestthree{\meanstd{0.894}{0.002}}} & \multicolumn{1}{l}{\bestthree{\meanstd{0.097}{0.002}}} & \multicolumn{1}{l}{\bestthree{\meanstd{0.014}{2e-4}}} & \multicolumn{1}{l}{\bestone{\meanstd{0.011}{0.001}}} & \multicolumn{1}{l}{\besttwo{\meanstd{0.711}{0.009}}} & \multicolumn{1}{l}{\besttwo{\meanstd{0.437}{0.006}}} \\
          & FisherRF+GAVIS (ours) & \multicolumn{1}{l}{\bestone{\meanstd{24.55}{0.17}}} & \multicolumn{1}{l}{\bestone{\meanstd{0.898}{0.002}}} & \multicolumn{1}{l}{\bestone{\meanstd{0.092}{0.002}}} & \multicolumn{1}{l}{\besttwo{\meanstd{0.013}{3e-4}}} & \multicolumn{1}{l}{\bestthree{\meanstd{0.012}{4e-4}}} & \multicolumn{1}{l}{\bestthree{\meanstd{0.703}{0.008}}} & \multicolumn{1}{l}{\bestone{\meanstd{0.440}{0.005}}} \\
          & VIMC+GAVIS (ours) & \besttwo{\meanstd{24.30}{0.18}} & \besttwo{\meanstd{0.894}{0.002}} & \besttwo{\meanstd{0.096}{0.001}} & \meanstd{0.014}{2e-4} & \besttwo{\meanstd{0.012}{3e-4}} & \bestone{\meanstd{0.713}{0.006}} & \bestthree{\meanstd{0.433}{0.005}} \\
    \midrule
    \multirow{6}[2]{*}{Space} & FisherRF & \meanstd{24.17}{0.08} & \meanstd{0.834}{0.004} & \meanstd{0.158}{0.002} & \meanstd{0.015}{3e-4} & \meanstd{0.020}{0.001} & \meanstd{0.547}{0.017} & \meanstd{0.474}{0.016} \\
          & VIMC  & \meanstd{24.56}{0.55} & \meanstd{0.841}{0.004} & \meanstd{0.150}{0.002} & \bestthree{\meanstd{0.015}{4e-4}} & \meanstd{0.017}{0.003} & \meanstd{0.612}{0.008} & \meanstd{0.510}{0.009} \\
          & NVF   & \meanstd{23.76}{0.46} & \meanstd{0.796}{0.012} & \meanstd{0.202}{0.010} & \meanstd{0.037}{0.003} & \meanstd{0.015}{0.002} & \meanstd{0.499}{0.019} & \meanstd{0.564}{0.017} \\
          & GAVIS (ours) & \multicolumn{1}{l}{\bestone{\meanstd{26.14}{0.10}}} & \multicolumn{1}{l}{\bestone{\meanstd{0.857}{0.003}}} & \multicolumn{1}{l}{\besttwo{\meanstd{0.140}{0.002}}} & \multicolumn{1}{l}{\besttwo{\meanstd{0.014}{4e-4}}} & \multicolumn{1}{l}{\bestone{\meanstd{0.013}{0.002}}} & \multicolumn{1}{l}{\bestthree{\meanstd{0.630}{0.019}}} & \multicolumn{1}{l}{\bestthree{\meanstd{0.582}{0.017}}} \\
          & FisherRF+GAVIS (ours) & \multicolumn{1}{l}{\bestthree{\meanstd{25.44}{0.31}}} & \multicolumn{1}{l}{\bestthree{\meanstd{0.857}{0.003}}} & \multicolumn{1}{l}{\bestone{\meanstd{0.139}{0.002}}} & \multicolumn{1}{l}{\bestone{\meanstd{0.013}{0.001}}} & \multicolumn{1}{l}{\bestthree{\meanstd{0.014}{0.001}}} & \multicolumn{1}{l}{\besttwo{\meanstd{0.631}{0.014}}} & \multicolumn{1}{l}{\besttwo{\meanstd{0.588}{0.023}}} \\
          & VIMC+GAVIS (ours) & \besttwo{\meanstd{25.47}{0.34}} & \besttwo{\meanstd{0.857}{0.002}} & \bestthree{\meanstd{0.141}{0.001}} & \meanstd{0.016}{0.001} & \besttwo{\meanstd{0.013}{2e-4}} & \bestone{\meanstd{0.640}{0.008}} & \bestone{\meanstd{0.597}{0.009}} \\
    \midrule
    \multirow{6}[2]{*}{Gibson} & FisherRF & \meanstd{18.11}{0.50} & \meanstd{0.720}{0.007} & \meanstd{0.419}{0.006} & \besttwo{\meanstd{0.038}{0.001}} & \meanstd{0.891}{0.139} & \meanstd{0.431}{0.031} & \meanstd{0.469}{0.035} \\
          & VIMC  & \meanstd{15.70}{0.20} & \meanstd{0.668}{0.003} & \meanstd{0.465}{0.003} & \meanstd{0.046}{0.002} & \meanstd{1.006}{0.067} & \meanstd{0.337}{0.013} & \meanstd{0.366}{0.014} \\
          & NVF   & \meanstd{23.29}{0.12} & \meanstd{0.798}{0.001} & \meanstd{0.402}{0.002} & \bestone{\meanstd{0.029}{3e-4}} & \bestone{\meanstd{0.033}{0.001}} & \bestone{\meanstd{0.880}{0.002}} & \bestone{\meanstd{0.915}{0.002}} \\
          & GAVIS (ours) & \multicolumn{1}{l}{\besttwo{\meanstd{24.42}{0.14}}} & \multicolumn{1}{l}{\besttwo{\meanstd{0.812}{0.003}}} & \multicolumn{1}{l}{\besttwo{\meanstd{0.323}{0.004}}} & \multicolumn{1}{l}{\bestthree{\meanstd{0.040}{0.001}}} & \multicolumn{1}{l}{\bestthree{\meanstd{0.044}{0.002}}} & \multicolumn{1}{l}{\bestthree{\meanstd{0.831}{0.005}}} & \multicolumn{1}{l}{\besttwo{\meanstd{0.890}{0.003}}} \\
          & FisherRF+GAVIS (ours) & \multicolumn{1}{l}{\bestone{\meanstd{24.58}{0.12}}} & \multicolumn{1}{l}{\bestone{\meanstd{0.815}{0.002}}} & \multicolumn{1}{l}{\bestone{\meanstd{0.319}{0.003}}} & \multicolumn{1}{l}{\meanstd{0.041}{0.001}} & \multicolumn{1}{l}{\besttwo{\meanstd{0.044}{0.002}}} & \multicolumn{1}{l}{\besttwo{\meanstd{0.836}{0.006}}} & \multicolumn{1}{l}{\bestthree{\meanstd{0.890}{0.007}}} \\
          & VIMC+GAVIS (ours) & \bestthree{\meanstd{24.04}{0.19}} & \bestthree{\meanstd{0.807}{0.002}} & \bestthree{\meanstd{0.324}{0.003}} & \meanstd{0.041}{3e-4} & \meanstd{0.075}{0.006} & \meanstd{0.772}{0.009} & \meanstd{0.834}{0.010} \\
    \midrule
    \multirow{6}[2]{*}{HM3D} & FisherRF & \meanstd{18.32}{0.44} & \meanstd{0.693}{0.009} & \meanstd{0.446}{0.009} & \besttwo{\meanstd{0.040}{0.001}} & \meanstd{0.503}{0.097} & \meanstd{0.447}{0.025} & \meanstd{0.558}{0.030} \\
          & VIMC  & \meanstd{17.15}{0.26} & \meanstd{0.645}{0.006} & \meanstd{0.477}{0.006} & \meanstd{0.052}{0.001} & \meanstd{0.178}{0.038} & \meanstd{0.476}{0.017} & \meanstd{0.618}{0.020} \\
          & NVF   & \meanstd{22.69}{0.14} & \meanstd{0.760}{0.002} & \meanstd{0.434}{0.002} & \bestone{\meanstd{0.036}{0.001}} & \bestone{\meanstd{0.040}{0.001}} & \bestthree{\meanstd{0.819}{0.003}} & \bestthree{\meanstd{0.873}{0.001}} \\
          & GAVIS (ours) & \multicolumn{1}{l}{\besttwo{\meanstd{23.97}{0.07}}} & \multicolumn{1}{l}{\besttwo{\meanstd{0.791}{0.001}}} & \multicolumn{1}{l}{\besttwo{\meanstd{0.338}{0.002}}} & \multicolumn{1}{l}{\meanstd{0.043}{4e-4}} & \multicolumn{1}{l}{\besttwo{\meanstd{0.040}{3e-4}}} & \multicolumn{1}{l}{\besttwo{\meanstd{0.820}{0.003}}} & \multicolumn{1}{l}{\besttwo{\meanstd{0.876}{0.001}}} \\
          & FisherRF+GAVIS (ours) & \multicolumn{1}{l}{\bestone{\meanstd{24.23}{0.07}}} & \multicolumn{1}{l}{\bestone{\meanstd{0.797}{0.001}}} & \multicolumn{1}{l}{\bestone{\meanstd{0.330}{0.002}}} & \multicolumn{1}{l}{\bestthree{\meanstd{0.043}{3e-4}}} & \multicolumn{1}{l}{\bestthree{\meanstd{0.040}{3e-4}}} & \multicolumn{1}{l}{\bestone{\meanstd{0.821}{0.002}}} & \multicolumn{1}{l}{\bestone{\meanstd{0.877}{0.001}}} \\
          & VIMC+GAVIS (ours) & \bestthree{\meanstd{23.04}{0.13}} & \bestthree{\meanstd{0.774}{0.002}} & \bestthree{\meanstd{0.347}{0.002}} & \meanstd{0.043}{5e-4} & \meanstd{0.057}{0.002} & \meanstd{0.750}{0.008} & \meanstd{0.826}{0.005} \\
    \bottomrule
    \end{tabular}%
    }
    \caption{\textbf{Quantitative results.} Active mapping performance across all datasets and methods. Best results are shown in \textbf{bold}; second and third best are \underline{underlined}.}
\label{tab:appx:am}
\end{table*}

\begin{table*}[htbp]
    \centering
    \begin{tabular}{cccccccc}
    \toprule
    Setting & PSNR $\uparrow$ & SSIM $\uparrow$ & LPIPS $\downarrow$ & Acc $\downarrow$ & Comp $\downarrow$ & CR $\uparrow$ & VIS $\uparrow$ \\
    \midrule
    GAVIS (default) & \meanstd{24.70}{0.08} & \meanstd{0.839}{0.001} & \meanstd{0.224}{0.001} & \meanstd{0.028}{2e-4} & \meanstd{0.027}{0.001} & \meanstd{0.748}{0.006} & \meanstd{0.697}{0.005} \\
    \midrule
    $\kappa=0$ & \meanstd{24.22}{0.14} & \meanstd{0.832}{0.001} & \meanstd{0.223}{0.001} & \meanstd{0.027}{3e-4} & \meanstd{0.027}{0.001} & \meanstd{0.744}{0.005} & \meanstd{0.666}{0.005} \\
    $\kappa=0.1$ & \meanstd{24.26}{0.18} & \meanstd{0.834}{0.001} & \meanstd{0.222}{0.001} & \meanstd{0.027}{4e-4} & \meanstd{0.026}{0.001} & \meanstd{0.747}{0.005} & \meanstd{0.671}{0.005} \\
    $\kappa=0.3$ & \meanstd{24.39}{0.13} & \meanstd{0.836}{0.002} & \meanstd{0.222}{0.001} & \meanstd{0.028}{3e-4} & \meanstd{0.026}{4e-4} & \meanstd{0.753}{0.006} & \meanstd{0.686}{0.003} \\
    $\kappa=3$ & \meanstd{24.71}{0.11} & \meanstd{0.843}{0.001} & \meanstd{0.216}{0.001} & \meanstd{0.026}{2e-4} & \meanstd{0.043}{0.004} & \meanstd{0.725}{0.007} & \meanstd{0.676}{0.006} \\
    $\kappa=10$ & \meanstd{24.04}{0.16} & \meanstd{0.835}{0.002} & \meanstd{0.223}{0.002} & \meanstd{0.024}{4e-4} & \meanstd{0.112}{0.017} & \meanstd{0.662}{0.009} & \meanstd{0.610}{0.013} \\
    \midrule
    $L=1$ & \meanstd{24.61}{0.10} & \meanstd{0.837}{0.002} & \meanstd{0.224}{0.002} & \meanstd{0.027}{3e-4} & \meanstd{0.028}{0.001} & \meanstd{0.749}{0.005} & \meanstd{0.694}{0.004} \\
    $L=3$ & \meanstd{24.90}{0.09} & \meanstd{0.844}{0.002} & \meanstd{0.217}{0.001} & \meanstd{0.027}{2e-4} & \meanstd{0.027}{0.001} & \meanstd{0.761}{0.003} & \meanstd{0.702}{0.006} \\
    $L=4$ & \meanstd{24.73}{0.10} & \meanstd{0.841}{0.001} & \meanstd{0.220}{0.001} & \meanstd{0.027}{2e-4} & \meanstd{0.027}{0.001} & \meanstd{0.758}{0.004} & \meanstd{0.700}{0.006} \\
    $L=5$ & \meanstd{24.74}{0.15} & \meanstd{0.840}{0.003} & \meanstd{0.220}{0.002} & \meanstd{0.028}{3e-4} & \meanstd{0.027}{0.001} & \meanstd{0.755}{0.008} & \meanstd{0.697}{0.008} \\
    \midrule
    $\rho=10$ & \meanstd{24.53}{0.08} & \meanstd{0.837}{0.001} & \meanstd{0.225}{0.001} & \meanstd{0.028}{3e-4} & \meanstd{0.033}{0.002} & \meanstd{0.738}{0.005} & \meanstd{0.685}{0.006} \\
    $\rho=25$ & \meanstd{24.54}{0.12} & \meanstd{0.836}{0.001} & \meanstd{0.224}{0.001} & \meanstd{0.028}{2e-4} & \meanstd{0.031}{0.001} & \meanstd{0.734}{0.003} & \meanstd{0.683}{0.003} \\
    $\rho=50$ & \meanstd{24.54}{0.11} & \meanstd{0.839}{0.001} & \meanstd{0.222}{0.001} & \meanstd{0.028}{3e-4} & \meanstd{0.029}{0.001} & \meanstd{0.750}{0.003} & \meanstd{0.693}{0.003} \\
    $\rho=200$ & \meanstd{24.99}{0.08} & \meanstd{0.844}{0.001} & \meanstd{0.216}{0.001} & \meanstd{0.027}{2e-4} & \meanstd{0.026}{2e-4} & \meanstd{0.765}{0.003} & \meanstd{0.703}{0.004} \\
    $\rho=400$ & \meanstd{24.95}{0.09} & \meanstd{0.844}{0.001} & \meanstd{0.215}{0.001} & \meanstd{0.027}{2e-4} & \meanstd{0.025}{3e-4} & \meanstd{0.766}{0.005} & \meanstd{0.701}{0.004} \\
    $\rho=1000$ & \meanstd{25.07}{0.10} & \meanstd{0.847}{0.001} & \meanstd{0.212}{0.001} & \meanstd{0.026}{2e-4} & \meanstd{0.030}{0.004} & \meanstd{0.763}{0.004} & \meanstd{0.700}{0.005} \\
    \bottomrule
    \end{tabular}%
    \caption{\textbf{Additional ablation studies.} Active mapping performance of GAVIS with different hyperparameter settings. GAVIS uses $\kappa=1$, $L=2$ and $\rho=100$ by default. } \label{tab:appx:ablation}
\end{table*}

\newpage

\clearpage

\clearpage

\clearpage

\clearpage

\end{document}